\documentclass{article}\usepackage{knitr}\usepackage{knitr}

\title{Experimental Investigation and Evaluation of Model-based Hyperparameter Optimization}
\author{
Eva Bartz\\
Bartz \& Bartz GmbH\\
Goebenstr.~10\\ 
51643 Gummersbach\\
eva.bartz@bartzundbartz.de\\
\and
Martin Zaefferer\\
Bartz \& Bartz GmbH\\
Goebenstr.~10\\ 
51643 Gummersbach\\
martin.zaefferer@bartzundbartz.de\\
\and
Olaf Mersmann\\
Bartz \& Bartz GmbH\\
Goebenstr.~10\\ 
51643 Gummersbach\\
olaf.mersmann@bartzundbartz.de\\
\and
Thomas Bartz-Beielstein\\
IDE+A, TH Köln \\
Steinmüllerallee 1\\ 
51643 Gummersbach\\
thomas.bartz-beielstein@th-koeln.de\\
}

\date{\today}

\usepackage[nottoc]{tocbibind}
\usepackage{graphicx}
\usepackage{arxiv}
\usepackage{gensymb}
\usepackage{placeins}
\usepackage{url}
\usepackage{dirtree}
\usepackage{xcolor}
\usepackage{listings}
\usepackage{paralist}
\usepackage{tabulary} 
\usepackage{todonotes}
\usepackage{amsmath}
\usepackage{amssymb}
\usepackage[margin=10pt,font=small,labelfont=bf]{caption}
\usepackage{hhline}
\usepackage{booktabs}
\usepackage{pdfpages}
\usepackage{placeins}
\usepackage{multicol}
\usepackage{gensymb}
\usepackage{fancyhdr}
\usepackage{xspace}
\usepackage{natbib}
\usepackage{pdfpages}
\usepackage{xspace}
\usepackage{nomencl}
\usepackage{glossaries}
\usepackage{csquotes}

\usepackage{bartr,bartvec,bartacronyms,bartmisc}

\newacronym{AI}{AI}{Artificial Intelligence}
\newacronym{AID}{AID}{Automatic Interaction Detection}
\newacronym{BCE}{BCE}{Binary Cross Entropy}
\newacronym{BLAS}{BLAS}{Basic Linear Algebra Subprograms}
\newacronym{CAM}{CAM}{Class Activation Map}
\newacronym{CART}{CART}{Classification And Regression Trees}
\newacronym{CCE}{CCE}{Categorical Cross Entropy}
\newacronym{CID}{CID}{Census-Income (KDD) Data Set}
\newacronym{CNN}{CNN}{Convolutionary Neural Network}
\newacronym{CTC}{CTC}{Connectionist Temporal Classification}
\newacronym{cuDNN}{cuDNN}{CUDA Deep Neural Network}
\newacronym{CV}{CV}{Cross Validation}
\newacronym{DL}{DL}{Deep Learning}
\newacronym{DT}{DT}{Decision Tree}
\newacronym{IMDB}{IMDB}{Internet Movie Data Base}
\newacronym{JPG}{JPG}{Joint Photographic Experts Group}
\newacronym{LOOCV}{LOOCV}{Leave One Out Cross Validation}
\newacronym{MAE}{MAE}{Mean Absolute Error}
\newacronym{ML}{ML}{Machine Learning}
\newacronym{MLR}{mlr}{Machine Learning in R}
\newacronym{MSE}{MSE}{Mean Squared Error}
\newacronym{GB}{GB}{Gradient Boosting}
\newacronym{HPO}{HPO}{Hyperparameter Optimization}
\newacronym{HPT}{HPT}{Hyperparameter Tuning}
\newacronym{GBM}{GBM}{Gradient Boosting Model}
\newacronym{MNIST}{MNIST}{Modified National Institute of Standards and Technology}
\newacronym{NN}{NN}{Neural Network}
\newacronym{R2}{$R^2$}{R-squared}
\newacronym{RGB}{RGB}{Red, Green, and Blue color space}
\newacronym{RMSE}{RMSE}{Root Mean Squared Error}
\newacronym{SGD}{SGD}{Stochastic Gradient Descent}
\newacronym{SPOT}{SPOT}{Sequential Parameter Optimization Toolbox}
\newacronym{SPOTMisc}{SPOTMisc}{Sequential Parameter Optimization Toolbox -- Miscelleanous Functions}
\newacronym{SSQ}{SSQ}{Sum of Squares}
\newacronym{SVM}{SVM}{Support Vector Machine}
\newacronym{XGBoost}{XGBoost}{Extreme Gradient Boosting}

\usepackage{acronym}

\definecolor{Gray}{gray}{0.85}

\usepackage{amsthm}
\theoremstyle{definition}

\IfFileExists{upquote.sty}{\usepackage{upquote}}{}
\begin{document}

\graphicspath{{.}{Figures.d/}{./figures/}}

\maketitle

\begin{abstract}
Machine learning algorithms such as random forests or xgboost are gaining more 
importance and are increasingly incorporated into production processes in order 
to enable comprehensive digitization and, if possible, automation of processes. 
Hyperparameters of these algorithms  used have to be set appropriately, which can 
be referred to as \emph{hyperparameter tuning} or optimization. Based on the concept of 
\emph{tunability}, this article presents an overview of theoretical and practical 
results for popular machine learning algorithms. 
This overview is accompanied by an experimental analysis of 30 hyperparameters from six relevant machine learning algorithms. 
In particular, it provides
(i) a survey of important hyperparameters, 
(ii) two parameter tuning studies, and
(iii) one extensive global parameter tuning study,  as well as 
(iv) a new way, based on consensus ranking, to  analyze results from multiple algorithms.
The R package mlr is used as a uniform interface to the machine learning models. The R package SPOT is used to perform the actual tuning (optimization).
All additional code is provided together with this paper.
\end{abstract}


\section{Introduction}\label{ch:intro}


\gls{AI} and \gls{ML} methods are gaining more importance and are increasingly incorporated into production processes in order to enable comprehensive digitization and, if possible, automation of processes. 
As a rule, several hyperparameters of the methods used have to be set appropriately. 
Previous projects carried out produced inconsistent results in this regard. 
For example, with \glspl{SVM} it could be observed that the tuning of the hyperparameters is critical to success with the same data material, with random forests the results do not differ too much from one another despite different selected hyperparameter values. 
While some methods have only one or a few hyperparameters, others provide a large number. 
In the latter case, optimization using a (more or less) fine grid (grid search) quickly becomes very time-consuming and can therefore no longer be implemented. In addition, the question of how the optimality of a selection can be measured in a statistically valid way (test problem: training / validation / test data and resampling methods) arises for both many and a few hyperparameters.


This article deals with the hyperparameter tuning of \ac{ml} algorithms.
In particular, it provides
\begin{itemize}
\item a survey of important model parameters, 
\item two parameter tuning studies and 
\item one extensive global parameter tuning study,  and 
\item a new way, based on consensus ranking, to  analyze results from multiple algorithms. 
\end{itemize}

More than 30 hyperparameters from six relevant \gls{ML} models are analyzed. 
This paper is structured as follows: Sec.~\ref{sec:survey} presents models (algorithms) and hyperparameters. \gls{HPT} methods are introduced in Sec.~\ref{ch:tuning}. 
Three case studies are presented next. These hyperparameter tuning studies are using the \gls{CID}, which will be described in Sec.~\ref{sec:census}.
The first case study, which describes \gls{HPT} for decision trees (\Rrpart), is presented in Sec.~\ref{sec:case1}. This case study might serve as a starting point for the interested reader, because \Rrpart can be evaluated very quickly. 
The second case study analyses \acl{xgb} and is presented in Sec.~\ref{sec:case2}. A global case study, which analyses \emph{tunability}, is presented in Sec.~\ref{sec:case3}.
Results are discussed in Sec.~\ref{sec:conclusion}. 

\section{Models and Hyperparameters}\label{sec:survey}
\subsection{Overview}
\subsubsection{Methodology}
In the following, we provide a survey and description of relevant parameters
of six different \ac{ml} models.
We emphasize that this is not a complete list of their parameters,
but covers parameters that are set quite frequently according to the literature.

Since the specific names and meaning of hyperparameters
may depend on the actual implementation used,
we have chosen a reference implementation for each model.
The implementations chosen are all packages from the statistical programming language
R.
Thus, we provide a description that is consistent with
what users experience, so that they can identify the relevant parameters
when tuning \ac{ml} algorithms in practice.

In particular, we cover the following models, with the respective 
R packages:

\begin{itemize}
\item \acl{knn}: \modkknn
\item \acl{en}: \modglmnet
\item \acl{eb}: \modrpart
\item \acl{rf}: \modranger
\item \acl{xgb}: \modxgb
\item \acl{svm}: \modsvm
\end{itemize}

\begin{table}
\centering
\caption{Overview. Models and parameters analyzed in this study.}\label{tab:paraOver}
\begin{tabulary}{\textwidth}{L | L | L}
\hline
Model & Hyperparameter \hspace*{2cm} & Comment \\
\hline
knn &\paramk  & number of neighbors  \\
&\paramp  & $p$ norm  \\
\hline
Elastic net &\paramalpha  & weight term of the loss-function \\
&\paramlambda &  trade-off between model quality and complexity\\
&\paramthresh  & threshold for model convergence, i.e., convergence of the internal coordinate descent\\
\hline
Decision trees&\paramminsplit  & minimum number of observations required for a split\\
&\paramminbucket & minimum number of observations in an end node (leaf) \\
&\paramcp   &  complexity parameter\\
&\parammaxdepth  & maximum depth of a leaf in the decision tree\\
\hline
Random forest & \paramnumtrees &  number of trees that are combined in the overall ensemble model \\
&\parammtry & number of randomly chosen features are considered for each split\\
&\paramsamplefraction & number of observations that are randomly drawn for training a specific tree\\
&\paramreplace   & replacement of randomly drawn samples\\
&\texttt{respect.-}  &  \\
&\texttt{unordered.factors}  & handling of splits of categorical variables \\
\hline
xgBoost & \parameta & learning rate, also called \enquote{shrinkage} parameter.\\
&\paramnrounds & number of boosting steps\\
&\paramlambda   & regularization of the model\\
&\paramalpha   & parameter for the L1 regularization of the weights \\
&\paramsubsample  & portion of the observations that is randomly selected in each iteration\\
&\paramcolsample   & number of features that chosen for the splits of a tree\\
&\paramgamma & number of splits of a tree by assuming a minimal improvement for each split\\
&\parammaxdepthx & maximum depth of a leaf in the decision trees\\
&\paramminchild  & restriction of the number of splits of each tree\\
\hline
Support vector machines &\paramdegree   & degree of the polynomial (parameter of the polynomial kernel function)\\
&\paramgamma & parameter of the polynomial, radial  basis, and sigmoid kernel functions\\
&\paramcoefz & parameter of the polynomial and sigmoid kernel functions\\
&\paramcost & regularization parameter, weights constraint violations of the model\\
&\paramepsilon & regularization parameter, defines ribbon around predictions\\
\hline
\end{tabulary}
\end{table}

Table~\ref{tab:paraOver} presents an overview of these algorithms and their associated hyperparameters.
After a short, general description of the specific hyperparameter, the following features will be described for every hyperparameter:
\begin{description}
\item[Type:] Describes the type (e.g., integer) and complexity (e.g., scalar). These data types are described in Sec.~\ref{sec:datatypes}.
\item[Sensitivity / robustness:]
Describes how much the parameter is affected by changes, i.e, how much its setting depends on the considered problem and data. There is a close relationship between sensitivity and tunability as defined by \citep{Prob19a}, because  tunability is the potential for improvement of the parameter in the vicinity of a reference value.
\item[Heuristics for determination:] 
Describes ways to find good hyperparameter settings.
\item[Plausible range of values:] Describes feasible values, i.e., lower and upper bounds, constraints, etc.
\item[Further constraints:] Additional constraints, specific for certain settings or algorithms.
\item[Interactions:] Describes interactions between the parameters.
\end{description}
Each algorithm description concludes with a brief survey of examples from the literature, that gives hints how the algorithm was tuned.

\subsubsection{Data Types}\label{sec:datatypes}
Our implementation is done in the R programming language, where data and functions are represented as objects.
Each object has a data type. The basic (or \emph{atomic}) data types are shown in Table~\ref{tab:datatype}.

\begin{table}
\caption{Atomic data types of the programming language R}
\label{tab:datatype}
\begin{tabular}{lll}
\hline
Data type & Description & Examples \\
\hline
NULL & empty set & NULL \\
logical & boolean values & TRUE, FALSE \\
numeric & integer and real values & 1, 0.5\\
complex & complex values & 1+1i\\
character & characters and strings of characters & "123", "test"\\
\hline 
\end{tabular}
\end{table}
In addition to these data types, R uses an \emph{internal storage mode} which can be queried using \texttt{typeof()}\/.
Thus there are two storage modes for the \texttt{numeric} data type:
\begin{itemize}
\item \texttt{integer}\/ for integers and
\item \texttt{double}\/ for real values.
\end{itemize}
The corresponding variables are referred to as \emph{numeric}. \emph{Factors}\/ are used in R to represent nominal (qualitative) features. Ordinal features can also be represented by factors in R (see argument \texttt{ordered} of the function \texttt{factor()}\/). However, this case is not considered here. Factors are generated with the generating function \texttt{factor()}. 
Factors are not atomic data types. Internally in R, factors 
are stored by numbers (integers), externally the name of the factor is used.
We call the corresponding variables \emph{categorical}.

\subsubsection{Terminology}
This article deals with the tuning of the hyperparameters of 
methods from the field of supervised learning (both regression and classification).
In the following, the scope of application is explained, including a terminology of important terms.

The \emph{data points}\/ $\xtrain$ come from an input data space $\mathcal{X}$ ($\xtrain \in \mathcal{X}$) and 
can have different scale levels (nominal: no order; ordinal: order, no distance; cardinal: order, distances). 
Nominal and ordinal data are mostly discrete, cardinal data are continuous.
Usually, the data points are $n$-dimensional vectors. The vector elements are also called features or dependent variables.
The number of data points in a data set is $m$.

In addition, we consider output data (dependent variables) $\ytrain \in \mathcal{Y}$. These can also have different scale levels (nominal, ordinal, cardinal). Output data are usually scalar.
Variable identifiers like \xtrain and \ytrain represent scalar or vector quantities, the meaning results from the context.

We study \gls{ML} algorithms,  also referred to as models or methods.
\emph{Supervised learning}\/ means that for each observed data point $\xtrain_i$ with $i=1,2,...,m$ of the input space
 an associated dependent output value $\ytrain_i$ is known. 
Supervised learning models learn the relationship between input and output. The goal is to predict the output value for a new/unknown data point $\xtest$: $\ytest = \f(\xtest)$, with $f: \mathcal{X} \mapsto \mathcal{Y}$.

\emph{Regression}\/ deals with a real output space ($\ytrain_i \in \mathbb{R}$). 
Integer output spaces are also included in this domain.
Example question: how does relative humidity depend on temperature?

\emph{Classification}\/ deals with an output space of discrete classes (categories), $\ytrain_i \in \{a_1, a_2, ... ,a_c\}$. Where $c \in \mathbb{N}$ is the natural number of classes.
Example: How does the default on a loan (yes, no) depend on the income of the borrower?

The \emph{model quality}\/ describes the performance of a model. 
This can be expressed in different ways, e.g. as a measure of the fit of the model to the observed data values.

\emph{Hyperparameters}\/ are settings or configurations of the methods (models), 
which are freely selectable within a certain range and influence model quality 
Example of hyperparameters: The number of units or layers of a neural network.
The number of hyperparameters of a model is $d$.
Hyperparameters are distinct from model parameters, among others. Model parameters are chosen during the learning process by the model itself.
Example of model parameters: The weights of each connection in a neural network.

The determination of the best possible hyperparameters is called \emph{tuning} (\gls{HPT}).
For the tuning procedure (or tuner for short), we use a model-based search. 
The corresponding model is called a surrogate model.

The term \emph{tunability}\/ is used according to the definition presented in~\citet{Prob19a}.
Tunability describes a measure for modeling algorithms as well as for individual hyperparameters. 
It is the difference between the model quality for default values (or reference values) 
and the model quality for optimized values (after tuning is completed).
Or in the words of ~\citet{Prob19a}: \enquote{measures for quantifying the tunability of the whole algorithm
and specific hyperparameters based on the differences between the performance of default
hyperparameters and the performance of the hyperparameters when this hyperparameter is
set to an optimal value}.
Tunability of individual hyperparameters can also be used as a measure of their 
\emph{relevance}, 
\emph{importance}, or 
\emph{sensitivity}.
Parameters with high tunability are 
accordingly of greater importance for the model. 
The model reacts strongly to (i.e., is sensitive to) changes in these hyperparameters.

The term \emph{complexity}\/ or model complexity generally describes, 
how many functions of different difficulty can be represented by a model. 
Example 1: for linear models, this can be influenced by the number of model coefficients.
Example 2: for support vector machines, this can be influenced by the parameter \paramcost.

\subsection{Model: \acl{knn}}\label{sec:knn}
The \ac{knn} model determines for each $\xtrain'$ the $n$ neighbors with the least distance to $\xtrain'$,
e.g., based on the Minkowski distance: 
\begin{equation}\label{eq:minkowski}
\text{d}(\xtrain,\xtrain')= \left(\sum_{i=1}^n |x_i - x'_i|^\paramp \right)^{1/\paramp} = ||\xtrain - \xtrain' ||_\paramp.
\end{equation}
For regression, the mean of the neighbors is used~\citep{Jame17a}.
For classification, the prediction of the model is the most frequent class observed in the neighborhood.
Two relevant hyperparameters (\paramk, \paramp) result from this.
Additionally, one categorical hyperparameter could be considered: 
the choice of evaluation algorithm (e.g., choosing between brute force or KD-Tree)~\citep{Frie77a}.
However, this mainly influences computational efficiency, rather than actual performance.

We consider the implementation from the R package \modkknn\footnote{\url{https://cran.r-project.org/package=kknn}}~\citep{Schl16a}.

\subsubsection{\acl{knn} Parameter: \paramk}
The parameter \paramk determines the number of neighbors that are considered
by the model.
In case of regression, it affects how smooth the predicted function of the model is.
Similarly, it influences the smoothness of the decision boundary in case of classification.

Small values of \paramk lead to fairly non-linear predictors (or decision boundaries),
while larger values tend towards more linear shapes~\citep{Jame17a}. 
The error of the model at any training data sample is zero if \paramk=1 
but this does not determine anything about the generalization error~\citep{Jame17a}. 
Larger values of \paramk may help to deal with rather noisy data.
Moreover, larger values of \paramk increase the runtime of the model.

\begin{description}
\item[Type:] integer, scalar.
\item[Sensitivity / robustness:]
Determining the size of the neighborhood via \paramk is a fairly sensitive decision.
\citet{Jame17a} describe this as a drastic effect.
However, this is only true as long as the individual classes are hard to separate (in case of classification).
If there is a large margin between classes, the shape of the decision boundary becomes less relevant (see \citep[Fig. 3]{Domi12a}). 
Thus, the sensitivity of the parameter depends on the considered problem and data.
\citet{Prob19a} also identify \paramk as a sensitive (or \textit{tunable}) parameter.
There is a close relationship between sensitivity and tunability as defined by Probst et al,
since tunability is the potential for improvement of the parameter in the vicinity of a reference value.
\item[Heuristics for determination:] 
As mentioned above, the choice of \paramk may depend on properties of the data.
Hence, no general rule can be provided.
In individual cases, determining the distance between and within classes may help to find an approximate value: 
\paramk=1 is better than $\paramk>1$,
if the distance within classes is larger than the distance between classes~\citep{Cove67a}.
Another empirical suggestion from the literature is $\paramk=\sqrt{m}$~\citep{Lall96a,Prob19a}.
\item[Plausible range of values:] $\paramk\geq 1$, $\paramk \ll m$, where $m$ is the number of data samples. Only integer values are valid.
\item[Further constraints:] none.
\item[Interactions:] 
We are not aware of any interactions between the parameters.
However, both \paramk and \paramp change the perceived 
neighborhood of samples and thus the shape of the decision boundaries.
Hence, an interaction between these parameters is likely.
\end{description}

\subsubsection{\acl{knn} Parameter: \paramp}
The parameter \paramp affects the distance measure that is used to determine the nearest neighbors in \ac{knn}.
Frequently, this is the Minkowski distance: $\text{d}(\xtrain,\xtrain')= (\sum_{i=1}^n |x_i - x'_i|^\paramp )^{1/\paramp}$.
Moreover, it has to be considered that other distances could be chosen for non-numerical features of the data set (i.e., Hamming distance for
categorical features).
The implementation used in this article (\modkknn) transforms categorical variables into numerical variables via dummy-coding, then
using the Minkowski distance on the resulting data.
Similarly to  \paramk, \paramp changes the observed neighborhood.
While \paramp does not change the number of neighbors, it still affects the choice of neighbors.

\begin{description}
\item[Type:] double, scalar.
\item[Sensitivity / robustness:]
It has to be expected, that the model is less sensitive to changes in \paramp than to changes in \paramk,
since fairly extreme changes are required to change the neighborhood set of a specific data sample.
This explains, why many publications do not consider \paramp during tuning, see Table~\ref{table:survey:knn}.
However, the detailed investigation of~\citet{Alfe19a} showed, that changes of the distance measure
can have a significant effect on the model accuracy. Alfeilat et al. 
only tested special cases of the Minkowski distance: 
Manhattan distance ($\paramp=1$), Euclidean distance ($\paramp=2$) and Chebyshev distance ($\paramp=\infty$).
They give no indication whether other values may be of interest as well.
\item[Heuristics for determination:] 
The choice of distance measure (and hence \paramp) depends on the data,
a general recommendation or rule-of-thumb is hard to derive~\citep{Alfe19a}.
\item[Plausible range of values:] 
Often, the interval $1 \leq \paramp \leq 2$ is considered. 
The lower boundary is $\paramp>0$.
Note: The Minkowski distance is not a metric if $\paramp<1$~\citep{Alfe19a}.
Theoretically, a value of $\paramp=\infty$ is possible (resulting into Chebyshev distance), 
but this is not possible in the \modkknn implementation.
\item[Further constraints:] none.
\item[Interactions:] 
We are not aware of any known interactions between parameters.
However, both \paramk and \paramp change what is perceived as 
the neighborhood of samples, and hence the shape of decision boundaries.
Hence, an interaction between those parameters is likely.
\end{description}

In conclusion, Table~\ref{table:survey:knn} provides a brief survey of
examples from the literature, where \ac{knn} was tuned.

\begin{table}[!htbp]
\caption{Survey of examples from the literature, for tuning of \ac{knn}. Asterisk (*) denotes results that depend on data set (multiple data sets).}
\centering
\begin{tabulary}{\textwidth}{L | R  | R | R | L}
\hline
Hyperparameter & Lower bound & Upper bound & Result & Notes  \\
\hline
\multicolumn{5}{l}{\citep{Schr19a}, weighted \ac{knn} variant, spacial data, 1 data set}\\
\hline
\paramk&10&400& NA & \\
\paramp&1&100& NA & integer\\
\hline
\multicolumn{5}{l}{\citep{Khan20a}, detection of bugs, 5 data sets}\\
\hline
\paramk&1&17& NA & \\
\paramp&0,5&5& NA &\\
\hline
\multicolumn{5}{l}{\citep{Osma17a}, detection of bugs, 5 data sets}\\
\hline
\paramk&1&5& 2 or  5 *& \\
\hline
\multicolumn{5}{l}{\citep{Prob19a}, various applications, 38 data sets}\\
\hline
\paramk&1&30& 2 to 30 *& \\
\hline
\multicolumn{5}{l}{\citep{Doan20a}, impact damage on reinforced concrete, 1 data set}\\
\hline
\paramk&7&51&9& \\
\paramp&1&11&3&\\
\hline
\end{tabulary}
\label{table:survey:knn}
\end{table}

\subsection{Model: Regularized regression---\acl{en}}\label{sec:en}
\ac{en} is a  regularized regression method~\citep{Zou05a}.
Regularized regression can be employed to fit regression models with a reduced number
of model coefficients. Special cases of \ac{en} are Lasso  and Ridge regression.

Regularization is useful, when data sets are high-dimensional (especially but not exclusively
if $n>m$),
or when variables in the data sets are heavily correlated with each other~\citep{Zou05a}.
Less complex models (i.e., with fewer coefficients) help to reduce overfitting\footnote{Overfitting means, that the model
is extremely well adapted to the training data, but generalizes poorly as a result (i.e., predicts
poorly for unseen data).}.
The resulting models are also easier to understand for humans, due to their reduced complexity.

During training, non-regularized regression reduces only the model error (e.g., via the least squares method).
\ac{en} also considers a penalty term, which grows with the number of coefficients included in the model (i.e., are non-zero).

As a reference implementation, we use the R package \modglmnet\footnote{\url{https://cran.r-project.org/package=glmnet}}~\citep{Hast20b,Simo11a}.

\subsubsection{\acl{en} Parameter: \paramalpha}
The parameter \paramalpha ($\alpha$) weights the two elements of the penalty term of the loss-function in the \ac{en} model~\citep{Frie10a}:
\begin{equation}\label{eq:enloss}
\min_{\beta_0,\beta} \frac{1}{2n}\sum_{i=1}^n (\ytrain_i-\beta_0 - \xtrain_i^{\T} \beta)^2 + \lambda P(\alpha,\beta).
\end{equation}
 The penalty term $P(\alpha,\beta)$ is~\citep{Frie10a}:
\begin{equation}
(1-\alpha) \frac{1}{2}||\beta||_2^2+\alpha||\beta||_1,
\end{equation}
with the vector of $p$ model coefficients $\beta\in \RR^p$ and the `intercept' coefficient $\beta_0 \in \RR $.
The value \paramalpha=0 corresponds to the special case of Ridge regression, \paramalpha=1 corresponds to Lasso regression~\citep{Frie10a}.

The parameter \paramalpha allows to find a compromise or trade-off between
Lasso and Ridge regression.
This can be advantageous, since both variants have different consequences.
Ridge regression affects that coefficients of strongly correlated variables
match to each other (extreme case: identical variables receive identical coefficients)~\citep{Frie10a}.
In contrast, Lasso regression tends to lead to a single coefficient in such a case (the other coefficients being zero)~\citep{Frie10a}.

\begin{description}
\item[Type:] double, scalar.
\item[Sensitivity / robustness:]
Empirical results from Friedman et al. show that the
\ac{en} model can be rather sensitive to changes in \paramalpha~\citep{Frie10a}.
\item[Heuristics for determination:] 
We are not aware of any heuristics to set this parameter. 
As described by~\citet{Frie10a}, \paramalpha can be set to a value of close to 1,
if a model with few coefficients without risk of degeneration is desired.
\item[Plausible range of values:] \paramalpha $\in[0,1]$.
\item[Further constraints:] none.
\item[Interactions:] \paramlambda interacts with \paramalpha, see  Sec.~~\ref{sub:en:lambda}.
\end{description}

\subsubsection{\acl{en} Parameter: \paramlambda}~\label{sub:en:lambda}
The hyperparameter \paramlambda influences the impact of the penalty term $P(\alpha,\beta)$ in equation~\ref{eq:enloss}.
Very large \paramlambda lead to many model coefficients ($\beta$) being set to zero. 
Correspondingly, only few model coefficients become zero if \paramlambda is small (close to zero).
Thus, \paramlambda is often treated differently than other hyperparameters:
In many cases, several values of \paramlambda are of interest, rather than a single value~\citep{Simo11a}.
There is no singular, optimal solution for \paramlambda, as it controls the trade-off between model quality and complexity (number of coefficients that are not zero).
Hence, a whole set of \paramlambda values will often be suggested to users, who
are then able to choose a resulting model that provides a specific trade-off
of their liking.

\begin{description}
\item[Type:] double, scalar.
\item[Sensitivity / robustness:] \ac{en} is necessarily sensitive to \paramlambda, since extreme values
lead to completely different models (i.e., all coefficients are zero or none are zero).
This is also shown in figure 1 by~\citet{Frie10a}.
\item[Heuristics for determination:] 
Often, \paramlambda gets determined by a type of grid-search, 
where a sequence of decreasing \paramlambda is tested~\citep{Frie10a,Simo11a}.
The sequence starts with a sufficiently large value of \paramlambda,
such that $\beta=0$. 
The sequence ends, if the resulting model starts to approximate the unregularized model~\citep{Simo11a}.
\item[Plausible range of values:] 
\paramlambda $\in (0,\infty)$ (Note: \paramlambda=0 is possible, but leads to a simple unregularized model). 
Using a logarithmic scale seems reasonable, as used in the study by~\citet{Prob19a}, 
to cover a broad spectrum of very small and very large values.
\item[Further constraints:] none.
\item[Interactions:] \paramlambda interacts with \paramalpha.
Both are central for determining the coefficients
$\beta$ (see also \citep[Fig. 1]{Frie10a}).
\end{description}

\subsubsection{\acl{en} Parameter: \paramthresh}

The parameter \paramthresh is a threshold for model convergence (i.e., convergence of the internal coordinate descent).
Model training ends, when the change after an update of the coefficients drops below this value~\citep{Hast20b}.
Unlike parameters like \paramlambda, \paramthresh is not a regularization parameter, 
hence there is clear connection between \paramthresh and the number of model coefficients.

As a stopping criterion, \paramthresh influences the duration of model training
(larger values of \paramthresh result into faster training), 
and the quality of the model (larger values of \paramthresh may decrease quality).

\begin{description}
\item[Type:] double, scalar.
\item[Sensitivity / robustness:] 
As long as \paramthresh is in a reasonable range of values, 
the model will not be sensitive to changes.
Extremely large values can lead to fairly poor models,
extremely small values may result into significantly larger training times.
\item[Heuristics for determination:] none are known.
\item[Plausible range of values:]  
\paramthresh $\approx 0$, \paramthresh $>0$.
It seems reasonable to set \paramthresh on a log-scale with fairly coarse granularity,
since \paramthresh has a low sensitivity for the most part. Example: \paramthresh $= 10^{-20},10^{-18}, ... , 10^{-4}$.
\item[Further constraints:] none.
\item[Interactions:] none are known.
\end{description}

In conclusion, Table~\ref{table:survey:en} provides a brief survey of
examples from the literature, where \ac{en} was tuned.
\begin{table}[!htbp]
\caption{Survey of examples from the literature, for tuning of \ac{en}.\\
*: Results depend on data set (multiple data sets).\\
**: the integrated, automatic tuning procedure from \modglmnet was used.
}
\begin{tabular}{c | c  | c | c | c}
\hline
Hyperparameter & Lower bound & Upper bound & Result & Notes  \\
\hline
\multicolumn{5}{l}{\citep{Prob19a}, various applications, 38 data sets}\\
\hline
\paramalpha&0&1& 0,003 to 0,981 *& \\
\paramlambda&$2^{-10}$&$2^{10}$& 0,001 to 0,223 *& \\
\hline
\multicolumn{5}{l}{\citep{Wong19a}, medical data, 1 data set }\\
\hline
\paramalpha&1&1&1& not tuned, constant \\
\paramlambda&**&**&0.001& \\
\hline
\end{tabular}
\label{table:survey:en}
\end{table}

\subsection{Model: Decision Trees}
\label{sec:eb}
Decision and regression trees are models that divide the data space into individual 
segments with successive decisions (called splits).

Basically, the procedure of a decision tree is as follows: 
Starting from a root node (which contains all observations) a first split is carried out. 
Each split affects a variable (or a feature). 
This variable is compared with a threshold value. 
All observations that are less than the threshold are assigned to a new node. 
All other observations are assigned to another new node. 
This procedure is then repeated for each node until a termination criterion is reached or until there is only one observation in each end node. 

End nodes are also called leaves (following the tree analogy). 

A detailed description of tree-based models is given by~\cite{Jame14a}. 
An overview of decision tree  implementations and algorithms is given by~\cite{Zhar19a}.
\citet{Mant18a} describe the tuning of hyperparameters of several implementations.
As a reference implementation, we refer to the R package \Rrpart~\citep{Ther19a, Ther19b}.

\subsubsection{\acl{eb} Parameter: \paramminsplit}
If there are fewer than \paramminsplit observations in a node of the tree, no further split is carried out at this node.
Thus, \paramminsplit limits the complexity (number of nodes) of the tree. 
With large \paramminsplit values, fewer splits are made. 
A suitable choice of \paramminsplit can thus avoid overfitting. 
In addition, the parameter influences the duration of the training of an decision tree~\citep{Hast17a}.
\begin{description}
\item[Type:] integer, scalar.
\item[Sensitivity / robustness:]
Trees can react very sensitively to parameters that influence their complexity. 
Together with \paramminbucket, \paramcp and \parammaxdepth, \paramminsplit is one of the most important hyperparameters~\citep{Mant18a}.
\item[Heuristics for determination:]
\paramminsplit is set to three times \paramminbucket in certain implementations, if this parameter is available~\citep{Ther19a}.
\item[Plausible range of values:]
\paramminsplit $ \in [1, m]$, 
where \paramminsplit $ <<  m$  is recommended, since otherwise trees with extremely few nodes will arise. Only integer values are valid.
\item[Further constraints:]
\paramminsplit $>$ \paramminbucket. This is a soft constraint, i.e., valid models are created even if violated, but \paramminsplit would no longer have any effect.
\item[Interactions:]
The parameters \paramminsplit, \paramminbucket, \paramcp and \parammaxdepth all influence the complexity of the tree. 
Interactions between these parameters are therefore likely. In addition, 
\paramminsplit has no effect for certain values of \paramminbucket (see Constraints). 
Similar relationships (depending on the data) are also conceivable for the other parameter combinations.
\end{description}

\subsubsection{\acl{eb} Parameter: \paramminbucket}
\paramminbucket specifies the minimum number of data points in an end node (leaf) of the tree.
The meaning in practice is similar to that of \paramminsplit. 
With larger values, \paramminbucket also increasingly limits the number of splits and thus the complexity of the tree.
\begin{description}
\item[Type:] integer, scalar.
\item[Sensitivity / robustness:] see \paramminsplit.
\item[Heuristics for determination:] \paramminbucket is set to a third of \paramminsplit in the reference implementations, if this parameter is available 
~\cite{Ther19b}.
\item[Plausible range of values:]
\paramminbucket $\in [1, m]$, where \paramminbucket $ << m$  is recommended, as otherwise trees with extremely few nodes will arise. Only integer values are valid.
\item[Further constraints:] \paramminsplit $>$ \paramminbucket (this is a soft constraint, i.e., valid models are created even if violated, but \paramminsplit would no longer have any effect).
\item[Interactions:] see \paramminsplit. 
Due to the similarity of \paramminsplit and \paramminbucket, it can make sense to only tune one of the two parameters.
\end{description}

\subsubsection{\acl{eb} Parameter: \paramcp}
The threshold complexity \paramcp controls the complexity of the model in that split decisions are linked to a minimal improvement. 
This means that if a split does not improve the tree-based model by at least the factor \paramcp, this split will not be carried out.
With larger values, \paramcp increasingly limits the number of splits and thus the complexity of the tree.

\citet{Ther19a} describe the \paramcp parameter as follows:
\begin{quote}
The complexity parameter \paramcp is, like \paramminsplit, an advisory parameter, but is considerably more useful. 
It is specified according to the formula 
\begin{equation}
R_{\paramcp}(T) \equiv R(T) + \paramcp \times  |T| \times R(T_1)
\end{equation}
where $T_1$ is the tree with no splits, $|T|$ is the number of splits for a tree, and $R$ is the risk. 
This scaled version is much more user-friendly than the original CART formula  since it is unit less. 
A value of $\paramcp = 1$ will always result in a tree with no splits. 
For regression models the scaled \paramcp has a very direct interpretation: 
if any split does not increase the overall $R^2$ of the model by at least \paramcp (where $R^2$
is the usual linear-models definition) 
then that split is decreed to be, a priori, not worth pursuing. 
The program does not split said branch any further, and saves considerable computational effort. 
The default value of .01 has been reasonably successful at ‘pre-pruning’ 
trees so that the cross-validation step need only remove 1 or 2 layers, 
but it sometimes over prunes, particularly for large data sets.
\end{quote}

\begin{description}
\item[Type:] double, scalar.
\item[Sensitivity / robustness:] see \paramminsplit.
\item[Heuristics for determination:] none known.
\item[Plausible range of values:] paramcp $\in  [0, 1[$.
\item[Further constraints:] none.
\item[Interactions:] see \paramminsplit. 
Since \paramcp expresses a relative factor for the improvement of the model, 
an interaction with the corresponding quality measure is also possible (split parameter).
\end{description}

\subsubsection{\acl{eb} Parameter: \parammaxdepth}
The parameter \parammaxdepth limits the maximum depth of a leaf in the decision tree. 
The depth of a leaf is the number of nodes that lie on the path between the root and the leaf. 
The root node itself is not counted~\citep{Ther19a}.

The meaning in practice is similar to that of \paramminsplit. 
Both \paramminsplit and \parammaxdepth can be used to limit the complexity of the tree. 
However, smaller values of \parammaxdepth lead to a lower complexity of the tree.
With \paramminsplit it is the other way round (larger values lead to less complexity).

\begin{description}
\item[Type:] integer, scalar.
\item[Sensitivity / robustness:] see \paramminsplit
\item[Heuristics for determination:] none known.
\item[Plausible range of values:] \parammaxdepth $\in [0, m]$. 
Only integer values are valid.
\item[Further constraints:] none.
\item[Interactions:] see \paramminsplit.
\end{description}

Table~\ref{table:survey:eb} shows examples from the literature.
\begin{table}
\caption{\acl{eb}: survey of examples from the literature. Tree-based tuning example configurations. $*$ denotes that results depend on data sets.
}
\begin{tabular}{l | r  | r | r }
\hline
Hyperparameter & Lower bound & Upper bound & Result   \\
\hline
\multicolumn{4}{l}{\citep{Prob19a}, various applications, 38 data sets}\\
\hline
\paramminsplit&1&60& 6.7 to 49.15 * \\
\parammaxdepth&1&30& 9 to 28 * \\
\paramcp&0&1&0 to 0.528 * \\
\paramminbucket&1&60& 1 to 44.1 * \\
\hline
\multicolumn{4}{l}{\citep{Wong19a}, medical data, 1 data set}\\
\hline
\paramcp&$10^{-6}$&$10^{-1}$&$10^{-2}$  \\
\hline
\multicolumn{4}{l}{\citep{Khan20a}, software bug detection, 5 data sets}\\
\hline
\paramminbucket & 1 & 50 & NA \\
\hline
\multicolumn{4}{l}{\citep{Mant18a}, various data sets, 94 data sets}\\
\hline
\paramminsplit&1&50&  \\
\paramminbucket&1&50& \\
\paramcp&0.0001&0.1& \\
\parammaxdepth&1&50& \\
\hline
\multicolumn{4}{l}{This study, see Sec.~\ref{sec:case1}, \gls{CID}}\\
\hline
\paramminsplit & 1 & 300 &  $16$ (not relevant)\\
\paramminbucket& 0.1& 0.5 & $0.17$ (not relevant)\\
\paramcp       & $10^{-10}$ & 1  & $10^{-3}$ (most relevant hyperparameter) \\
\parammaxdepth & 1          & 30 &  $>10$ \\
\hline
\end{tabular}
\label{table:survey:eb}
\end{table}

\subsection{Model: \acl{rf}}
The model quality of decision trees can often be improved with ensemble methods.
Here, many individual models (i.e., many individual trees) are merged into one overall model (the ensemble).

An example is the \ac{rf} model.
In this process, many decision trees are created at the same time, and their prediction is then usually made using the mean (in case of regression) or by majority vote (in case of classification).
The variant of \ac{rf} described by~\citet{Brei01a} uses two important steps to reduce generalization error:
First, when creating individual trees, only a random subset of the features is considered for each split.
Second, each tree is given a randomly drawn subset of the observations to train. 
Typically, the approach of bootstrap aggregating or bagging~\citep{Jame17a} is used.
A comprehensive discussion of random forest models is provided by~\citet{Loup15a} and a detailed discussion of hyperparameters is provided by~\citet{Loup15a}.
Theoretical results on hyperparameters of \ac{rf} models are summarized by~\citet{Scor17a}.
Often, tuning of \ac{rf} also takes into account parameters for the decision trees themselves see  Sec.~~\ref{sec:eb}.
Our reference implementation studied in this report is from the R package \modranger\footnote{\url{https://cran.r-project.org/package=ranger}}~\citep{Wrig20a,Wrig17a}.

\subsubsection{\acl{rf} Parameter:  \paramnumtrees}\label{sub:rf:numtrees}
\paramnumtrees determines the number of trees that are combined in the overall ensemble model.
In practice, this influences the quality of the model (more trees lead to a better quality) 
and the runtime of the model (more trees lead to longer runtimes for training and prediction). 

\begin{description}
\item[Type:] integer, scalar.
\item[Sensitivity / robustness:] 
According to \citet{Brei01a}, the generalization error of the model converges with increasing number of trees towards a lower bound. 
This means, that the model will become less sensitive to changes of 
\paramnumtrees with increasing values of \paramnumtrees.
This is also shown in the benchmarks of~\citet{Loup15a}.
Only with relatively small values (\paramnumtrees $<50$) the model is rather sensitive to changes in that parameter.
The empirical results of~\citet{Prob19a} also show that the \textit{tunability} of \paramnumtrees is estimated to be rather low. 
\item[Heuristics for determination:] 
There are theoretical results about the convergence of the model in relation
to \paramnumtrees~\citep{Brei01a,Scor17a}. 
This however does not result into a clear heuristic approach to setting this parameter.
One common recommendation is to choose \paramnumtrees sufficiently high~\citep{Prob19b} (since more trees are usually better),
while making sure that the runtime of the model does not become too large.
\item[Plausible range of values:] \paramnumtrees $\in [1,\infty)$.  
Several hundred or thousands of trees are commonly used, 
see also Table~\ref{table:survey:rf}.
\item[Further constraints:] none.
\item[Interactions:] none are known.
\end{description}
\subsubsection{\acl{rf} Parameter: \parammtry}\label{sub:rf:mtry}
\parammtry determines, how many randomly chosen features are considered for each split. 
Thus, it controls an important aspect, the randomization of individual trees.
Values of \parammtry $\ll n$ imply that differences between trees will be larger (more randomness).
This increases the potential error of individual trees, but the overall ensemble benefits~\citep{Brei01a,Loup15a}.
One useful side effect is, that \parammtry $\ll n$ may also reduce the runtime considerably~\citep{Loup15a}.
Nevertheless, findings about this parameter largely depend on heuristics and empirical results.
According to Scornet, no theoretical results about the randomization of split features are available~\citep{Scor17a}.

\begin{description}
\item[Type:] integer, scalar.
\item[Sensitivity / robustness:] 
According to Breiman, \ac{rf} is relatively insensitive to changes of \parammtry:
``\textit{But the procedure is not overly
sensitive to the value of F. The average absolute difference between the error
rate using F=1 and the higher value of F is less than 1.}''~\citep{Brei01a} (here: F corresponds to \parammtry).

This seems to be at odds with the benchmarks by~\citet{Loup15a}, 
which determine that \parammtry may indeed have a considerable impact, especially for low values of \parammtry.
The investigation of tunability by~\citet{Prob19a} also identifies \parammtry as an important (i.e., tunable) parameter.
This is not necessarily a contradiction to Breiman's observation,
since~\citet{Prob19a} determine \ac{rf} as the least tunable model in their 
experimental investigation. So while \parammtry might have some impact (compared to other parameters),
it may be less sensitive when compared relative to hyperparameters of other models.

\item[Heuristics for determination:]  
\citet{Brei01a} propose the following heuristic: 
\begin{equation*}
\text{\parammtry} =\text{floor}(\log_2(n)+1).
\end{equation*}
Should categorical features be present, Breiman suggests doubling or tripling that value. 
No theoretical motivation is given.

Another frequent suggestion is \parammtry $=\sqrt{n}$ (or \parammtry $=\text{floor}(\sqrt{n})$).
While these are used in various implementations of \ac{rf}, there is no clear theoretical motivation given.
For $n<20$ both heuristics provide very similar values.

Some implementations distinguish between classification (\parammtry $=\sqrt{n}$)
and regression (\parammtry $=n/3$). 
Empirical results with these heuristics are described by~\citet{Prob19b}.

\item[Plausible range of values:] \parammtry $\in [1,n]$.
\item[Further constraints:] none.
\item[Interactions:] none are known.
\end{description}
\subsubsection{\acl{rf} Parameter: \paramsamplefraction}
The parameter \paramsamplefraction determines, how many
observations are randomly drawn for training a specific tree. 

\citet{Prob19b} write, that \paramsamplefraction a similar effect as \parammtry.
That means, it influences the properties of the trees:
With small \paramsamplefraction (corresponding to small \parammtry) individual trees
are weaker (in terms of predictive quality), yet the diversity of trees is increased.
This improves the ensemble model quality.
Smaller values of \paramsamplefraction reduce the runtime~\citep{Prob19b} (if all other parameters are equal).

\begin{description}
\item[Type:] double, scalar.
\item[Sensitivity / robustness:] 
\paramsamplefraction can have a relevant impact on model quality.
Scornet reports: 
``\textit{However, according
to empirical results, there is no justification for default values in random forests for sub-sampling or
tree depth, since optimizing either leads to better performance.}''
\item[Heuristics for determination:] none known.
\item[Plausible range of values:] \paramsamplefraction $\in (0,1]$.
\item[Further constraints:] none.
\item[Interactions:] Potentially, \paramsamplefraction interacts with parameters that influence training individual trees \ac{eb}
(e.g., \parammaxdepth, \paramminsplit, \paramcp).
Scornet:
``\textit{According to the theoretical analysis of median forests, we know that there is no need to optimize
both the \paramsubsample size and the tree depth: optimizing only one of these two parameters leads to the
same performance as optimizing both of them}''~\citep{Scor17a}.
However, this theoretical observation is only valid for the respective \textit{median trees}
and not necessarily for the classical \ac{rf} model we consider
\end{description}
\subsubsection{\acl{rf} Parameter: \paramreplace}
The parameter \paramreplace specifies,
whether randomly drawn samples are replaced, i.e.,
whether individual samples can be drawn multiple times for
training of a tree (\paramreplace=\true) or not (\paramreplace=\false). 
If \paramreplace=\true, the probability that two trees
receive the same data sample is reduced.
This may further decorrelate trees and improve quality.
\begin{description}
\item[Type:] logical, scalar.
\item[Sensitivity / robustness:] 
The sensitivity of \paramreplace is often rather small.
Yet, the survey of~\citet{Prob19b} notes 
a potentially detrimental bias for \paramreplace=\true,
if categorical variables with a variable number of levels 
are present.
\item[Heuristics for determination:] 
Due to the aforementioned bias, 
the choice could be made depending on the variance
of the cardinality in the data features.
However, a quantifiable recommendation is not available.
\item[Plausible range of values:] \paramreplace $\in\{\true, \false\}$.
\item[Further constraints:] none.
\item[Interactions:] 
One obvious interaction occurs with \paramsamplefraction.
Both parameters control the random choice of training data for each tree.
The setting (\paramreplace = \true $\land$ \paramsamplefraction $=1$) as well as the setting (\paramreplace = \false $\land$ \paramsamplefraction $<1$)
implies that individual trees will not see the whole data set.
\end{description}

\subsubsection{\acl{rf} Parameter: \paramrespect}\label{sec:paramrespect}
This parameter decides how splits of categorical variables are
handled.
A detailed discussion is given by~\citet{Wrig19a}.
A standard that is also used by~\citep{Brei01a}, is \paramrespect=\texttt{partition}. 
In that case, all potential splits of a nominal, categorical variable are considered.
This leads to a good model, but the large number of considered splits can lead to an
unfavorable runtime.

A naive alternative is \paramrespect=\texttt{ignore}.
Here, the categorical nature of a variable will be ignored.
Instead, it is assumed that the variables is ordinal,
and splits are chosen just as with numerical variables.
This reduces runtime but can decrease model quality.

A better choice should be \paramrespect=\texttt{order}.
Here, each categorical variable first is sorted, depending on
the frequency of each level in the first of two classes (in case of classification
or depending on the average dependent variable value (regression).
After this sorting, the variable is considered to be numerical.
This allows for a runtime similar to that with \paramrespect=\texttt{ignore} 
but with potentially better model quality.
This may not be feasible for classification with more than two classes, 
due to lacking a clear sorting criterion~\citep{Wrig19a,Wrig20a}.

In specific cases, \paramrespect=\texttt{ignore} may work well in practice.
This could be the case, when the variable is actually nominal (unknown to the analyst).

\begin{description}
\item[Type:] character, scalar.
\item[Sensitivity / robustness:] unknown.
\item[Heuristics for determination:]  none.
\item[Plausible range of values:] \paramrespect $\in\{$\texttt{ignore}, \texttt{order},\\ \texttt{partition}$\}$.
The parameter \paramrespect can also be understood as a binary value.
Then \true corresponds to \texttt{order} and \false to \texttt{ignore}~\citep{Wrig20a}.
\item[Further constraints:] none.
\item[Interactions:] none are known.
\end{description}

In conclusion, Table~\ref{table:survey:rf} provides a brief survey of
examples from the literature, where \ac{rf} was tuned.
\begin{table}[!htbp]
\caption{\ac{rf}: survey of examples from the literature for tuning of random forest. *: Results depend on data set (multiple data sets).
}
\begin{tabular}{l | r  | r | r | l}
\hline
Hyperparameter & Lower bound & Upper bound & Result & Notes  \\
\hline
\multicolumn{5}{l}{\citep{Prob19a}, various applications, 38 data sets}\\
\hline
\paramnumtrees&1&2000& 187,85 to 1908,25 *& \\
\paramreplace& & & \false & binary\\
\paramsamplefraction&0.1 &1& 0,257 to 0,974 *& \\
\parammtry&0&1&0,035 to 0,954 *& transformed:  \parammtry$\times m$\\
\paramrespect&&&\false oder \true & binary \\
min.node.size&0&1& 0,007 to 0,513 *& transformed:  $n^\text{min.node.size}$\\
\hline
\multicolumn{5}{l}{\citep{Schr19a}, spatial data, 1 data set}\\
\hline
\paramnumtrees&10&10000& NA & \\
\parammtry&1&11& NA &\\
\hline
\multicolumn{5}{l}{\citep{Wong19a}, medical data, 1 data set}\\
\hline
\paramnumtrees&10&2000& 1000 & \\
\parammtry&10&200& 50 &\\
\hline
\end{tabular}
\label{table:survey:rf}
\end{table}

\subsection{Model: \acl{xgb}}\label{sec:xgboost}
Boosting is an ensemble process. 
In contrast to random forests, 
the individual models (here: decision trees) are not created and evaluated at the same time, but rather sequentially. 
The basic idea is that each subsequent model tries to compensate for the weaknesses of the previous models.

For this purpose, a model is created repeatedly. The model is trained with weighted data. 
At the beginning these weights are identically distributed. 
Data that are poorly predicted or recognized by the model are given larger 
weights in the next step and thus have a greater influence on the next model. 
All models generated in this way are combined as a linear combination to form an overall model~\citep{Freu97a, Druc95a}.

An intuitive description of this approach is \emph{slow learning}, as the attempt is not made to understand the entire database in a single step, but to improve the understanding step by step~\citep{Jame14a}.
\gls{GB} is a variant of this approach, with one crucial difference: instead of changing the weighting of the data, models are created sequentially that follow the gradient of a loss function. 
In the case of regression, the models learn with residuals of the sum of all previous models. Each individual model tries to reduce the weaknesses (here: residuals) of the ensemble~\citep{Frie01a}. 

In the following we consider the hyperparameters of one version of \gls{GB}: \gls{XGBoost}~\citep{Chen16a}.
In principle, any models can be connected in ensembles via boosting. 
We apply \gls{XGBoost} to decision trees.
As a reference implementation, we refer to the R package xgboost~\citep{Chen20a}.
\citet{Brow18a} describes some empirical hyperparameter values for tuning  \gls{XGBoost}.

\subsubsection{\acl{xgb} Parameter: \parameta}
The parameter \parameta is a learning rate and is also called \enquote{shrinkage} parameter. 
The parameter controls the lowering of the weights in each boosting step~\citep{Chen16a, Frie02a}.
It has the following practical meaning: lowering the weights helps to reduce the influence of individual trees on the ensemble. This can also avoid overfitting~\citep{Chen16a}.
\begin{description}
\item[Type:] double, scalar. 
\item[Sensitivity / robustness:]  Empirical results show that \gls{XGBoost} is more sensitive to \parameta when \parameta is large~\citep{Frie01a}. 
Generally speaking, smaller values are better. In an empirical study, \citet{Prob18a} describe \parameta as a parameter with comparatively high tunability.
\item[Heuristics for determination:] A heuristic is difficult to formulate due to the dependence on other parameters and the data situation, but ~\cite{Hast17a} recommend
\begin{quote}
$\ldots$ the best strategy appears to be to set \parameta to be very small (\parameta $<0.1$) and then choose \paramnrounds by early stopping.
\end{quote}
This may lead to correspondingly longer runtimes due to large \paramnrounds. 
\citet{Brow18a} mentions a heuristic, which describes a search range depending on \paramnrounds.
\item[Plausible range of values:] \parameta $\in  [0, 1]$.
Using a logarithmic scale seems reasonable, e.g., $2^{-10}, \ldots, 2^0$), 
as used in the studies by~\cite{Prob18a} or ~\cite{Sigr20a}, because values close to zero often show good results.
\item[Further constraints:] none.
\item[Interactions:] 
There is a connection between \parameta and \paramnrounds: 
If one of the two parameters increases, the other should be decreased if the error remains the same~\citep{Frie01a}, \citep{Prob19a}. This is also demonstrated by ~\citet{Hast17a}: 
\begin{quote}
Smaller values of \parameta lead to larger values of \paramnrounds for the same training risk, so that there is a trade-off between them. 
\end{quote}
In addition, ~\citet{Hast17a} also point to correlations with the \paramsubsample parameter: 
In an empirical study, \paramsubsample = 1 and \parameta = 1 show significantly worse results than \paramsubsample = 0.5 and \parameta = 0.1. 
If \paramsubsample = 0.5 and \parameta = 1, the results are even worse than for \parameta = 1 and \paramsubsample = 1. 
In the best case (\paramsubsample = 0.5 and \parameta = 0.1), however, larger values of \paramnrounds are required to achieve optimal results.
\end{description}

\subsubsection{\acl{xgb} Parameter: \paramnrounds}
The parameter \paramnrounds specifies the number of boosting steps. 
Since a tree is created in each individual boosting step, \paramnrounds also controls the number of trees that are integrated into the ensemble as a whole.
Its practical meaning can be described as follows: larger values of \paramnrounds mean a more complex and possibly more precise model, but also cause a longer running time.
The practical meaning is therefore very similar to that of num.trees in random forests. In contrast to num.trees, overfitting is a risk with very large values, 
depending on other parameters such as \parameta, \paramlambda, \paramalpha. For example, the empirical results of~\citet{Frie01a} show that with a low \parameta, even a high value of \paramnrounds does not lead to overfitting.
\begin{description}
\item[Type:] integer, scalar.
\item[Sensitivity / robustness] Similar to the random forests parameter num.trees, \paramnrounds also has a higher sensitivity, especially with low values~\citep{Frie01a}.
\item[Heuristics for determination:] Heuristics cannot be derived from the literature. Often values of several hundred to several thousand trees are set as the upper limit~\citep{Brow18a}.
\item[Plausible range of values:]   $\in  [1, \infty [$. Only integer values are valid.
\item[Further constraints:] none.
\item[Interactions:] There is a connection between the hyperparameters beta, rounds, and \paramsubsample.
\end{description}

\subsubsection{\acl{xgb} Parameter: \paramlambda}
The parameter \paramlambda is used for the regularization of the model. 
This parameter influences the complexity of the model~\citep{Chen16a, Chen20a} (similar to the parameter of the same name in elastic net).
Its practical significance can be described as follows: as a regularization parameter, \paramlambda helps to prevent overfitting~\citep{Chen16a}. With larger values, smoother or simpler models are to be expected.
\begin{description}
\item[Type:] double, scalar.
\item[Sensitivity / robustness:] not known.
\item[Heuristics for determination:] none known.
\item[Plausible range of values:] \paramlambda $\in [0, \infty[$.
A logarithmic scale seems to be useful, e.g., $ 2^{-10}, \dots, 2^{10}$, as used in the study by~\citet{Prob19a} to cover a wide range of very small and very large values.
\item[Further constraints:] none.
\item[Interactions:] Because both \paramlambda and \paramalpha control the regularization of the model, an interaction is likely.
\end{description}

\subsubsection{\acl{xgb} Parameter: \paramalpha}
\citet{Chen16a} did not mention this parameter.
The documentation of the reference implementation does not provide any detailed information on \paramalpha either. 
Due to the description as a parameter for the L1 regularization of the weights~\citep{Chen20a}, a highly similar use as for the parameter of the same name in elastic net  is to be assumed.
Its practical meaning can be described as follows: similar to \paramlambda, \paramalpha also functions as a regularization parameter.
\begin{description}
\item[Type:] double, scalar.
\item[Sensitivity / robustness:] unknown.
\item[Heuristics for determination:] No heuristics are known.
\item[Plausible range of values:] \paramalpha $\in [0, \infty[$. 
A logarithmic scale seems to be useful, e.g., $ 2^{-10}, \ldots, 2^{10}$, as used in the study by~\citet{Prob19a} to cover a wide range of very small and very large values.
\item[Further constraints:] none.
\item[Interactions:] Since both \paramlambda and \paramalpha control the regularization of the model, an interaction is likely.
\end{description}

\subsubsection{\acl{xgb} Parameter: \paramsubsample}
In each boosting step, the new tree to be created is
usually only trained on a subset of the entire data set, similar to random forest~\citep{Frie02a}.
The \paramsubsample parameter specifies the portion of the data approach that is randomly selected in each iteration.
Its practical significance can be described as follows: an obvious effect of small \paramsubsample values is a shorter running time for the training of individual trees, which is proportional to the \paramsubsample~\citep{Hast17a}.
\begin{description}
\item[Type:] double, scalar.
\item[Sensitivity / robustness:]  The study by~\citet{Frie02a} shows a high sensitivity for very small or large values of \paramsubsample. 
In a relatively large range of values from \paramsubsample (around 0.3 to 0.6), however, hardly any differences in model quality are observed.
\item[Determination heuristics:] ~\citet{Hast17a} suggest \paramsubsample = 0.5 as a good starting value, 
but point out that this value can be reduced if \paramnrounds increases. 
With many trees (nround is large) it is sufficient if each individual tree sees a smaller part of the data, since the unseen data is more likely to be taken into account in other trees.
\item[Plausible range of values:] \paramsubsample $\in ]0,1]$.
Based on the empirical results~\cite{Frie02a, Hast17a}, a logarithmic scale is not recommended.
\item[Further constraints:] none.
\item[Interactions:] There is a connection between the \parameta, \paramnrounds, and \paramsubsample.
\end{description}

\subsubsection{\acl{xgb} Parameter: \paramcolsample}
The parameter \paramcolsample 
has similarities to the mtry parameter in random forests. 
Here, too, a random number of features is chosen for the splits of a tree. 
In \gls{XGBoost}, however, this choice is made only once for each tree that is created, 
instead of for each split~\citep{xgbo20a}.
Here \paramcolsample is a relative factor. The number of selected features is therefore \paramcolsample $ \times n$.
Its practical meaning is similar to  mtry: \paramcolsample enables the trees of the ensemble to have a greater diversity.
The runtime is also reduced, since a smaller number of splits have to be checked each time (if \paramcolsample $< 1$).
\begin{description}
\item[Type:] double, scalar.
\item[Sensitivity / robustness:]
The empirical study by~\citet{Prob19a} shows that the model is particularly sensitive to changes in the area of \paramcolsample = 1. 
However, this sensitivity decreases in the vicinity of more suitable values.
\item[Heuristics for determination:] none known.
\item[Plausible range of values:] \paramcolsample $\in ]0,1]$.
\citet{Brow18a} mentions search ranges such as \paramcolsample = $0.4, 0.6, 0.8, 1$, 
but mostly works with \paramcolsample = $0.1, 0.2, \ldots, 1$.
\item[Further constraints:] none.
\item[Interactions:] none known.
\end{description}

\subsubsection{\acl{xgb} Parameter: \paramgamma}
This parameter of a single decision tree  is very similar to the parameter \paramcp: 
Like \paramcp, \paramgamma controls the number of splits of a tree by assuming a minimal improvement for each split. 
According to the documentation~\citep{Chen20a}:
\begin{quote}
Minimum loss reduction required to make a further partition on a leaf node of the tree. the larger, the more conservative the algorithm will be.
\end{quote}
The main difference between \paramcp and \paramgamma is the definition of \paramcp as a relative factor, 
while \paramgamma is defined as an absolute value. This also means that the value range differs.
\begin{description}
\item[Plausible range of values:] \paramgamma $ \in [0, \infty[$.
A logarithmic scale seems to make sense, e.g., $2^{-10}, \ldots, 2^{10}$, as, e.g., in the study 
by~\citet{Thom18a} to cover a wide range of very small and very large values.
\end{description}

\subsubsection{\acl{xgb} Parameter: \parammaxdepth}
This parameter of a single decision tree is already known as \parammaxdepth.
\begin{description}
\item[Sensitivity / heuristics:] \citet{Hast17a} state:
\begin{quote}
Although in many applications $J = 2$ will be insufficient, 
it is unlikely that $J > 10$ will be required. 
Experience so far indicates that $ 4 \leq J \leq 8$ 
works well in the context of boosting, with results being fairly insensitive to particular choices in this 
range.\footnote{$J$ is the number of nodes in a tree that is strongly influenced by \parammaxdepth.} 
\end{quote}
\end{description}

\subsubsection{\acl{xgb} Parameter: \paramminchild}
Like \paramgamma and \parammaxdepth, \paramminchild restricts the number of splits of each tree. 
In the case of \paramminchild, this restriction is determined using the Hessian matrix of the loss function (summed over all observations in each new terminal node)~\citep{Chen20a, Sigr20a}.
In experiments by ~\citet{Sigr20a}, this parameter turns out to be comparatively difficult to tune: 
The results show that tuning with \paramminchild gives worse results than tuning with a similar parameter (limitation of the number of samples per sheet)~\citep{Sigr20a}.
\begin{description}
\item[Type:] double, scalar.
\item[Sensitivity / robustness:] unknown.
\item[Heuristics for determination:] none known.
\item[Plausible range of values:] \paramminchild $\in [0, \infty [$.
A logarithmic scale seems to make sense, e.g., $2^{-10}, \ldots, 2^{10}$, as used in the study 
by~\citet{Prob19a} to cover a wide range of very small and very large values.
\item[Further constraints:] none.
\item[Interactions:] Interactions with parameters such as \paramgamma and \parammaxdepth are probable, since all three parameters influence the complexity of the individual trees in the ensemble.
\end{description}
Table~\ref{table:survey:xgbparam} shows example parameter settings from the literature.

\begin{table}
\caption{Survey: examples from literature about \gls{XGBoost} tuning. $*$ denotes that results depend on the data (several data sets).
}
\begin{tabular}{l | r  | r | r | l}
\hline
Hyperparameter & Lower bound & Upper bound & Result & Comments  \\
\hline
\multicolumn{5}{l}{\citep{Prob19a}, several applications, 38 data sets}\\
\hline
\paramnrounds & 1 & 5000 &  920.7 to 4847.15 *& \\
\parameta & $2^{-10}$ & $2^0$ &  0.002 to 0.445 *& \\
\paramsubsample & 0.1 & 1 & 0.545 to 0.964 *& \\
\parammaxdepthx &1 & 15 & 2.6 to 14 *& \\
\paramminchild &$2^{0}$  &$2^{7}$  & 1.061 to 7.502 *& \\
\paramcolsample &0 &1 & 0.334 to 0.922 *& \\
\paramlambda &$2^{-10}$  &$2^{10}$  & 0.004 to 29.755 *& \\
\paramalpha &$2^{-10}$  &$2^{10}$  & 0.002 to 6.105 *& \\
\hline
\multicolumn{5}{l}{\citep{Thom18a}, several applications, 16 data sets}\\
\hline
\parameta & 0.01 & 0.2 & & \\
\paramgamma & $2^{-7}$ & $2^6$ & & \\
\paramsubsample & 0.5 & 1 && \\
\parammaxdepthx &3& 20 & & \\
\paramcolsample &0.5 &1 & & \\
\paramlambda &$2^{-10}$  &$2^{10}$  & & \\
\paramalpha &$2^{-10}$  &$2^{10}$  & & \\
\hline
\multicolumn{5}{l}{\citep{Wang19a}, Risk Classification, 1 data set}\\
\hline
\parameta & 0.005 & 0.2 & & \\
\paramsubsample & 0.8 & 1 & & \\
\parammaxdepthx &5 & 30 & & \\
\paramminchild &0  &10  & & \\
\paramcolsample & 0.8 & 1 & & \\
\paramgamma & 0 & 0.02 && \\
\hline
\multicolumn{5}{l}{\citep{Zhou20a}, Tunnel construction, 1 data set}\\
\hline
\paramnrounds &1 & 150 &  103& \\
\parameta & 0.00001 & 1 &0.152 & \\
\parammaxdepthx &1 & 15 &15 & \\
\paramlambda &1  &15  & 13& \\
\paramalpha &1 &15 &1 & \\
\hline
\multicolumn{5}{l}{This study, see Sec.~\ref{sec:case2}, \gls{CID}}\\
\hline
\paramnrounds & 0 & 32 &  256 & \\
\parameta & $2^{-10}$ & 0 & 0.125 & \\
\hline
\end{tabular}
\label{table:survey:xgbparam}
\end{table}

\subsection{Model: \acl{svm}}

The \ac{svm} is a kernel-based model.
A kernel is a real-valued, symmetrical function $\text{k}(\xtrain,\xtrain')$
(usually positive-definite),
which often expresses some form of similarity between two observations $\xtrain, \xtrain'$.
The usefulness of kernels can be explained by the Kernel-Trick. 
The Kernel-Trick describes the ability of kernels, to transfer data into a higher-dimensional 
feature space.
This allows, e.g., classification with linear decision boundaries (hyperplanes)
even in cases where the data in the original features space are not linearly separable~\citep{Scho01a}.

As reference implementation, we use the R package \modsvm\footnote{\url{https://cran.r-project.org/package=e1071}}~\citep{Meye20a}. 
Based on libsvm~\citep{Chan19a}.

\subsubsection{\acl{svm} Parameter: \paramkernel}
The parameter \paramkernel is central for the \ac{svm} model.
It describes the choice of the function $\text{k}(\xtrain,\xtrain')$.
In practice, $\text{k}(\xtrain,\xtrain')$ can often be understood to be a measure of similarity. 
That is, the kernel function describes how similar to observations are to each other,
depending on their feature values.
In that context, the kernel function can express important properties about
the relation between observations, such as differentiability.

\begin{description}
\item[Type:] character, scalar.
\item[Sensitivity / robustness:]
The empirical investigation of~\citet{Prob19a} shows:
``\textit{In svm the biggest gain in performance can be achieved by tuning the kernel, \paramgamma or
\paramdegree, while the cost parameter does not seem to be very tunable.}''
This does not necessarily mean that \paramcost should not be tuned,
as the tunability investigated by~\citet{Prob19a} always considers a reference value (e.g., the default).
\item[Heuristics for determination:] 
Informally, it is often recommended to use \paramkernel = \texttt{radial basis}. 
This also matches well to results and observations from the literature~\citep{Prob19a,Guen16a}.
With very large numbers of observations and/or features~\citet{Hsu16a} suggest to use \paramkernel = \texttt{linear}.
These are infallible rules, other kernels may perform better depending on the data set.
This stresses the necessity of using hyperparameter tuning to choose kernels.
\item[Plausible range of values:] 
\begin{itemize}[\textbullet]
\item linear: $\text{k}(\xtrain,\xtrain') =\xtrain^\T \xtrain'$
\item polynomial: $\text{k}(\xtrain,\xtrain') =(\paramgamma~\xtrain^\T \xtrain' + \paramcoefz)^\paramdegree$
\item radial basis: $\text{k}(\xtrain,\xtrain') =\exp(-\paramgamma~|| \xtrain-\xtrain'||^2)$
\item sigmoid: $\text{k}(\xtrain,\xtrain') =\tanh(\paramgamma~\xtrain^\T \xtrain' + \paramcoefz)$
\end{itemize}
\item[Further constraints:] none.
\item[Interactions:]
The kernel functions themselves have parameters (\paramdegree, \paramgamma, and \paramcoefz), 
whose values only matter if the respective function is chosen.
\end{description}

\subsubsection{\acl{svm}  Parameter: \paramdegree}\label{sub:svm:degree}
The parameter \paramdegree influences the kernel function:
\begin{itemize}
\item polynomial: $\text{k}(\xtrain,\xtrain') =(\paramgamma~\xtrain^\T \xtrain' + \paramcoefz)^\paramdegree$
\end{itemize}
Integer values of \paramdegree determine the degree of the polynomial.
Non-integer values are possible, even though not leading to a polynomial in the classical sense.
If \paramdegree has a value close to one, the polynomial kernel approximates the linear kernel.
Else, the kernel becomes correspondingly non-linear.

\begin{description}
\item[Type:] double, scalar.
\item[Sensitivity / robustness:]
The empirical investigation of~\citet{Prob19a} shows:
``\textit{In svm the biggest gain in performance can be achieved by tuning the kernel, \paramgamma or
\paramdegree, while the cost parameter does not seem to be very tunable.}''
\item[Heuristics for determination:]
none are known.
\item[Plausible range of values:] \paramdegree $\in (0,\infty)$.
\item[Further constraints:] none.
\item[Interactions:] 
The parameter only has an impact if \paramkernel=polynomial.
\end{description}

\subsubsection{\acl{svm} Parameter: \paramgamma}
The parameter \paramgamma influences three kernel functions: 
\begin{itemize}[\textbullet]
\item polynomial: $\text{k}(\xtrain,\xtrain') =(\paramgamma~\xtrain^\T \xtrain' + \paramcoefz)^\paramdegree$
\item radial basis: $\text{k}(\xtrain,\xtrain') =\exp(-\paramgamma~|| \xtrain-\xtrain'||^2)$
\item sigmoid: $\text{k}(\xtrain,\xtrain') =\tanh(\paramgamma~\xtrain^\T \xtrain' + \paramcoefz)$
\end{itemize}
In case of polynomial and sigmoid, \paramgamma acts as a multiplier for the scalar product of two feature vectors.
For radial basis, \paramgamma acts as a multiplier for the distance of two feature vectors.

In practice \paramgamma scales how far the impact of a single data sample reaches in terms of influencing the model. 
With small \paramgamma, an individual observation my potentially influence the prediction in a larger vicinity,
since with increasing distance between \xtrain and \xtrain', their similarity will
decrease more slowly (esp. with \paramkernel = radial basis).

\begin{description}
\item[Type:] double, scalar.
\item[Sensitivity / robustness:] 
The empirical investigation of~\citet{Rijn18a} shows that \paramgamma is rather sensitive.
\item[Heuristics for determination:] 
The reference implementation uses a simple heuristic, to determine \paramgamma:
\paramgamma $=1/n$~\citep{Meye20a}.
Another implementation (the \texttt{sigest} function in \texttt{kernlab}\footnote{\url{https://cran.r-project.org/package=kernlab}}) first scales all input data, so that each feature has zero mean and unit variance.
Afterwards, a good interval for \paramgamma is determined, by using the 10\% and 90\% quantile of the distances between the scaled
data samples. By default, 50\% randomly chosen samples from the input data are used.
\item[Plausible range of values:] \paramgamma $\in [0,\infty)$. 
Using a logarithmic scale seems reasonable 
(e.g.: $2^{-10}, ..., 2^{10}$ as used by~\citet{Prob19a}),
to cover a broad spectrum of very small and very large values.
\item[Further constraints:] none.
\item[Interactions:] 
This parameter has no effect when \paramkernel = linear.
In addition, empirical results show a clear interaction with \paramcost~\citep{Rijn18a}.
\end{description}

\subsubsection{\acl{svm} Parameter: \paramcoefz}
The parameter \paramcoefz influences two kernel functions: 
\begin{itemize}[\textbullet]
\item polynomial: $\text{k}(\xtrain,\xtrain') =(\paramgamma~\xtrain^\T \xtrain' + \paramcoefz)^\paramdegree$
\item sigmoid: $\text{k}(\xtrain,\xtrain') =\tanh(\paramgamma~\xtrain^\T \xtrain' + \paramcoefz)$
\end{itemize}
In both cases, \paramcoefz is added to the scalar product of two feature vectors.

\begin{description}
\item[Type:] double, scalar.
\item[Sensitivity / robustness:] 
Empirical results of~\citet{Zhou11a} show that \paramcoefz has a strong impact in case of the polynomial kernel
(but only for \paramdegree=2). 
\item[Heuristics for determination:]
\citet{Guen16a} suggest to leave this parameter at \paramcoefz=0.
\item[Plausible range of values:] \paramcoefz $\in \mathbb{R}$.
\item[Further constraints:] none.
\item[Interactions:] 
This parameter is only active if \paramkernel=polynomial or \paramkernel=sigmoid.
\end{description}

\subsubsection{\acl{svm} Parameter: \paramcost}
The parameter \paramcost (often written as $C$) 
is a constant that weights constraint violations of the model.
$C$ is a typical regularization parameter, which controls the complexity of the model~\citep{Cher04a},
and may help to avoid overfitting or dealing with noisy data.\footnote{
Here, complexity does not mean the number of model coefficients (as in linear models) or splits (decision trees),
but the potential to generate more active / rugged functions. 
In that context, $C$ influences the number of support vectors in the model. 
A high model complexity (many support vectors) can create
functions with many peaks. This may lead to overfitting.}

\begin{description}
\item[Type:] double, scalar.
\item[Sensitivity / robustness:]
The empirical results of~\citet{Rijn18a} show that \paramcost has a strong impact on the model, 
while the investigation of~\citet{Prob19a} determines only a minor tunability.
This disagreement may be explained, since \paramcost may have a huge impact in extreme cases,
yet good parameter values are found close to the default values.
\item[Heuristics for determination:] 
\citet{Cher04a} suggest the following: 
\paramcost $= \max(|\bar{\ytrainv}+3\sigma_\ytrainv|,|\bar{\ytrainv}-3\sigma_\ytrainv|)$.
Here, $\bar{\ytrainv}$ is the mean of the observed \ytrain values in the training data,
and $\sigma_\ytrainv$ is the standard deviation.
They justify this heuristic, by pointing out a connection between \paramcost and the predicted \ytrain:
As a constraint, \paramcost limits the output values of the \ac{svm} model (regression) 
and should hence be set in a similar order of magnitude as the observed \ytrain \citep{Cher04a}.
\item[Plausible range of values:] \paramcost $\in [0,\infty)$.
Using a logarithmic scale seems reasonable 
(e.g.: $2^{-10}, ..., 2^{10}$ as used by~\citet{Prob19a}),
to cover a broad spectrum of very small and very large values.
\item[Further constraints:] none.
\item[Interactions:]
Empirical results show a clear interaction with \paramgamma~\citep{Rijn18a}.
\end{description}

\subsubsection{\acl{svm} Parameter: \paramepsilon}
The parameter \paramepsilon defines a ribbon around predictions.
Residuals within that ribbon are tolerated by the model, i.e., are not penalized~\citep{Scho01a}.
The parameter is only used for regression with \ac{svm}, not for classification. 
In the experiments of this investigation (see  Sec.~~\ref{sec:case3}) 
\paramepsilon is only considered when \ac{svm} is used for regression.

Similarly to \paramcost, \paramepsilon is a regularization parameter
With larger values, \paramepsilon allows for larger errors / residuals.
This reduces the number of support vectors (and incidentally, also the run time).
The model becomes more smooth (cf.~\citep[Fig. 9.4]{Scho01a}).
This can be useful, e.g., to deal with noisy data and avoid overfitting.
However, the model quality may be decreased.

\begin{description}
\item[Type:] double, scalar.
\item[Sensitivity / robustness:]
As described above, \paramepsilon has a significant impact on the model.
\item[Heuristics for determination:] 
For \ac{svm} regression, \citet{Cher04a} suggest based on simplified assumptions
and empirical results:
\paramepsilon $= 3 \sigma \sqrt{\frac{\ln(n)}{n}}$.
Here, $\sigma^2$ is the noise variance,
which has to be estimated from the data, see, e.g., equations (22), (23), and (24) in~\citet{Cher04a}.
The noise variance is the remaining variance of the observations \ytrain, which
can not be explained by an ideal model.
This ideal model has to be approximated, e.g., with the nearest neighbor model~\citep{Cher04a}), resulting into additional computational effort.
\item[Plausible range of values:] \paramepsilon $\in (0,\infty)$. 
\item[Further constraints:]  none.
\item[Interactions:] none are known. 
\end{description}

In conclusion, Table~\ref{table:survey:svm} provides a brief survey of
examples from the literature, where \ac{svm} was tuned.
\begin{table}[!htbp]
\caption{Survey of examples from the literature, for tuning of \ac{svm} 
 (\enquote{*} denotes that results depend on data set (multiple data sets)).
}
\begin{tabular}{l | r  | r | r | l}
\hline
Hyperparameter & Lower bound & Upper bound & Result & Notes  \\
\hline
\multicolumn{5}{l}{\citep{Prob19a}, various applications, 38 data sets}\\
\hline
\paramkernel & & & radial basis & \\ 
\paramcost &$2^{-10} $ &$2^{10}$ &  0,002 to 963,81 *& \\ 
\paramgamma &$2^{-10} $ & $2^{10}$& 0,003 to 276,02 *& \\ 
\paramdegree & 2 &5 &  2 to 4 *& \\ 
\hline
\multicolumn{5}{l}{\citep{Mant15a}, various applications, 70 data sets}\\
\hline
\paramcost &$2^{-2} $ &$2^{15}$ & & \\ 
\paramgamma &$2^{-15} $ & $2^{3}$&& \\ 
\hline
\multicolumn{5}{l}{\citep{Rijn18a}, various applications, 100 data sets}\\
\hline
\paramcost &$2^{-5} $ &$2^{15}$ &  & \\ 
\paramgamma &$2^{-15} $ & $2^{3}$&& \\ 
\paramcoefz  & $-1$ &1 &  & only sigmoid \\ 
\paramtolerance &$10^{-5} $ & $10^{-1}$& & \\ 
\hline
\multicolumn{5}{l}{\citep{Sudh13a}, flow rate prediction (hydrology), 1 data set}\\
\hline
\paramcost &$10^{-5} $ &$10^{5}$ &  1,12 to 1,93 *& \\ 
\paramepsilon &0 & 10& 0,023 to 0,983 * & \\ 
\paramgamma &0& 10& 0,59 to 0,87 * & \\ 
\hline
\end{tabular}
\label{table:survey:svm}
\end{table}

\subsection{Concluding Overview}
On the basis of our literature survey, 
we recommend tuning the hyperparameters.
Reasonable bounds are summarized in Table~\ref{table:hyperExp}.

\begin{table}[!htbp]
\centering
\caption{Overview of hyperparameters in the experiments. 
For data type, we employ the signifiers used in R. 
For categorical parameters, 
we list categories instead of providing bounds.
}
\begin{tabular}{l | l | l |  r | r | r | r | l  }
\hline
Model & Hyperparameter & Data type & Lower & Upper & Lower & Upper  & Transformation \\
 &  &  & bound & bound & bound & bound  &  \\
  &  &  &  &  & (trans.) & (trans.)  &  \\
\hline
\ac{knn}&\paramk & integer & 1 & 30& 1 & 30 & -- \\
&\paramp  & numeric  & 0.1 & 100 & -1 & 2 & $10^\paramp$ \\
\hline
\ac{en}&\paramalpha  & numeric & 0 & 1 & 0 & 1 & --\\
&\paramthresh  & numeric & $10^{-8}$ & $10^{-1}$ & -8 & -1 & $10^\paramthresh$  \\
\hline
\ac{eb}&\paramminsplit  & integer & 1 & 300& 1 & 300 & -- \\
&\paramminbucket  & integer & 1 &150 & 0.1 & 0.5 & $\text{round}(\max(\paramminsplit \times $ \\
&\ &  &  &  &  &  & $\paramminbucket,1))$ \\
&\paramcp  & numeric & $10^{-10}$ & $10^{0}$ & -10 & 0 &  $10^\paramcp$  \\
&\parammaxdepth  & integer & 1 & 30& 1 & 30 & -- \\
\hline
\ac{rf}&\paramnumtrees  & integer& 1 & 2048 & 0 & 11 & $\text{round}(2^\paramnumtrees)$\\
&\parammtry  & integer & 1 & $n$ & 1 & $n$ & --\\
&\paramsamplefraction  & numeric & 0.1 & 1 & 0.1 & 1 & -- \\
&\paramreplace & factor &\multicolumn{4}{c|}{\texttt{TRUE}, \texttt{FALSE} }  & -- \\
&\texttt{respect.-}   &  &  \multicolumn{4}{c|}{\texttt{ } \texttt{ } }  &  \\
&\texttt{unordered.factors}  & factor &  \multicolumn{4}{c|}{\texttt{ignore}, \texttt{order} }  & -- \\
\hline
xgBoost &\parameta  & numeric & $2^{-10}$ & $2^{0}$& -10 & 0 & $2^\parameta$ \\
&\paramnrounds  & integer & 1 & 2048& 0 & 11 &  $2^\paramnrounds$ \\
&\paramlambda  & numeric & $2^{-10}$ & $2^{10}$ & -10 & 10 & $2^\paramlambda$ \\
&\paramalpha  & numeric & $2^{-10}$ & $2^{10}$  & -10 & 10 & $2^\paramalpha$\\
&\paramsubsample & numeric & 0.1 & 1& 0.1 & 1 & -- \\
&\paramcolsample  & numeric  & $1/n$ & 1& $1/n$ & 1 & -- \\
&\paramgamma  & numeric  & $2^{-10}$ & $2^{10}$  & -10 & 10 &  $2^\paramgamma$ \\
&\parammaxdepthx  & integer & 1 & 15  & 1 & 15 &  -- \\
&\paramminchild  & numeric & 1 & 128 & 0 & 7 &  $2^\paramminchild$ \\
\hline
\ac{svm}&\paramkernel & factor & \multicolumn{4}{c|}{\texttt{radial}, \texttt{sigmoid} } & --\\
&\paramgamma  & numeric & $2^{-10}$ & $2^{10}$  & -10 & 10 & $2^\paramgamma$ \\
&\paramcoefz  & numeric & -1 & 1 & -1 & 1 & -- \\
&\paramcost & numeric & $2^{-10}$ & $2^{10}$ & -10 & 10 & $2^\paramcoefz$ \\
&\paramepsilon  & numeric & $10^{-8}$ & $10^{0}$ & -8 & 0 & $10^\paramepsilon$ \\
\end{tabular}
\label{table:hyperExp}
\end{table}

\section{Tuning Procedures}\label{ch:tuning}
This section gives an overview of the various relevant tuning procedures,
which could be applied to the problem of \ac{ml} algorithm tuning.

\subsection{Model-free Search}
\subsubsection{Manual Search}
A frequently applied approach is that \ac{ml} algorithms are configured manually~\citep{Berg12a}.
Users apply their own experience and trial-and-error to find reasonable parameter values.

In individual cases, this approach may indeed yield good results: 
when expert knowledge about data, algorithms and parameters is available.
At the same time, this approach has major weaknesses,
e.g., it may require significant amount of work time by the users,
bias may be introduced due to wrong assumptions,
limited options for parallel computation,
and extremely limited reproducibility.
An automated approach is hence of interest.

\subsubsection{Undirected Search}
Undirected search algorithms determine new hyperparameter values independently of
any results of their evaluation. Two important examples are Grid Search and Random search.

Grid Search covers the search space with a regular grid. Each grid point is evaluated.
Random Search selects new values at random (usually independently, uniform distributed)
in the search space.

Grid Search is a frequently used approach, as it is easy to understand and implement (including parallelization).
As discussed by \citet{Berg12a}, Random Search shares the advantages of Grid Search.
However, they show that Random Search may be preferable to Grid Search, especially 
in high-dimensional spaces or when the importance of individual parameters is fairly heterogeneous.
They hence suggest to use Random Search instead Grid Search if such simple procedures are required.
\citet{Prob19a} also use a Random Search variant to determine the tunability of models and hyperparameters.
For these reasons, we employ Random Search as a baseline for the comparison in our experimental investigation.

Next to Grid Search and Random Search, there are other undirected search methods.
Hyperband is an extension of Random Search, which controls the use of certain resources (e.g., iterations, training time)~\citep{Li18a}.
Another relevant set of methods are Design of Experiments methods, such as Latin Hypercube Designs~\citep{Lear03a}.

\subsubsection{Directed Search}
One obvious disadvantage of undirected search is, that a large amount of
the computational effort may be spent on evaluating solutions that cover the whole search space.
Hence, only a comparatively small amount of the computational budget will be spent on
potentially optimal or at least promising regions of the search space.

Directed search on the other hand may provide a more purposeful approach.
Basically any gradient-free, global optimization algorithm could be employed.
Prominent examples are Iterative Local Search~\citep{Hutt09a} and Iterative Racing~\citep{Lope16a}.
Metaheuristics like Evolutionary Algorithms or Swarm Optimization are also applicable~\citep{Yang20a}.
In comparison to undirected search procedures, directed search has two frequent drawbacks:
an increased complexity that makes implementation a larger issue, and being more complicated to parallelize.

In our own investigation, we employ a \textit{model-based} directed search procedure.
This is described in the following  Sec.~~\ref{sec:modelbased}.

\subsection{Model-based Search}\label{sec:modelbased}
A disadvantage of model-free, directed search procedures is that 
they may require a relatively large number of evaluations (i.e., long runtimes)
to approximate the values of optimal hyperparameters.

When tuning \ac{ml} algorithms, this can easily become problematic:
if complex \ac{ml} algorithms are tuned on large data sets, runtimes
of a single hyperparameter evaluation may easily go up into the range of hours.

Model-based search is one approach to resolve this issue.
These search procedures use information gathered during the search to
learn the relationship between hyperparameter values and quality measures (e.g., misclassification error).
The model that encodes this learned relationship is called the surrogate model.

The advantage of this approach is that a considerable part of the evaluation burden (i.e., the computational effort)
can be shifted from `real' evaluations to evaluations of the surrogate model, which should be faster
to evaluate.

\begin{figure}
\includegraphics[width=\textwidth]{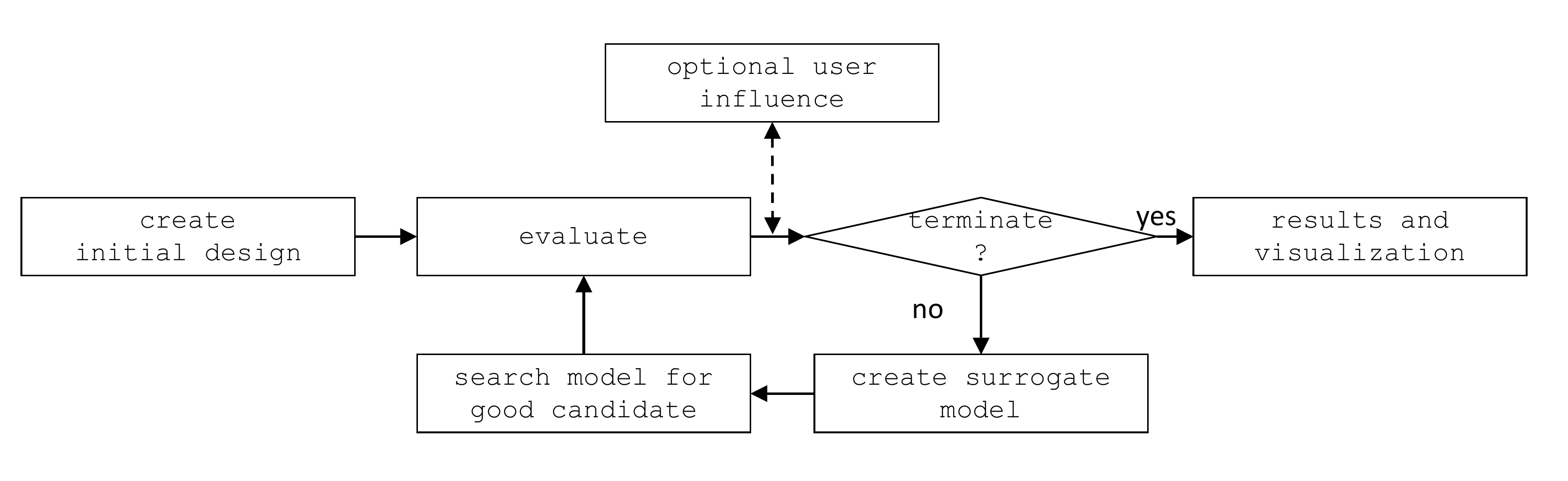}
\caption{Visual representation of model-based search with \gls{SPOT}.} \label{fig:spo}
\end{figure}

One variant of model-based search is \gls{SPOT}~\citep{Bart05a}. 
Details of \gls{SPOT} its application in practice is given by~\citet{Bart21a}.
\gls{SPOT} was originally developed for tuning of optimization algorithms.
The requirements and challenges of algorithm tuning in optimization broadly reflect those of 
tuning machine learning models.
\gls{SPOT} uses the following approach.
A detailed, algorithmic description is given by~\cite{preu07a}.
\begin{enumerate}
\item Design: 
In a first step, several candidate solutions (here: different combinations of hyperparameter values).
\item Evaluation: 
All new candidate solutions are evaluated (here: training the respective \ac{ml} model with the specified hyperparameter values
and measuring the quality / performance).
\item Termination: 
Check whether a termination criterion has been reached (e.g., number of iterations, evaluations, runtime).
\item Training:  
If not terminated, the surrogate model will be trained with all data derived from the evaluated candidate solutions,
thus learning how hyperparameters affect model quality.
\item Surrogate search:
The trained model is used to perform a search for new, promising candidate solutions
\item Go to 2.
\end{enumerate}
Note, that it can be useful to allow for user interaction with the tuner after the evaluation step.
Thus, the user may affect changes of the search space (stretch or shrink bounds on parameters, eliminate parameters).
However, we will consider a purely automatic search in our experiments.
Figure~\ref{fig:spo} summarizes the approach of \gls{SPOT}.

We use the R implementation of \gls{SPOT}, as provided by the R package \packSPOT~\citep{Bart21b,Bart21a}.

Table~\ref{tab:softoverview} (taken from~\citet{Bart19g}) compares \packSPOT with tuning procedures.
\begin{table}[tb]
  \centering
  \caption{Overview of common features of tuning procedures.
  (\checkmark) means that a feature is included.  
  ($\bigcirc$) means that an implementation of that feature is possible with minimal effort, according to the respective authors. 
  Table taken from~\citet{Bart19g}.}
  \label{tab:softoverview}
  \begin{tabular}{l|c|c|c|c|c|c}
  \hline
  Features \textbackslash Software & IRACE & SMAC & ParamILS & GGA & SPOT & Manual \\ \hline
  Supports Numerical Variables     &  \checkmark     &   \checkmark   &  \checkmark      &  \checkmark   &  \checkmark  & \checkmark \\
  Supports Categorical Variables   &  \checkmark     &   \checkmark   &  \checkmark       &  \checkmark   &  \checkmark  & \checkmark \\
  Interactive Execution            &       &      &         &     &   \checkmark  & \checkmark\\
  Multi Instance Problems          &      &   \checkmark   &        &     &    & \\
  Parallelization                  &   \checkmark    &      &   $\bigcirc$      &    \checkmark &  $\bigcirc$   & \\    
  Model-Based                      &       &  \checkmark    &         &  \checkmark   &   \checkmark  & \\
  Visualization                    &  \checkmark     &      &         &     &   \checkmark  & \\
  Noise Handling                   &  \checkmark     &      &         &     &   \checkmark  & \\
  Graphical User Interface         &       &      &         &     &   \checkmark  & \\
  \end{tabular}
\end{table}

\section{Case Study I: Tuning Decision Trees (rpart)}\label{sec:case1}
This case study considers the \acl{eb} algorithm which is detailed in Sec.~\ref{sec:eb}.
The hyperparameter tuning study applies \acl{eb} to the \gls{CID}, which will be described in Sec.~\ref{sec:census}.
This case study considers a classification task.

We will explain how to find suitable parameter values and bounds, and perform experiments w.r.t. the following four \acl{eb} hyperparameters:
\begin{enumerate}
      \item \paramminsplit
      \item \paramminbucket
      \item \paramcp
      \item \parammaxdepth
\end{enumerate}
The R package \gls{MLR} is used to set up a learning task.
The corresponding  parameters of this \gls{MLR} task can be 
accessed and modified via \gls{SPOT}. Section~\ref{sec:setup} illustrates
how the R packages can be combined.

\subsection{rpart: Project Setup}\label{sec:setup}
We start with the general setup.
Since we are considering classification (income class, i.e, high income versus low income), the \gls{MLR}
task.type is set to \Rclassif.
Regression trees are specified via \Rrpart.
\begin{knitrout}\footnotesize
\definecolor{shadecolor}{rgb}{0.969, 0.969, 0.969}\color{fgcolor}\begin{kframe}
\begin{alltt}
\hlstd{task.type} \hlkwb{<-} \hlstr{"classif"}
\hlstd{model} \hlkwb{<-} \hlstr{"rpart"}
\hlstd{data.seed} \hlkwb{<-} \hlnum{1}
\hlstd{tuner.seed} \hlkwb{<-} \hlnum{1}
\hlstd{timebudget} \hlkwb{<-} \hlnum{60}  \hlcom{# budget for tuning in seconds}
\hlstd{timeout} \hlkwb{<-} \hlnum{NA}  \hlcom{# do not use timeout (avoid overhead for quick tests)}
\end{alltt}
\end{kframe}
\end{knitrout}

\subsection{rpart: The Data Set}\label{sec:census}
For the investigation we choose the \gls{CID}, which is made available, for example, 
via the UCI Machine Learning Repository.~\footnote{\url{https://archive.ics.uci.edu/ml/datasets/Census-Income+(KDD)}}
This data set contains weighted census data extracted from the 1994 and 1995 Current Population Surveys conducted by the U.S. Census Bureau. It includes 41 demographic and employment related variables. 
For our investigation, we will access the version of the data set that is available via the
platform openml.org under the data record ID 45353~\citep{Vans13a}.
This data set is an excerpt from the current population surveys for 1994 and 1995, compiled by the U.S.~Census Bureau. 
It contains $m=299,285$ observations with 41 features on demography and employment. The data set is comparatively large, 
has many categorical features with many levels and fits well with the field of application of official statistics.

The \gls{CID} can be downloaded as follows:
\begin{knitrout}\footnotesize
\definecolor{shadecolor}{rgb}{0.969, 0.969, 0.969}\color{fgcolor}\begin{kframe}
\begin{alltt}
\hlstd{dirName} \hlkwb{=} \hlstr{"oml.cache"}
\hlkwa{if} \hlstd{(}\hlopt{!}\hlstd{(}\hlkwd{dir.exists}\hlstd{(dirName)))} \hlkwd{dir.create}\hlstd{(dirName)}
\hlkwd{setOMLConfig}\hlstd{(}\hlkwc{apikey} \hlstd{=} \hlstr{"c1994bdb7ecb3c6f3c8f3b35f4b47f1f"}\hlstd{,} \hlkwc{cachedir} \hlstd{= dirName)}
\end{alltt}
\begin{verbatim}
## OpenML configuration:
##   server           : http://www.openml.org/api/v1
##   cachedir         : oml.cache
##   verbosity        : 1
##   arff.reader      : farff
##   confirm.upload   : TRUE
##   apikey           : ***************************47f1f
\end{verbatim}
\begin{alltt}
\hlstd{dataOML} \hlkwb{<-} \hlkwd{getOMLDataSet}\hlstd{(}\hlnum{4535}\hlstd{)}\hlopt{$}\hlstd{data}
\end{alltt}
\end{kframe}
\end{knitrout}

Its 42 columns contain the following information:

\begin{knitrout}\footnotesize
\definecolor{shadecolor}{rgb}{0.969, 0.969, 0.969}\color{fgcolor}\begin{kframe}
\begin{alltt}
\hlkwd{str}\hlstd{(dataOML)}
\end{alltt}
\begin{verbatim}
## 'data.frame':	299285 obs. of  42 variables:
##  $ V1 : num  73 58 18 9 10 48 42 28 47 34 ...
##  $ V2 : Factor w/ 9 levels " Federal government",..: 4 7 4 4 4 5 5 5 2 5 ...
##  $ V3 : num  0 4 0 0 0 40 34 4 43 4 ...
##  $ V4 : num  0 34 0 0 0 10 3 40 26 37 ...
##  $ V5 : Factor w/ 17 levels " 10th grade",..: 13 17 1 11 11 17 10 13 17 17 ...
##  $ V6 : num  0 0 0 0 0 1200 0 0 876 0 ...
##  $ V7 : Factor w/ 3 levels " College or university",..: 3 3 2 3 3 3 3 3 3 3 ...
##  $ V8 : Factor w/ 7 levels " Divorced"," Married-A F spouse present",..: 7 1 5 5 5 3 3 5 3 3 ...
##  $ V9 : Factor w/ 24 levels " Agriculture",..: 15 5 15 15 15 7 8 5 6 5 ...
##  $ V10: Factor w/ 15 levels " Adm support including clerical",..: 7 9 7 7 7 11 3 5 1 6 ...
##  $ V11: Factor w/ 5 levels " Amer Indian Aleut or Eskimo",..: 5 5 2 5 5 1 5 5 5 5 ...
##  $ V12: Factor w/ 10 levels " All other"," Central or South American",..: 1 1 1 1 1 1 1 1 1 1 ...
##  $ V13: Factor w/ 2 levels " Female"," Male": 1 2 1 1 1 1 2 1 1 2 ...
##  $ V14: Factor w/ 3 levels " No"," Not in universe",..: 2 2 2 2 2 1 2 2 1 2 ...
##  $ V15: Factor w/ 6 levels " Job leaver",..: 4 4 4 4 4 4 4 2 4 4 ...
##  $ V16: Factor w/ 8 levels " Children or Armed Forces",..: 3 1 3 1 1 2 1 7 2 1 ...
##  $ V17: num  0 0 0 0 0 ...
##  $ V18: num  0 0 0 0 0 0 0 0 0 0 ...
##  $ V19: num  0 0 0 0 0 0 0 0 0 0 ...
##  $ V20: Factor w/ 6 levels " Head of household",..: 5 1 5 5 5 3 3 6 3 3 ...
##  $ V21: Factor w/ 6 levels " Abroad"," Midwest",..: 4 5 4 4 4 4 4 4 4 4 ...
##  $ V22: Factor w/ 51 levels " ?"," Abroad",..: 37 6 37 37 37 37 37 37 37 37 ...
##  $ V23: Factor w/ 38 levels " Child <18 ever marr not in subfamily",..: 30 21 8 3 3 37 21 36 37 21 ...
##  $ V24: Factor w/ 8 levels " Child 18 or older",..: 7 5 1 3 3 8 5 6 8 5 ...
##  $ V25: num  1700 1054 992 1758 1069 ...
##  $ V26: Factor w/ 10 levels " ?"," Abroad to MSA",..: 1 4 1 6 6 1 6 1 1 6 ...
##  $ V27: Factor w/ 9 levels " ?"," Abroad",..: 1 9 1 7 7 1 7 1 1 7 ...
##  $ V28: Factor w/ 10 levels " ?"," Abroad",..: 1 10 1 8 8 1 8 1 1 8 ...
##  $ V29: Factor w/ 3 levels " No"," Not in universe under 1 year old",..: 2 1 2 3 3 2 3 2 2 3 ...
##  $ V30: Factor w/ 4 levels " ?"," No"," Not in universe",..: 1 4 1 3 3 1 3 1 1 3 ...
##  $ V31: num  0 1 0 0 0 1 6 4 5 6 ...
##  $ V32: Factor w/ 5 levels " Both parents present",..: 5 5 5 1 1 5 5 5 5 5 ...
##  $ V33: Factor w/ 43 levels " ?"," Cambodia",..: 41 41 42 41 41 32 41 41 41 41 ...
##  $ V34: Factor w/ 43 levels " ?"," Cambodia",..: 41 41 42 41 41 41 41 41 41 41 ...
##  $ V35: Factor w/ 43 levels " ?"," Cambodia",..: 41 41 42 41 41 41 41 41 41 41 ...
##  $ V36: Factor w/ 5 levels " Foreign born- Not a citizen of U S ",..: 5 5 1 5 5 5 5 5 5 5 ...
##  $ V37: num  0 0 0 0 0 2 0 0 0 0 ...
##  $ V38: Factor w/ 3 levels " No"," Not in universe",..: 2 2 2 2 2 2 2 2 2 2 ...
##  $ V39: num  2 2 2 0 0 2 2 2 2 2 ...
##  $ V40: num  0 52 0 0 0 52 52 30 52 52 ...
##  $ V41: num  95 94 95 94 94 95 94 95 95 94 ...
##  $ V42: Factor w/ 2 levels " - 50000."," 50000+.": 1 1 1 1 1 1 1 1 1 1 ...
\end{verbatim}
\end{kframe}
\end{knitrout}

\subsection{rpart: Create an mlr Task}
The  target variable of the data set is the income class (income below or above US \$ 50,000).
It is stored in column 42.
\begin{knitrout}\footnotesize
\definecolor{shadecolor}{rgb}{0.969, 0.969, 0.969}\color{fgcolor}\begin{kframe}
\begin{alltt}
\hlstd{task} \hlkwb{<-} \hlkwd{makeClassifTask}\hlstd{(}\hlkwc{data} \hlstd{= dataOML,} \hlkwc{target} \hlstd{=} \hlstr{"V42"}\hlstd{)}
\end{alltt}
\end{kframe}
\end{knitrout}

Using \gls{MLR}, a resampling strategy can be defined as follows:
\begin{knitrout}\footnotesize
\definecolor{shadecolor}{rgb}{0.969, 0.969, 0.969}\color{fgcolor}\begin{kframe}
\begin{alltt}
\hlstd{rsmpl} \hlkwb{<-} \hlkwd{makeResampleDesc}\hlstd{(}\hlstr{"Holdout"}\hlstd{,} \hlkwc{split} \hlstd{=} \hlnum{0.6}\hlstd{)}
\end{alltt}
\end{kframe}
\end{knitrout}

\subsection{rpart: Define the Learner (Algorithm)}

The we define the \gls{MLR} learner and print the learner's parameters. Please note, that the parameter bounds given by \gls{MLR} are not always correct.
If they are used, they may lead to invalid configurations 
(the tuning procedure will try to work around this via a \enquote{penalty},
but this may be an avoidable waste of run time).

\begin{knitrout}\footnotesize
\definecolor{shadecolor}{rgb}{0.969, 0.969, 0.969}\color{fgcolor}\begin{kframe}
\begin{alltt}
\hlstd{learner} \hlkwb{<-} \hlkwd{paste}\hlstd{(task.type, model,} \hlkwc{sep} \hlstd{=} \hlstr{"."}\hlstd{)}
\hlkwd{print}\hlstd{(}\hlkwd{getParamSet}\hlstd{(learner))}
\end{alltt}
\begin{verbatim}
##                    Type len  Def   Constr Req Tunable Trafo
## minsplit        integer   -   20 1 to Inf   -    TRUE     -
## minbucket       integer   -    - 1 to Inf   -    TRUE     -
## cp              numeric   - 0.01   0 to 1   -    TRUE     -
## maxcompete      integer   -    4 0 to Inf   -    TRUE     -
## maxsurrogate    integer   -    5 0 to Inf   -    TRUE     -
## usesurrogate   discrete   -    2    0,1,2   -    TRUE     -
## surrogatestyle discrete   -    0      0,1   -    TRUE     -
## maxdepth        integer   -   30  1 to 30   -    TRUE     -
## xval            integer   -   10 0 to Inf   -   FALSE     -
## parms           untyped   -    -        -   -    TRUE     -
\end{verbatim}
\end{kframe}
\end{knitrout}

\subsection{rpart: Define the Experiment Configuration}

The experiment configuration contains information about the \gls{MLR} learner to be tuned, the task,
the resampling strategy, parameter names as specified for the \gls{MLR} learner, 
lower and upper bounds (see also transformations), data type of each parameter, 
and further information.
Fixed parameters can be specified, in our case, there are none, so the list is empty.
The same holds for the factor levels for categorical parameters. There are none of them in our configuration.

Important note:
\paramminbucket is set relative to \paramminsplit, i.e., we are using numerical values for \paramminbucket that
represent \emph{percentages} relative to \paramminsplit.
If \paramminbucket = 1.0, then \paramminbucket = \paramminsplit (\paramminsplit values should be greater equal \paramminbucket values).

\begin{knitrout}\footnotesize
\definecolor{shadecolor}{rgb}{0.969, 0.969, 0.969}\color{fgcolor}\begin{kframe}
\begin{alltt}
\hlstd{cfg} \hlkwb{<-} \hlkwd{list}\hlstd{(}
  \hlkwc{learner} \hlstd{= learner,}
  \hlkwc{task} \hlstd{= task,}
  \hlkwc{resample} \hlstd{= rsmpl,}
  \hlkwc{tunepars} \hlstd{=} \hlkwd{c}\hlstd{(}\hlstr{"minsplit"}\hlstd{,} \hlstr{"minbucket"}\hlstd{,} \hlstr{"cp"}\hlstd{,} \hlstr{"maxdepth"}\hlstd{),}
  \hlkwc{lower} \hlstd{=} \hlkwd{c}\hlstd{(}\hlnum{1}\hlstd{,}
            \hlnum{0.1}\hlstd{,}\hlopt{-}\hlnum{10}\hlstd{,}
            \hlnum{1}\hlstd{),}
  \hlkwc{upper} \hlstd{=} \hlkwd{c}\hlstd{(}\hlnum{300}\hlstd{,}
            \hlnum{0.5}\hlstd{,}
            \hlnum{0}\hlstd{,}
            \hlnum{30}\hlstd{),}
  \hlkwc{type} \hlstd{=}  \hlkwd{c}\hlstd{(}\hlstr{"integer"}\hlstd{,} \hlstr{"numeric"}\hlstd{,} \hlstr{"numeric"}\hlstd{,} \hlstr{"integer"}\hlstd{),}
  \hlkwc{fixpars} \hlstd{=} \hlkwd{list}\hlstd{(),}
  \hlkwc{factorlevels} \hlstd{=} \hlkwd{list}\hlstd{(),}
  \hlkwc{transformations} \hlstd{=} \hlkwd{c}\hlstd{(trans_id,} \hlcom{#identity transformation}
                      \hlstd{trans_id,}
                      \hlstd{trans_10pow,} \hlcom{# x^10 transformation}
                      \hlstd{trans_id),}
  \hlkwc{dummy} \hlstd{=} \hlnum{FALSE}\hlstd{,}
  \hlcom{# set parameters relative to other parameters. here: minbucket relative to minsplit}
  \hlkwc{relpars} \hlstd{=} \hlkwd{list}\hlstd{(}\hlkwc{minbucket} \hlstd{=} \hlkwd{expression}\hlstd{(}\hlkwd{round}\hlstd{(}
    \hlkwd{max}\hlstd{(minsplit} \hlopt{*} \hlstd{minbucket,} \hlnum{1}\hlstd{)}
  \hlstd{)))}
  \hlcom{# if no relative parameters needed, use empty list for relpars}
\hlstd{)}
\end{alltt}
\end{kframe}
\end{knitrout}

The, the objective function can be defined:

\begin{knitrout}\footnotesize
\definecolor{shadecolor}{rgb}{0.969, 0.969, 0.969}\color{fgcolor}\begin{kframe}
\begin{alltt}
\hlstd{objf} \hlkwb{<-} \hlkwd{get_objf}\hlstd{(}\hlkwc{config} \hlstd{= cfg,} \hlkwc{timeout} \hlstd{= timeout)}
\end{alltt}
\end{kframe}
\end{knitrout}

\subsection{rpart: Run Tuning with SPOT}
Now, everything is prepared and the \gls{SPOT} run can be started.
Note, that using \texttt{timebudget/60} converts the timebudget to minutes.

\begin{knitrout}\footnotesize
\definecolor{shadecolor}{rgb}{0.969, 0.969, 0.969}\color{fgcolor}\begin{kframe}
\begin{alltt}
\hlstd{result} \hlkwb{<-} \hlkwd{spot}\hlstd{(}\hlkwc{fun} \hlstd{= objf,} \hlkwc{lower} \hlstd{= cfg}\hlopt{$}\hlstd{lower,} \hlkwc{upper} \hlstd{= cfg}\hlopt{$}\hlstd{upper,} \hlkwc{control} \hlstd{=} \hlkwd{list}\hlstd{(}\hlkwc{types} \hlstd{= cfg}\hlopt{$}\hlstd{type,}
    \hlkwc{maxTime} \hlstd{= timebudget}\hlopt{/}\hlnum{60}\hlstd{,} \hlkwc{plots} \hlstd{=} \hlnum{FALSE}\hlstd{,} \hlkwc{progress} \hlstd{=} \hlnum{FALSE}\hlstd{,} \hlkwc{model} \hlstd{= buildKriging,}
    \hlkwc{optimizer} \hlstd{= optimDE,} \hlkwc{noise} \hlstd{=} \hlnum{TRUE}\hlstd{,} \hlkwc{seedFun} \hlstd{=} \hlnum{123}\hlstd{,} \hlkwc{seedSPOT} \hlstd{= tuner.seed,} \hlkwc{designControl} \hlstd{=} \hlkwd{list}\hlstd{(}\hlkwc{size} \hlstd{=} \hlnum{5} \hlopt{*}
        \hlkwd{length}\hlstd{(cfg}\hlopt{$}\hlstd{lower)),} \hlkwc{funEvals} \hlstd{=} \hlnum{Inf}\hlstd{,} \hlkwc{modelControl} \hlstd{=} \hlkwd{list}\hlstd{(}\hlkwc{target} \hlstd{=} \hlstr{"y"}\hlstd{,} \hlkwc{useLambda} \hlstd{=} \hlnum{TRUE}\hlstd{,}
        \hlkwc{reinterpolate} \hlstd{=} \hlnum{TRUE}\hlstd{),} \hlkwc{optimizerControl} \hlstd{=} \hlkwd{list}\hlstd{(}\hlkwc{funEvals} \hlstd{=} \hlnum{100} \hlopt{*} \hlkwd{length}\hlstd{(cfg}\hlopt{$}\hlstd{lower))))}

\hlkwd{save}\hlstd{(result,} \hlkwc{file} \hlstd{=} \hlstr{"rpartResult.RData"}\hlstd{)}
\end{alltt}
\end{kframe}
\end{knitrout}

\begin{knitrout}\footnotesize
\definecolor{shadecolor}{rgb}{0.969, 0.969, 0.969}\color{fgcolor}\begin{kframe}
\begin{alltt}
\hlkwd{load}\hlstd{(}\hlkwc{file} \hlstd{=} \hlstr{"R.d/case1Result.RData"}\hlstd{)}
\end{alltt}
\end{kframe}
\end{knitrout}

\subsection{rpart: Visualize the Results}\label{sec:visRpart}

First, we visualize the progress of the \gls{SPOT} tuning procedure.

\begin{knitrout}\footnotesize
\definecolor{shadecolor}{rgb}{0.969, 0.969, 0.969}\color{fgcolor}\begin{kframe}
\begin{alltt}
\hlstd{size} \hlkwb{<-} \hlnum{5} \hlopt{*} \hlkwd{length}\hlstd{(cfg}\hlopt{$}\hlstd{lower)}
\hlkwd{plot}\hlstd{(result}\hlopt{$}\hlstd{y,} \hlkwc{type} \hlstd{=} \hlstr{"l"}\hlstd{)}
\hlkwd{abline}\hlstd{(}\hlkwc{v} \hlstd{= size,} \hlkwc{col} \hlstd{=} \hlstr{"red"}\hlstd{)}
\hlstd{y0} \hlkwb{<-} \hlkwd{min}\hlstd{(result}\hlopt{$}\hlstd{y[}\hlnum{1}\hlopt{:}\hlstd{size])}
\hlkwd{abline}\hlstd{(}\hlkwc{h} \hlstd{= y0,} \hlkwc{col} \hlstd{=} \hlstr{"blue"}\hlstd{)}
\end{alltt}
\end{kframe}\begin{figure}

{\centering \includegraphics{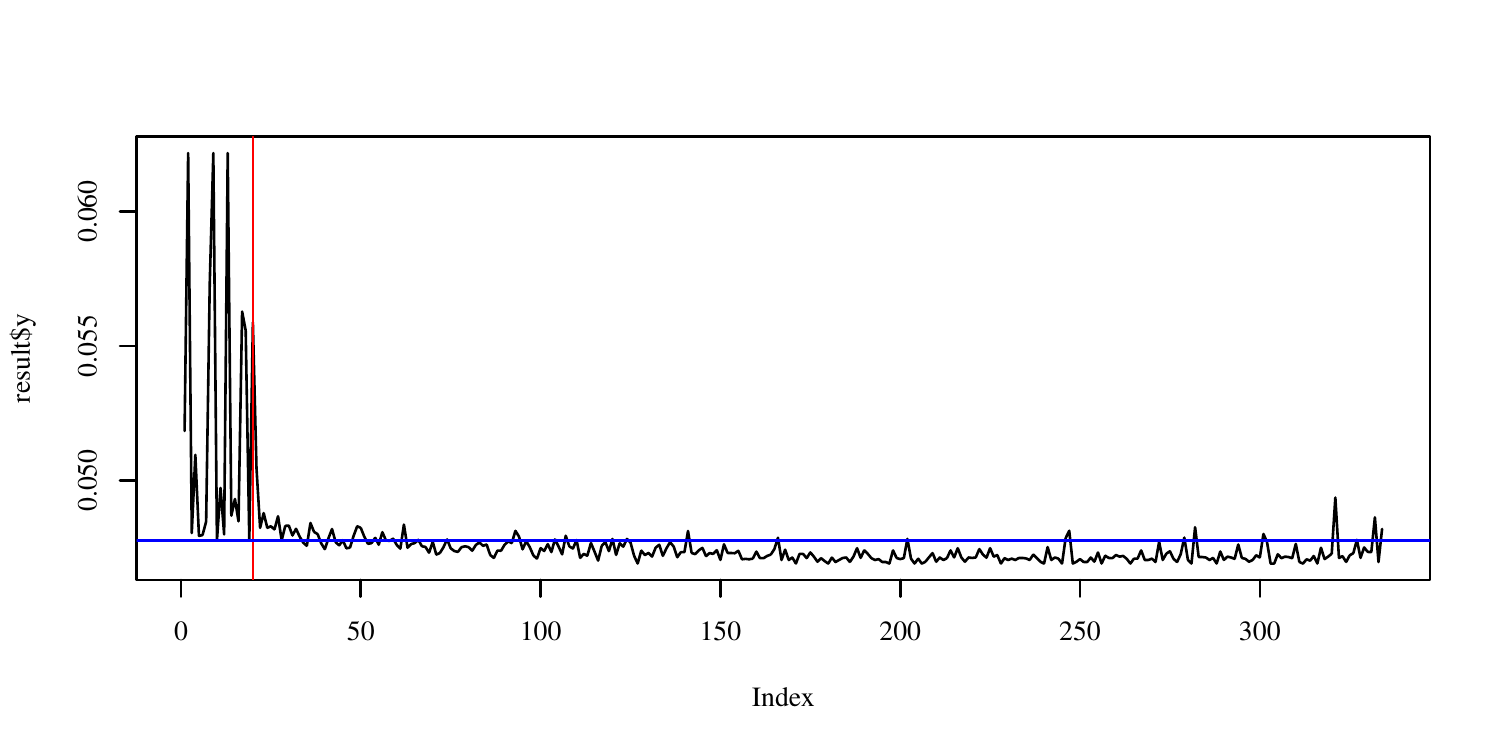} 

}

\caption[Case Study I]{Case Study I: SPOT search. The vertical red line denotes the end of the initial design phase. The best result from the (random) initial phase is shown in blue.}\label{fig:plotSpot1}
\end{figure}

\end{knitrout}

The best hyperparameter setting can be printed as follows:
\begin{knitrout}\footnotesize
\definecolor{shadecolor}{rgb}{0.969, 0.969, 0.969}\color{fgcolor}\begin{kframe}
\begin{alltt}
\hlstd{(paraNames} \hlkwb{<-} \hlstd{cfg}\hlopt{$}\hlstd{tunepars)}
\end{alltt}
\begin{verbatim}
## [1] "minsplit"  "minbucket" "cp"        "maxdepth"
\end{verbatim}
\begin{alltt}
\hlkwd{print}\hlstd{(result}\hlopt{$}\hlstd{xbest)}
\end{alltt}
\begin{verbatim}
##      [,1]      [,2]      [,3] [,4]
## [1,]   16 0.1770681 -3.213333   28
\end{verbatim}
\end{kframe}
\end{knitrout}

\begin{figure}
\centering
\includegraphics[width=0.5\linewidth]{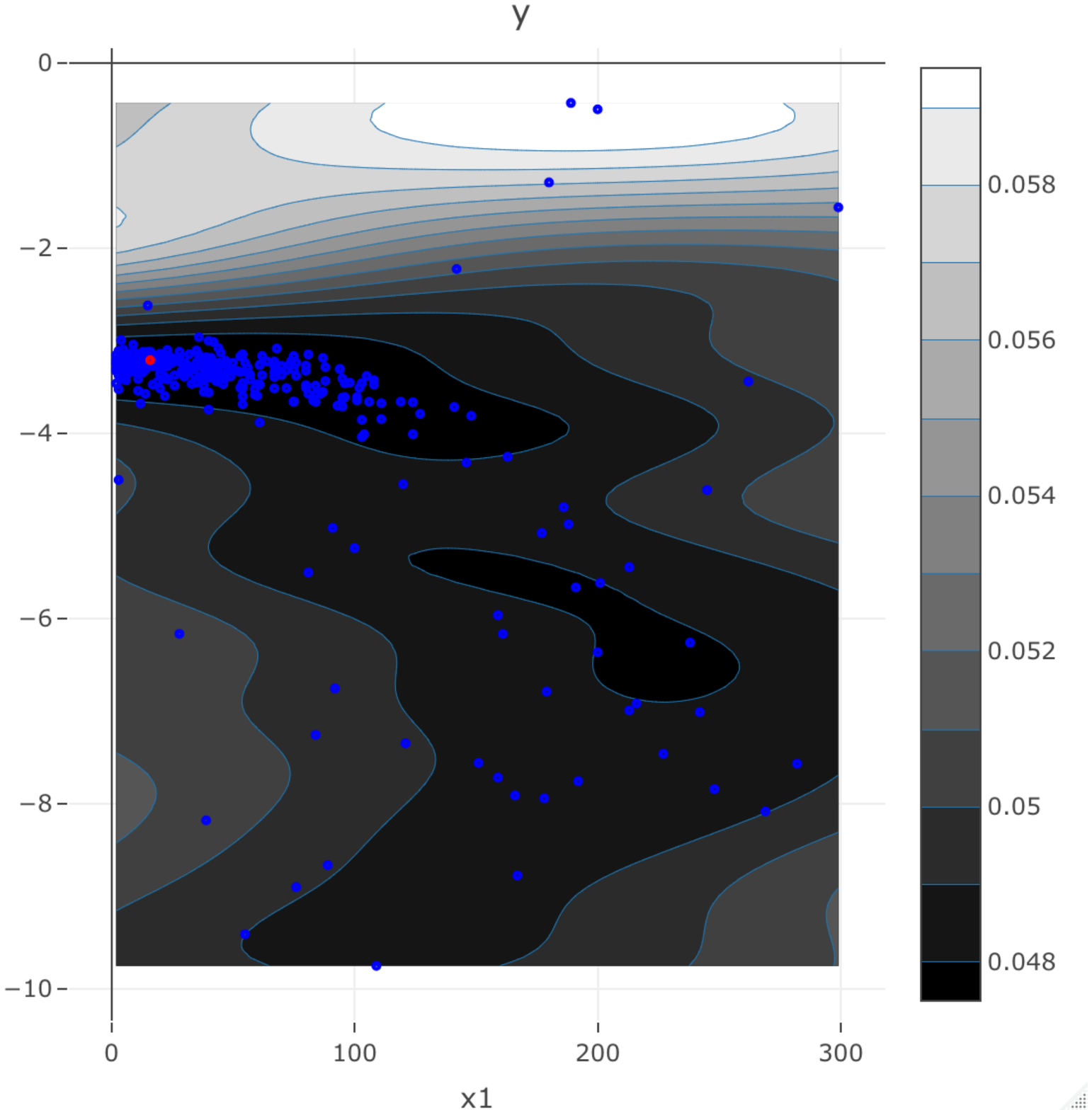}
\caption{Surface plot: mce as a funtion of minsplit and cp. Note that the cp values (on the vertical axis) use the $10^{x}$ transformation.}
\label{fig:rpartSpotTuning01}
\end{figure}

The error (mce) can be shown as a function of minsplit and cp (based on surrogate) as shown in Fig.~\ref{fig:rpartSpotTuning01}.
\begin{knitrout}\footnotesize
\definecolor{shadecolor}{rgb}{0.969, 0.969, 0.969}\color{fgcolor}\begin{kframe}
\begin{alltt}
\hlkwd{plot_surface}\hlstd{(result,} \hlkwc{yvar} \hlstd{=} \hlnum{1}\hlstd{,} \hlkwc{which} \hlstd{=} \hlkwd{c}\hlstd{(}\hlnum{1}\hlstd{,} \hlnum{3}\hlstd{))}
\end{alltt}
\end{kframe}
\end{knitrout}

\begin{figure}
\centering
\includegraphics[width=0.5\linewidth]{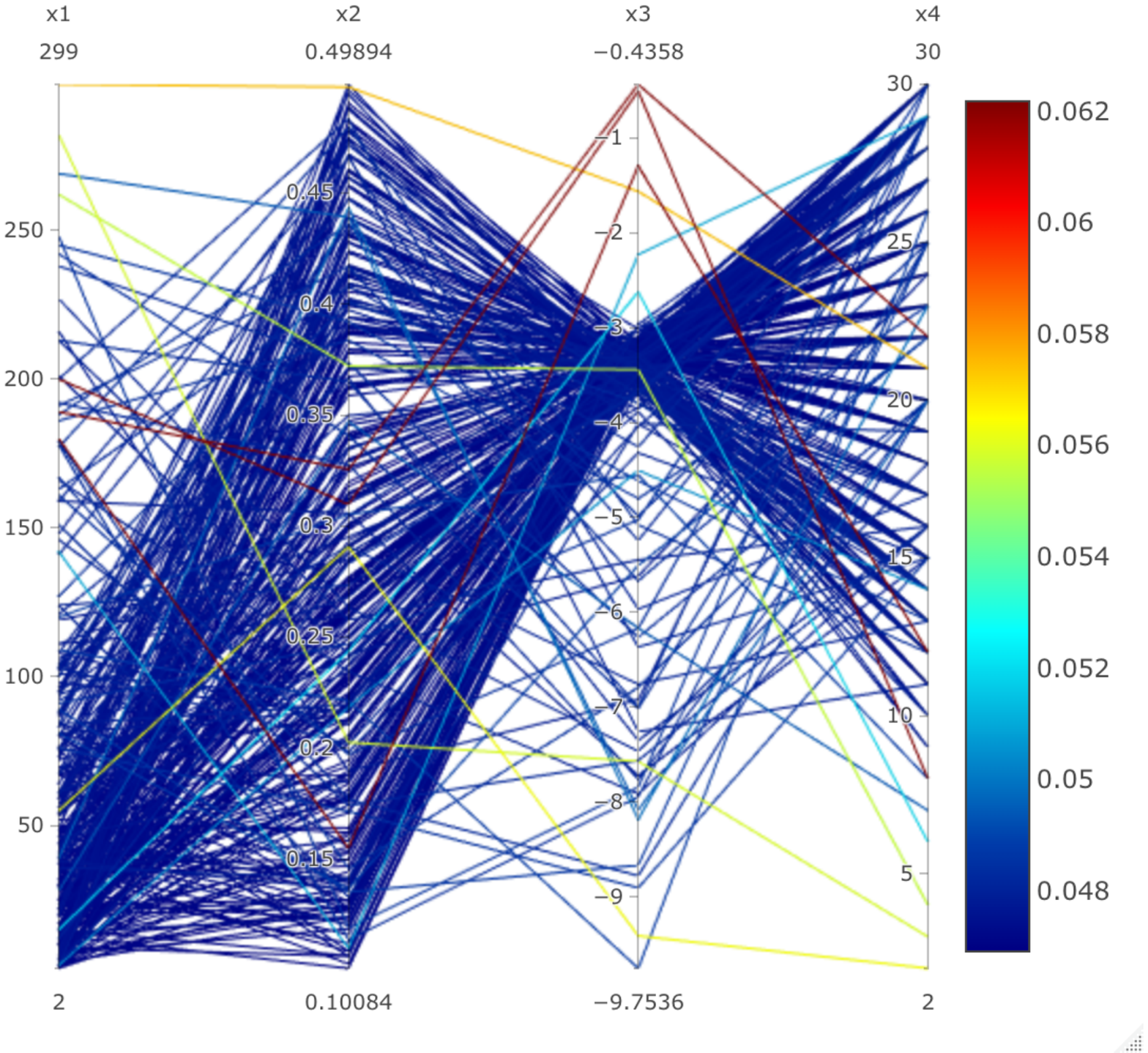}
\caption{Parallel plot. Red lines denote poor configurations, e.g., large $x_3$ (cp) values are not recommended.}
\label{fig:rpartSpotTuning02}
\end{figure}

Next, mce vs all parameters  (based on evaluations) is shown in the parallel plot in 
Fig.~\ref{fig:rpartSpotTuning02}.
\begin{knitrout}\footnotesize
\definecolor{shadecolor}{rgb}{0.969, 0.969, 0.969}\color{fgcolor}\begin{kframe}
\begin{alltt}
\hlkwd{plot_parallel}\hlstd{(result,} \hlkwc{yvar} \hlstd{=} \hlnum{1}\hlstd{)}
\end{alltt}
\end{kframe}
\end{knitrout}

In addition, a sensitivity plot can be shown.

\begin{knitrout}\footnotesize
\definecolor{shadecolor}{rgb}{0.969, 0.969, 0.969}\color{fgcolor}\begin{kframe}
\begin{alltt}
\hlkwd{plot_sensitivity}\hlstd{(result,} \hlkwc{type} \hlstd{=} \hlstr{"agg"}\hlstd{)}
\end{alltt}
\end{kframe}
\end{knitrout}

\begin{figure}
\centering
\includegraphics[width=0.5\linewidth]{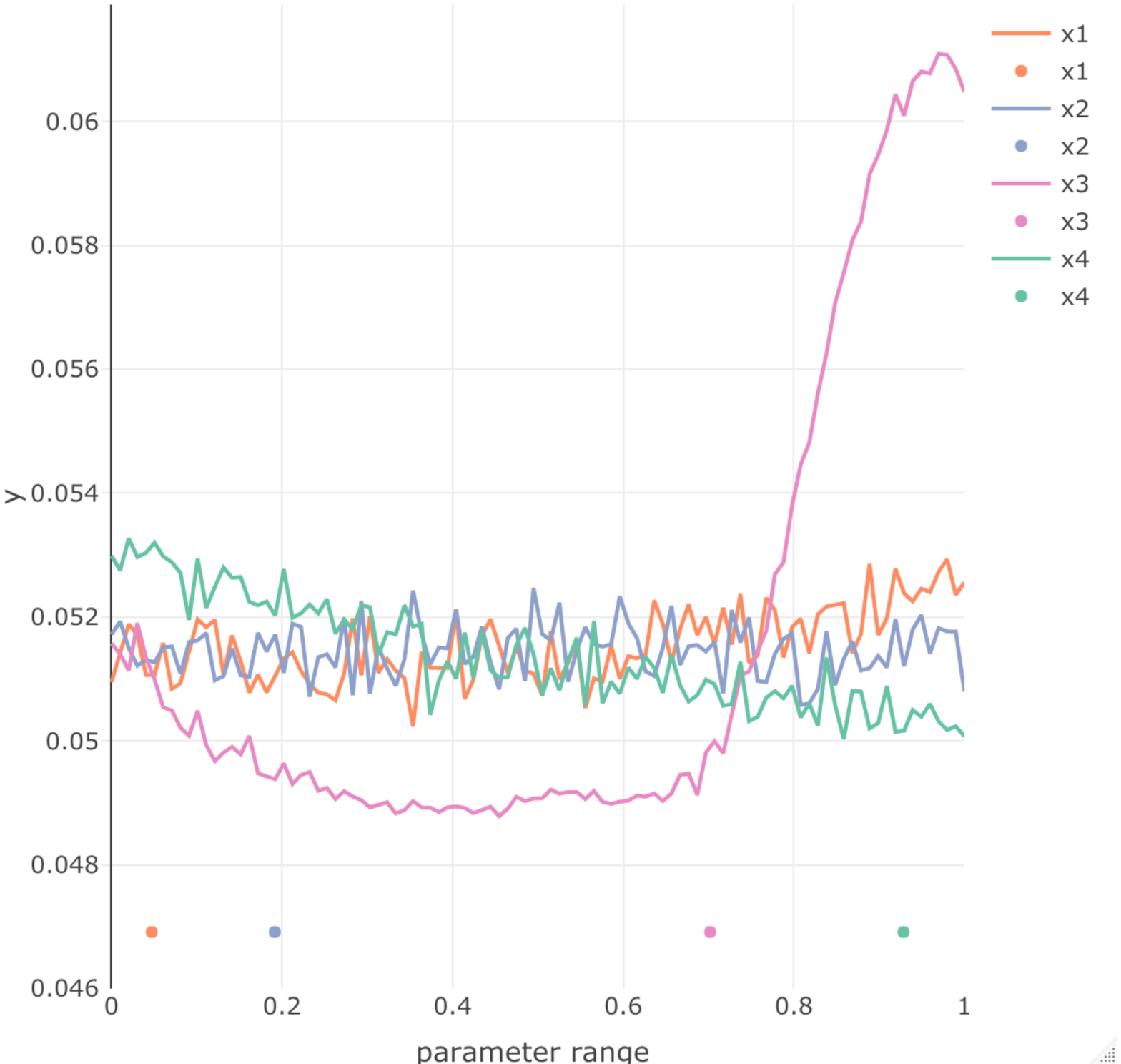}
\caption{Sensitivity plot. Effect of normalized hyperparamter values. \paramcp ($x_3$) appears to be the most important parameter.}
\label{fig:rpartSpotTuning03}
\end{figure}

Finally, we fit a simple regression tree to analyze the most important
hyperparameters.

\begin{knitrout}\footnotesize
\definecolor{shadecolor}{rgb}{0.969, 0.969, 0.969}\color{fgcolor}\begin{kframe}
\begin{alltt}
\hlkwd{library}\hlstd{(}\hlstr{"rpart"}\hlstd{)}
\hlkwd{library}\hlstd{(}\hlstr{"rpart.plot"}\hlstd{)}
\hlstd{fitTree} \hlkwb{<-} \hlkwd{buildTreeModel}\hlstd{(}\hlkwc{x} \hlstd{= result}\hlopt{$}\hlstd{x,} \hlkwc{y} \hlstd{= result}\hlopt{$}\hlstd{y,} \hlkwc{control} \hlstd{=} \hlkwd{list}\hlstd{(}\hlkwc{xnames} \hlstd{= paraNames))}  \hlcom{# paste0('x', 1:n)))}
\hlkwd{rpart.plot}\hlstd{(fitTree}\hlopt{$}\hlstd{fit)}
\end{alltt}
\end{kframe}

{\centering \includegraphics{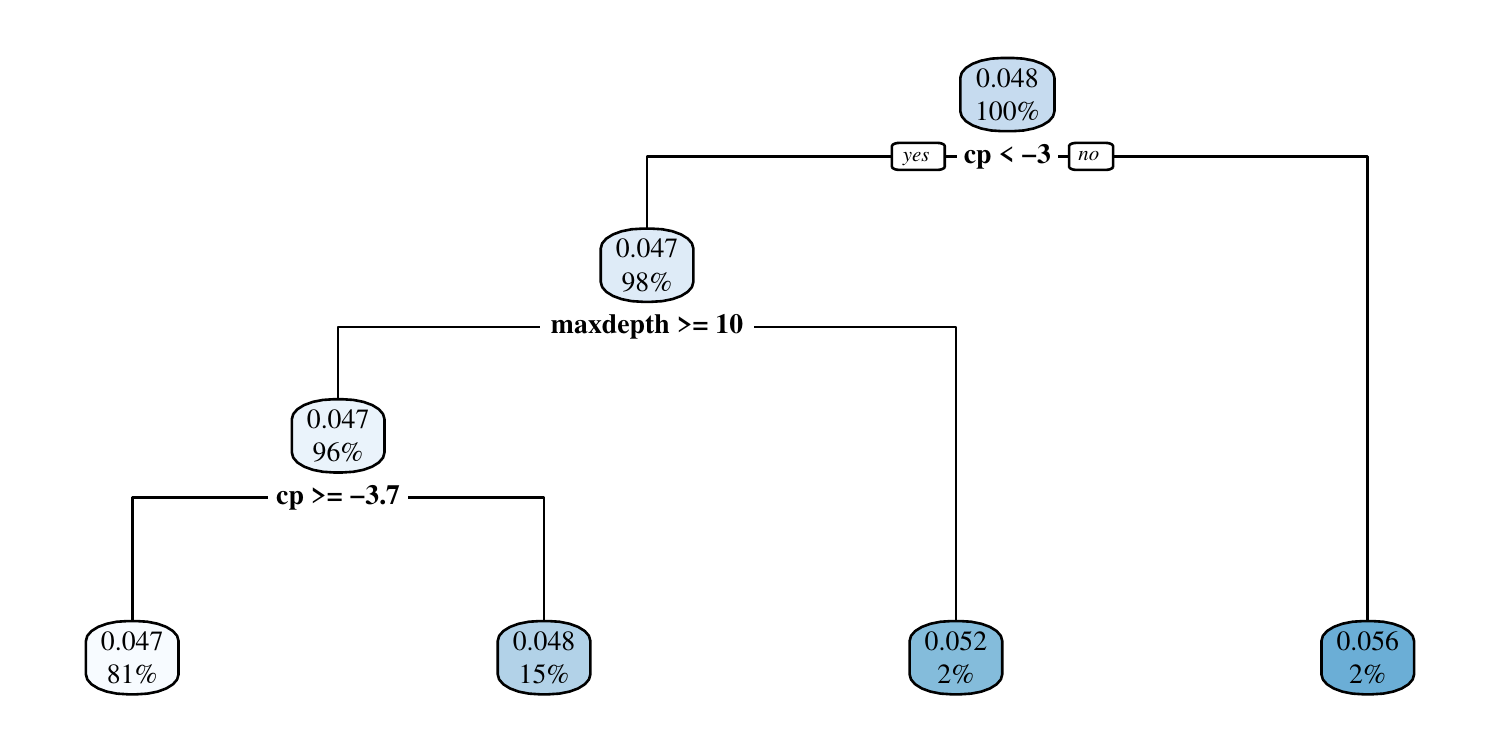} 

}

\end{knitrout}

Our visualizations are based on the transformed values, i.e.,
the power transformation was applied to the cp values.
Natural hyperparameter values can be obtained as follows:

\begin{knitrout}\footnotesize
\definecolor{shadecolor}{rgb}{0.969, 0.969, 0.969}\color{fgcolor}\begin{kframe}
\begin{alltt}
\hlstd{result}\hlopt{$}\hlstd{x[,} \hlnum{3}\hlstd{]} \hlkwb{<-} \hlkwd{trans_10pow}\hlstd{(result}\hlopt{$}\hlstd{x[,} \hlnum{3}\hlstd{])}
\end{alltt}
\end{kframe}
\end{knitrout}

\subsection{rpart: Summary}

Data visualizations shown in Sec.~\ref{sec:visRpart} indicate that hyperparameter
\paramcp has the greatest effect on the algorithm's performance.
The recommended value is $ \paramcp \approx 10^{-3} = 0.001$.

\section{Case Study II: Tuning XGBoost}\label{sec:case2}
Consider the \acl{xgb} algorithm which is detailed in Sec.~\ref{sec:xgboost}.
We will explain how to find suitable parameter values and bounds, and perform experiments w.r.t. the following \acl{xgb} parameters:
\begin{enumerate}
      \item \paramnrounds
      \item \parameta
      \item \paramlambda
      \item \paramalpha 
      \item \paramsubsample
      \item \paramcolsample
      \item \paramgamma
      \item \parammaxdepthx
      \item \paramminchild 
\end{enumerate}

\subsection{The Census-Income (KDD) Data Set}
The first step is identical to the step in the \Rrpart example in Sec.~\ref{sec:case1}.

\begin{knitrout}\footnotesize
\definecolor{shadecolor}{rgb}{0.969, 0.969, 0.969}\color{fgcolor}\begin{kframe}
\begin{alltt}
\hlstd{dirName} \hlkwb{=} \hlstr{"oml.cache"}
\hlkwa{if} \hlstd{(}\hlopt{!}\hlstd{(}\hlkwd{dir.exists}\hlstd{(dirName)))} \hlkwd{dir.create}\hlstd{(dirName)}
\hlkwd{setOMLConfig}\hlstd{(}\hlkwc{apikey} \hlstd{=} \hlstr{"c1994bdb7ecb3c6f3c8f3b35f4b47f1f"}\hlstd{,} \hlkwc{cachedir} \hlstd{= dirName)}
\end{alltt}
\begin{verbatim}
## OpenML configuration:
##   server           : http://www.openml.org/api/v1
##   cachedir         : oml.cache
##   verbosity        : 1
##   arff.reader      : farff
##   confirm.upload   : TRUE
##   apikey           : ***************************47f1f
\end{verbatim}
\begin{alltt}
\hlstd{dataOML} \hlkwb{<-} \hlkwd{getOMLDataSet}\hlstd{(}\hlnum{4535}\hlstd{)}\hlopt{$}\hlstd{data}
\hlstd{task} \hlkwb{<-} \hlkwd{makeClassifTask}\hlstd{(}\hlkwc{data} \hlstd{= dataOML,} \hlkwc{target} \hlstd{=} \hlstr{"V42"}\hlstd{)}
\hlstd{rsmpl} \hlkwb{<-} \hlkwd{makeResampleDesc}\hlstd{(}\hlstr{"Holdout"}\hlstd{,} \hlkwc{split} \hlstd{=} \hlnum{0.6}\hlstd{)}
\end{alltt}
\end{kframe}
\end{knitrout}

\subsection{\acl{xgb}: Project Setup}
The project setup is also similar to the setup described in Sec.~\ref{sec:case1}.
Because the \gls{XGBoost} algorithm is more complex than the \gls{DT}, 
the computational budget is increased.
The budget for tuning in seconds is set to $15 \times 3,600$ seconds or 15 hours.

\begin{knitrout}\footnotesize
\definecolor{shadecolor}{rgb}{0.969, 0.969, 0.969}\color{fgcolor}\begin{kframe}
\begin{alltt}
\hlstd{task.type} \hlkwb{<-} \hlstr{"classif"}
\hlstd{data.seed} \hlkwb{<-} \hlnum{1}
\hlstd{tuner.seed} \hlkwb{<-} \hlnum{1}
\hlstd{timebudget} \hlkwb{<-} \hlnum{15} \hlopt{*} \hlnum{3600}
\hlstd{timeout} \hlkwb{<-} \hlstd{timebudget}\hlopt{/}\hlnum{20}  \hlcom{# use 1/20 times the budget before tuning is stopped}
\end{alltt}
\end{kframe}
\end{knitrout}

\subsection{\acl{xgb}: Learner (Algorithm) Definition}
The mode name is \enquote{xgboost}.

\begin{knitrout}\footnotesize
\definecolor{shadecolor}{rgb}{0.969, 0.969, 0.969}\color{fgcolor}\begin{kframe}
\begin{alltt}
\hlstd{model} \hlkwb{<-} \hlstr{"xgboost"}
\hlstd{learner} \hlkwb{<-} \hlkwd{paste}\hlstd{(task.type, model,} \hlkwc{sep} \hlstd{=} \hlstr{"."}\hlstd{)}
\end{alltt}
\end{kframe}
\end{knitrout}

The complete list of \gls{XGBoost} hyperparameters can be shown using the \gls{MLR} function 
\texttt{getParamSet}. Since this is a long list, it is not shown here.

\begin{knitrout}\footnotesize
\definecolor{shadecolor}{rgb}{0.969, 0.969, 0.969}\color{fgcolor}\begin{kframe}
\begin{alltt}
\hlkwd{print}\hlstd{(}\hlkwd{getParamSet}\hlstd{(learner))}
\end{alltt}
\end{kframe}
\end{knitrout}

\subsection{\acl{xgb}: Experiment Configuration}
Now follows the  most important part of the \gls{HPT} task: the specification of a reasonable 
hyperparameter  setting.
We will use the examples from literature shown in Table~\ref{table:survey:xgbparam} in Sec.~\ref{sec:xgboost}
as a guideline.

\begin{knitrout}\footnotesize
\definecolor{shadecolor}{rgb}{0.969, 0.969, 0.969}\color{fgcolor}\begin{kframe}
\begin{alltt}
    \hlstd{tunepars} \hlkwb{<-} \hlkwd{c}\hlstd{(}\hlstr{"nrounds"}\hlstd{,}\hlstr{"eta"}\hlstd{,}\hlstr{"lambda"}\hlstd{,}\hlstr{"alpha"}\hlstd{,}\hlstr{"subsample"}\hlstd{,}\hlstr{"colsample_bytree"}\hlstd{,}\hlstr{"gamma"}\hlstd{,}\hlstr{"max_depth"}\hlstd{,}\hlstr{"min_child_weight"}\hlstd{)}
    \hlstd{lower} \hlkwb{<-} \hlkwd{c}\hlstd{(}\hlnum{0}\hlstd{,}
               \hlopt{-}\hlnum{10}\hlstd{,}
               \hlopt{-}\hlnum{10}\hlstd{,}
               \hlopt{-}\hlnum{10}\hlstd{,}
               \hlnum{0.1}\hlstd{,}
               \hlnum{1}\hlopt{/}\hlkwd{getTaskNFeats}\hlstd{(task),}
               \hlcom{## slight deviation from Prob19a.}
               \hlcom{## reason: 0 makes no sense. At least one feature should be chosen via colsample.}
               \hlopt{-}\hlnum{10}\hlstd{,} \hlcom{## smaller than Thom18a}
               \hlnum{1}\hlstd{,}
               \hlnum{0}\hlstd{)}
    \hlstd{upper} \hlkwb{<-} \hlkwd{c}\hlstd{(}\hlnum{5}\hlstd{,}  \hlcom{## set similar as random forest (which is less than Prob19a used: 5000)}
               \hlnum{0}\hlstd{,}
               \hlnum{10}\hlstd{,}
               \hlnum{10}\hlstd{,}
               \hlnum{1}\hlstd{,}
               \hlnum{1}\hlstd{,}
               \hlnum{10}\hlstd{,}  \hlcom{## larger than Thom18a}
               \hlnum{15}\hlstd{,}
               \hlnum{7}\hlstd{)}
    \hlstd{type} \hlkwb{<-}  \hlkwd{c}\hlstd{(}\hlstr{"numeric"}\hlstd{,}\hlstr{"numeric"}\hlstd{,}\hlstr{"numeric"}\hlstd{,}\hlstr{"numeric"}\hlstd{,}\hlstr{"numeric"}\hlstd{,}\hlstr{"numeric"}\hlstd{,}\hlstr{"numeric"}\hlstd{,}\hlstr{"integer"}\hlstd{,}\hlstr{"numeric"}\hlstd{)}
    \hlkwa{if}\hlstd{(task.type}\hlopt{==}\hlstr{"classif"}\hlstd{)\{}
      \hlstd{fixpars} \hlkwb{<-} \hlkwd{list}\hlstd{(}\hlkwc{eval_metric}\hlstd{=}\hlstr{"logloss"}\hlstd{,}\hlcom{# suppress warning given when default metric is used.}
                    \hlkwc{nthread}\hlstd{=}\hlnum{1}\hlstd{)} \hlcom{#one thread, not parallel}
    \hlstd{\}}\hlkwa{else}\hlstd{\{}
      \hlstd{fixpars} \hlkwb{<-} \hlkwd{list}\hlstd{(}\hlkwc{eval_metric}\hlstd{=}\hlstr{"rmse"}\hlstd{,}\hlcom{# suppress warning given when default metric is used.}
                    \hlkwc{nthread}\hlstd{=}\hlnum{1}\hlstd{)} \hlcom{#one thread, not parallel}
    \hlstd{\}}
    \hlstd{factorlevels} \hlkwb{<-} \hlkwd{list}\hlstd{()}
    \hlstd{transformations} \hlkwb{<-} \hlkwd{c}\hlstd{(trans_2pow_round,} \hlcom{## differs from Prob19a}
                         \hlstd{trans_2pow,}
                         \hlstd{trans_2pow,}
                         \hlstd{trans_2pow,}
                         \hlstd{trans_id,}
                         \hlstd{trans_id,}
                         \hlstd{trans_2pow,} \hlcom{## Thom18a}
                         \hlstd{trans_id,}
                         \hlstd{trans_2pow)}
    \hlstd{dummy}\hlkwb{=}\hlnum{TRUE}
    \hlstd{relpars} \hlkwb{<-} \hlkwd{list}\hlstd{()}
\end{alltt}
\end{kframe}
\end{knitrout}

The imputation of missing values can be implemented as follows.

\begin{knitrout}\footnotesize
\definecolor{shadecolor}{rgb}{0.969, 0.969, 0.969}\color{fgcolor}\begin{kframe}
\begin{alltt}
\hlstd{task} \hlkwb{<-} \hlkwd{impute}\hlstd{(task,} \hlkwc{classes} \hlstd{=} \hlkwd{list}\hlstd{(}\hlkwc{factor} \hlstd{=} \hlkwd{imputeMode}\hlstd{(),} \hlkwc{integer} \hlstd{=} \hlkwd{imputeMedian}\hlstd{(),}
    \hlkwc{numeric} \hlstd{=} \hlkwd{imputeMean}\hlstd{()))}\hlopt{$}\hlstd{task}
\end{alltt}
\end{kframe}
\end{knitrout}

All factor features will be replaces with their dummy variables. 
Internally model.matrix is used. Non-factor features will be left untouched and passed to the result.

\begin{knitrout}\footnotesize
\definecolor{shadecolor}{rgb}{0.969, 0.969, 0.969}\color{fgcolor}\begin{kframe}
\begin{alltt}
\hlkwa{if} \hlstd{(dummy) \{}
    \hlstd{task} \hlkwb{<-} \hlkwd{createDummyFeatures}\hlstd{(task)}
\hlstd{\}}
\hlkwd{str}\hlstd{(task)}
\end{alltt}
\begin{verbatim}
## List of 6
##  $ type       : chr "classif"
##  $ env        :<environment: 0x7fe0f65be088> 
##  $ weights    : NULL
##  $ blocking   : NULL
##  $ coordinates: NULL
##  $ task.desc  :List of 13
##   ..$ id                : chr "dataOML"
##   ..$ type              : chr "classif"
##   ..$ target            : chr "V42"
##   ..$ size              : int 299285
##   ..$ n.feat            : Named int [1:4] 409 0 0 0
##   .. ..- attr(*, "names")= chr [1:4] "numerics" "factors" "ordered" "functionals"
##   ..$ has.missings      : logi FALSE
##   ..$ has.weights       : logi FALSE
##   ..$ has.blocking      : logi FALSE
##   ..$ has.coordinates   : logi FALSE
##   ..$ class.levels      : chr [1:2] " - 50000." " 50000+."
##   ..$ positive          : chr " - 50000."
##   ..$ negative          : chr " 50000+."
##   ..$ class.distribution: 'table' int [1:2(1d)] 280717 18568
##   .. ..- attr(*, "dimnames")=List of 1
##   .. .. ..$ : chr [1:2] " - 50000." " 50000+."
##   ..- attr(*, "class")= chr [1:3] "ClassifTaskDesc" "SupervisedTaskDesc" "TaskDesc"
##  - attr(*, "class")= chr [1:3] "ClassifTask" "SupervisedTask" "Task"
\end{verbatim}
\end{kframe}
\end{knitrout}

We can set the seed to improve reproducibility:
\begin{knitrout}\footnotesize
\definecolor{shadecolor}{rgb}{0.969, 0.969, 0.969}\color{fgcolor}\begin{kframe}
\begin{alltt}
\hlkwd{set.seed}\hlstd{(data.seed)}
\end{alltt}
\end{kframe}
\end{knitrout}

The information is compiled to a list.

\begin{knitrout}\footnotesize
\definecolor{shadecolor}{rgb}{0.969, 0.969, 0.969}\color{fgcolor}\begin{kframe}
\begin{alltt}
\hlstd{cfg} \hlkwb{<-} \hlkwd{list}\hlstd{(}\hlkwc{learner} \hlstd{= learner,} \hlkwc{tunepars} \hlstd{= tunepars,} \hlkwc{lower} \hlstd{= lower,} \hlkwc{upper} \hlstd{= upper,}
    \hlkwc{type} \hlstd{= type,} \hlkwc{fixpars} \hlstd{= fixpars,} \hlkwc{factorlevels} \hlstd{= factorlevels,} \hlkwc{transformations} \hlstd{= transformations,}
    \hlkwc{dummy} \hlstd{= dummy,} \hlkwc{relpars} \hlstd{= relpars,} \hlkwc{task} \hlstd{= task,} \hlkwc{resample} \hlstd{= rsmpl)}
\end{alltt}
\end{kframe}
\end{knitrout}

The objective function is defined: 
this function receives a configuration for a tuning experiment, 
and returns an objective function to be tuned via SPOT.

\begin{knitrout}\footnotesize
\definecolor{shadecolor}{rgb}{0.969, 0.969, 0.969}\color{fgcolor}\begin{kframe}
\begin{alltt}
\hlstd{objf} \hlkwb{<-} \hlkwd{get_objf}\hlstd{(}\hlkwc{config} \hlstd{= cfg,} \hlkwc{timeout} \hlstd{= timeout)}
\end{alltt}
\end{kframe}
\end{knitrout}

\subsection{\acl{xgb}: Tuning Run with SPOT}
The \gls{SPOT} tuning run can be started as follows.
\begin{knitrout}\footnotesize
\definecolor{shadecolor}{rgb}{0.969, 0.969, 0.969}\color{fgcolor}\begin{kframe}
\begin{alltt}
\hlstd{resultXGB} \hlkwb{<-} \hlkwd{spot}\hlstd{(}\hlkwc{fun} \hlstd{= objf,}
               \hlkwc{lower}\hlstd{=cfg}\hlopt{$}\hlstd{lower,}
               \hlkwc{upper}\hlstd{=cfg}\hlopt{$}\hlstd{upper,}
               \hlkwc{control} \hlstd{=} \hlkwd{list}\hlstd{(}\hlkwc{types}\hlstd{=cfg}\hlopt{$}\hlstd{type,}
                              \hlkwc{maxTime} \hlstd{= timebudget}\hlopt{/}\hlnum{60}\hlstd{,} \hlcom{#convert to minutes}
                              \hlkwc{plots}\hlstd{=}\hlnum{TRUE}\hlstd{,}
                              \hlkwc{progress} \hlstd{=} \hlnum{TRUE}\hlstd{,}
                              \hlkwc{model}\hlstd{=buildKriging,}
                              \hlkwc{optimizer}\hlstd{=optimDE,}
                              \hlkwc{noise}\hlstd{=}\hlnum{TRUE}\hlstd{,}
                              \hlkwc{seedFun}\hlstd{=}\hlnum{123}\hlstd{,}
                              \hlkwc{seedSPOT}\hlstd{=tuner.seed,}
                              \hlkwc{designControl}\hlstd{=}\hlkwd{list}\hlstd{(}\hlkwc{size}\hlstd{=}\hlnum{5}\hlopt{*}\hlkwd{length}\hlstd{(cfg}\hlopt{$}\hlstd{lower)),}
                              \hlkwc{funEvals}\hlstd{=}\hlnum{Inf}\hlstd{,}
                              \hlkwc{modelControl}\hlstd{=}\hlkwd{list}\hlstd{(}\hlkwc{target}\hlstd{=}\hlstr{"y"}\hlstd{,}
                                                \hlkwc{useLambda}\hlstd{=}\hlnum{TRUE}\hlstd{,}
                                                \hlkwc{reinterpolate}\hlstd{=}\hlnum{TRUE}\hlstd{),}
                              \hlkwc{optimizerControl}\hlstd{=}\hlkwd{list}\hlstd{(}\hlkwc{funEvals}\hlstd{=}\hlnum{100}\hlopt{*}\hlkwd{length}\hlstd{(cfg}\hlopt{$}\hlstd{lower))}
               \hlstd{)}
\hlstd{)}
\hlkwd{save}\hlstd{(resultXGB,} \hlkwc{file}\hlstd{=}\hlstr{"xgbResult.RData"}\hlstd{)}
\end{alltt}
\end{kframe}
\end{knitrout}

\begin{knitrout}\footnotesize
\definecolor{shadecolor}{rgb}{0.969, 0.969, 0.969}\color{fgcolor}\begin{kframe}
\begin{alltt}
\hlkwd{load}\hlstd{(}\hlkwc{file} \hlstd{=} \hlstr{"xgbResult.RData"}\hlstd{)}
\hlstd{result} \hlkwb{<-} \hlstd{resultXGB}
\end{alltt}
\end{kframe}
\end{knitrout}

\subsection{\acl{xgb}: Visualization of the Results}

First, we visualize the progress of the \gls{SPOT} tuning procedure.

\begin{knitrout}\footnotesize
\definecolor{shadecolor}{rgb}{0.969, 0.969, 0.969}\color{fgcolor}\begin{kframe}
\begin{alltt}
\hlstd{size} \hlkwb{<-} \hlnum{5} \hlopt{*} \hlkwd{length}\hlstd{(cfg}\hlopt{$}\hlstd{lower)}
\hlkwd{plot}\hlstd{(result}\hlopt{$}\hlstd{y,} \hlkwc{type} \hlstd{=} \hlstr{"l"}\hlstd{)}
\hlkwd{abline}\hlstd{(}\hlkwc{v} \hlstd{= size,} \hlkwc{col} \hlstd{=} \hlstr{"red"}\hlstd{)}
\hlstd{y0} \hlkwb{<-} \hlkwd{min}\hlstd{(result}\hlopt{$}\hlstd{y[}\hlnum{1}\hlopt{:}\hlstd{size])}
\hlkwd{abline}\hlstd{(}\hlkwc{h} \hlstd{= y0,} \hlkwc{col} \hlstd{=} \hlstr{"blue"}\hlstd{)}
\end{alltt}
\end{kframe}\begin{figure}

{\centering \includegraphics{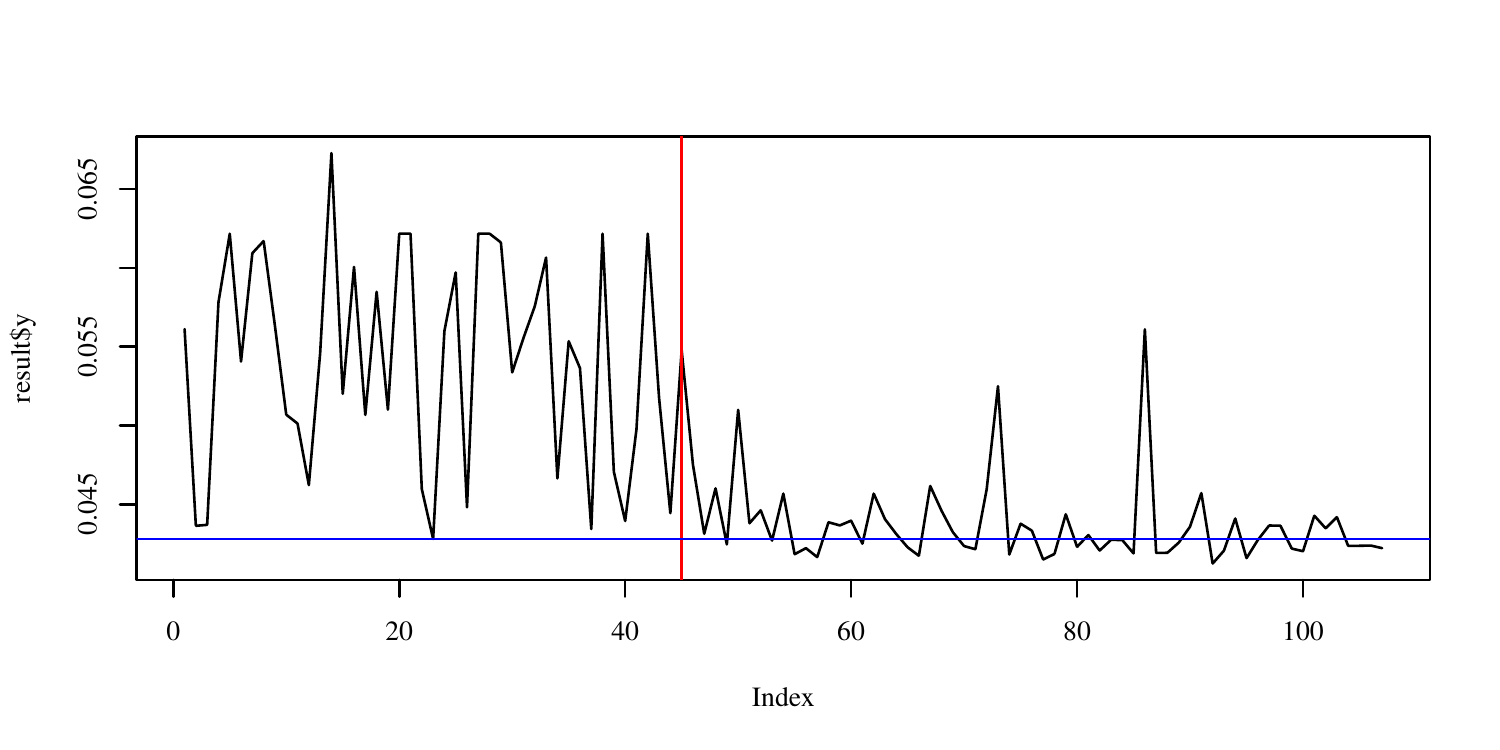} 

}

\caption[Case Study II]{Case Study II: SPOT search. The vertical red line denotes the end of the initial design phase. The best result from the (random) initial phase is shown in blue.}\label{fig:plotSpot02}
\end{figure}

\end{knitrout}

The best hyperparameter setting can be printed as follows:
\begin{knitrout}\footnotesize
\definecolor{shadecolor}{rgb}{0.969, 0.969, 0.969}\color{fgcolor}\begin{kframe}
\begin{alltt}
\hlstd{(paraNames} \hlkwb{<-} \hlstd{cfg}\hlopt{$}\hlstd{tunepars)}
\end{alltt}
\begin{verbatim}
## [1] "nrounds"          "eta"              "lambda"           "alpha"           
## [5] "subsample"        "colsample_bytree" "gamma"            "max_depth"       
## [9] "min_child_weight"
\end{verbatim}
\begin{alltt}
\hlkwd{print}\hlstd{(result}\hlopt{$}\hlstd{xbest)}
\end{alltt}
\begin{verbatim}
##         [,1]      [,2]     [,3]       [,4]      [,5]      [,6]      [,7] [,8]
## [1,] 8.22029 -3.009606 3.767863 -0.5625179 0.9684651 0.6746676 -2.636762    9
##          [,9]
## [1,] 2.756446
\end{verbatim}
\end{kframe}
\end{knitrout}

First plot, based on surrogate:

\begin{knitrout}\footnotesize
\definecolor{shadecolor}{rgb}{0.969, 0.969, 0.969}\color{fgcolor}\begin{kframe}
\begin{alltt}
\hlkwd{plot_surface}\hlstd{(result,} \hlkwc{yvar} \hlstd{=} \hlnum{1}\hlstd{)}
\end{alltt}
\end{kframe}
\end{knitrout}

\begin{figure}
\centering
\includegraphics[width=0.5\linewidth]{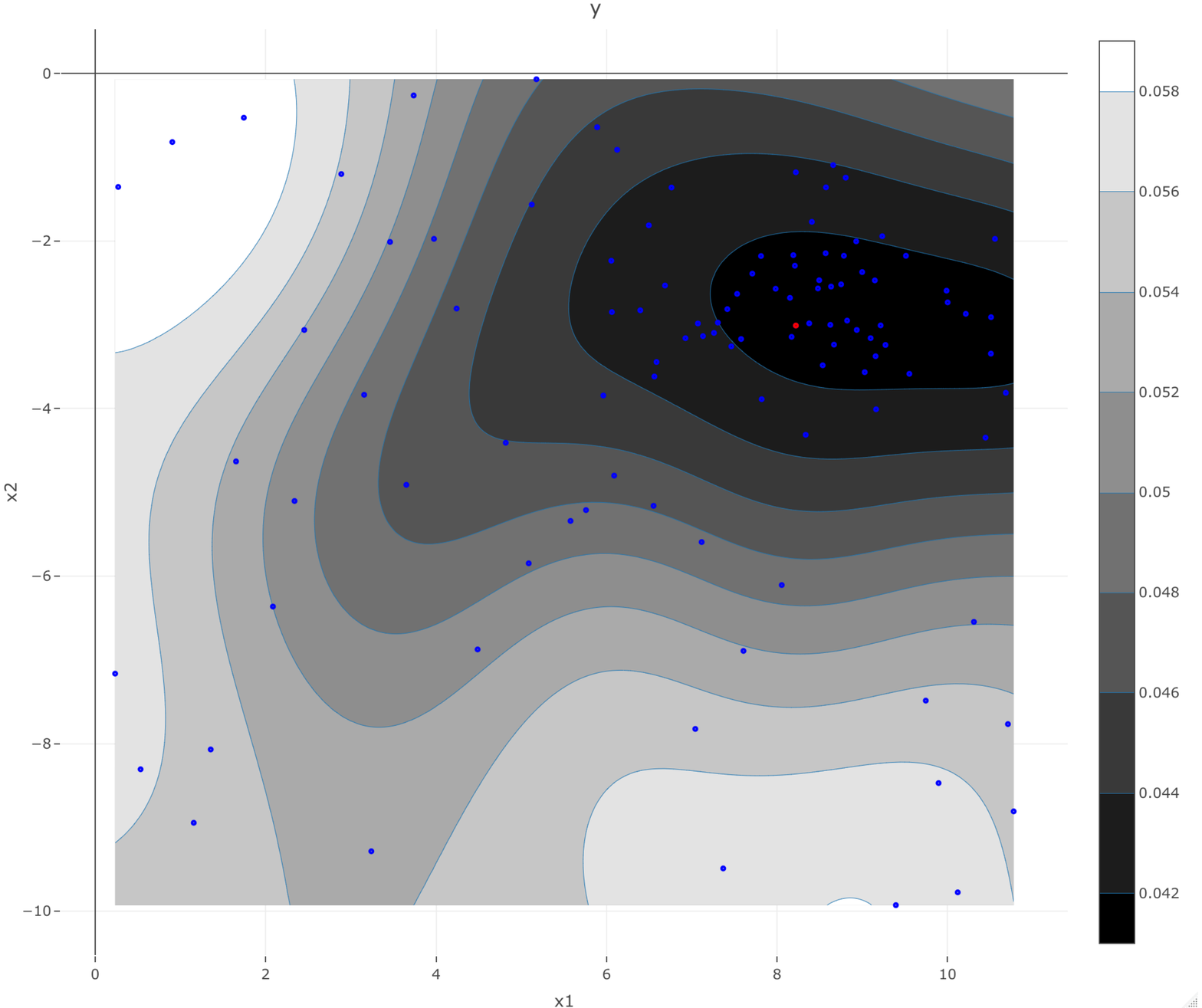}
\caption{mce versus \paramnrounds and \parameta. Values of $x_1$ (nrounds) are in the range from $2^0$ to $2^{10}$ and values of $x_2$ (eta) in the range from $2^{-10}$ to $2^0$.}
\label{fig:xgbSpotResult01}
\end{figure}

\begin{figure}
\centering
\includegraphics[width=0.5\linewidth]{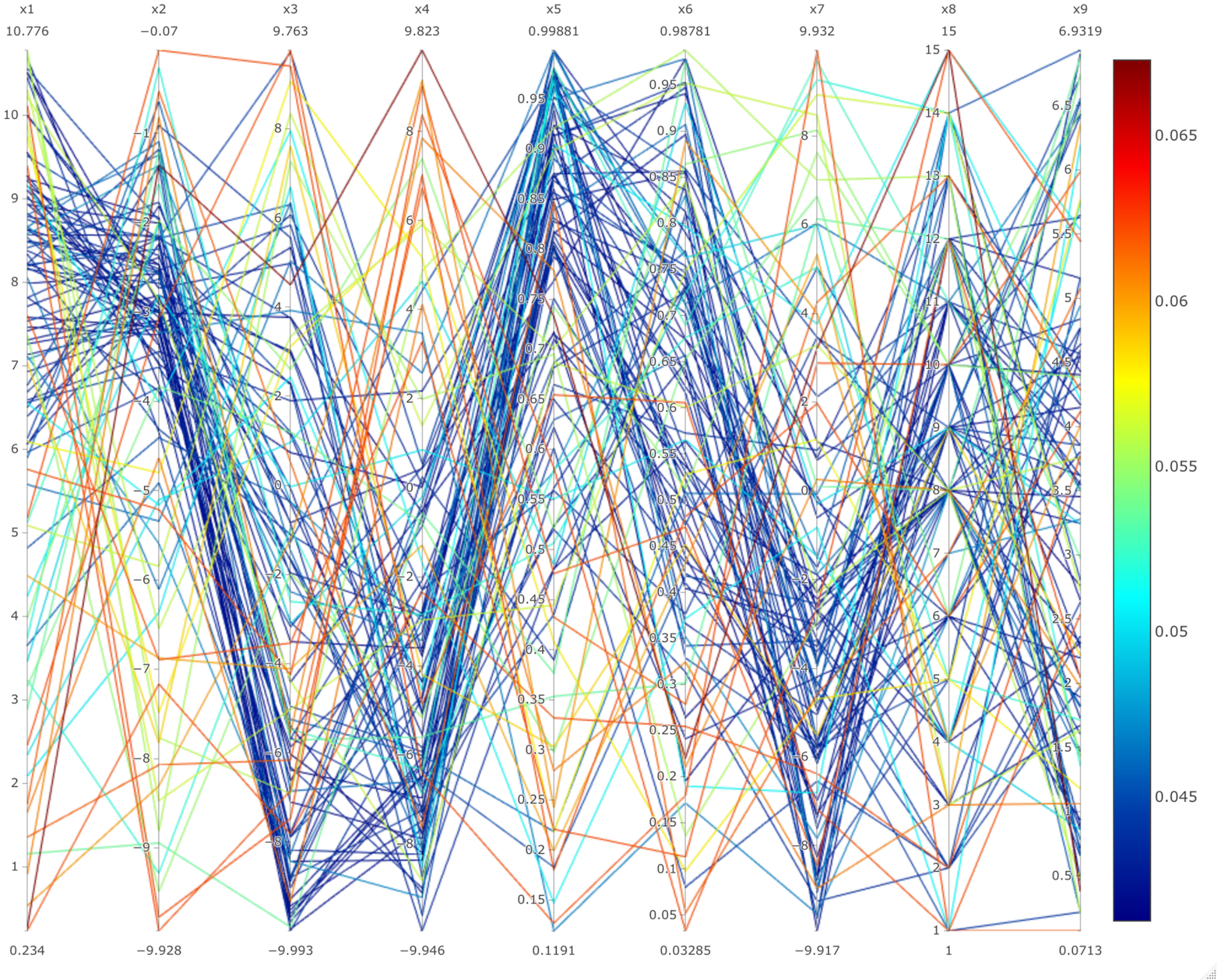}
\caption{mce: parallel plot. Red lines denote poor configurations, blue lines better configurations. This plot indicates that $x_1$ and $x_5$ should be large whereas $x_7$ values should be low.}
\label{fig:xgbSpotResult02}
\end{figure}

Next, mce vs all parameters  (based on evaluations) is shown in Fig.~\ref{fig:xgbSpotResult02}.

\begin{knitrout}\footnotesize
\definecolor{shadecolor}{rgb}{0.969, 0.969, 0.969}\color{fgcolor}\begin{kframe}
\begin{alltt}
\hlkwd{plot_parallel}\hlstd{(resultXGB,} \hlkwc{yvar} \hlstd{=} \hlnum{1}\hlstd{)}
\end{alltt}
\end{kframe}
\end{knitrout}

In addition, a sensitivity plot can be shown.

\begin{knitrout}\footnotesize
\definecolor{shadecolor}{rgb}{0.969, 0.969, 0.969}\color{fgcolor}\begin{kframe}
\begin{alltt}
\hlkwd{plot_sensitivity}\hlstd{(result,} \hlkwc{type} \hlstd{=} \hlstr{"agg"}\hlstd{)}
\end{alltt}
\end{kframe}
\end{knitrout}

\begin{figure}
\centering
\includegraphics[width=0.5\linewidth]{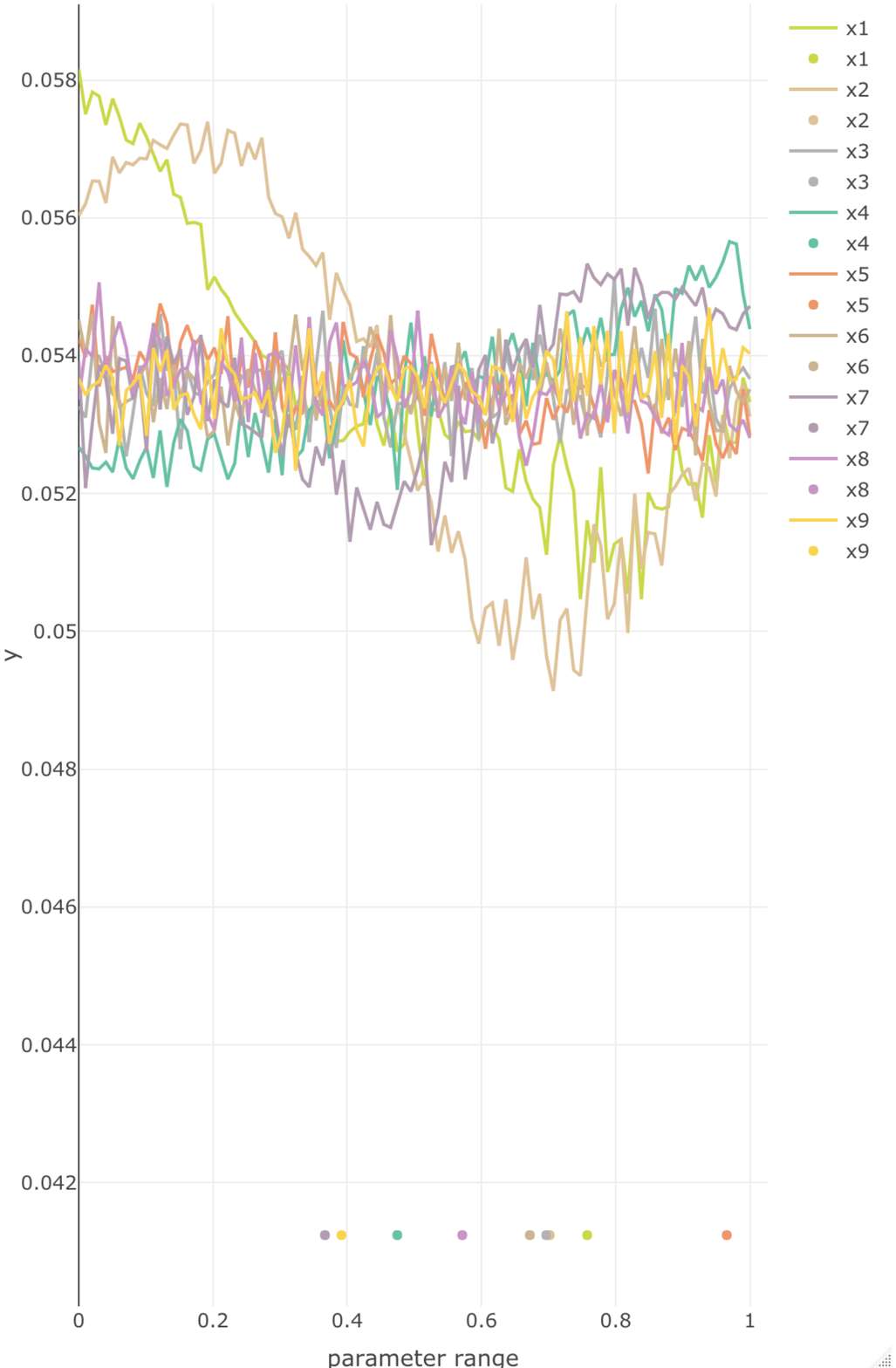}
\caption{Sensitivity plot. Effect of normalized hyperparamter values. $x_1$ and $x_2$ appear to be the most important parameters.}
\label{fig:xgbSpotTuning03}
\end{figure}

Finally, we fit a simple regression tree to analyze the most important
hyperparameters.

\begin{knitrout}\footnotesize
\definecolor{shadecolor}{rgb}{0.969, 0.969, 0.969}\color{fgcolor}\begin{kframe}
\begin{alltt}
\hlkwd{library}\hlstd{(}\hlstr{"rpart"}\hlstd{)}
\hlkwd{library}\hlstd{(}\hlstr{"rpart.plot"}\hlstd{)}
\hlstd{fitTree} \hlkwb{<-} \hlkwd{buildTreeModel}\hlstd{(}\hlkwc{x} \hlstd{= result}\hlopt{$}\hlstd{x,} \hlkwc{y} \hlstd{= result}\hlopt{$}\hlstd{y,} \hlkwc{control} \hlstd{=} \hlkwd{list}\hlstd{(}\hlkwc{xnames} \hlstd{= paraNames))}  \hlcom{# paste0('x', 1:n)))}
\hlkwd{rpart.plot}\hlstd{(fitTree}\hlopt{$}\hlstd{fit)}
\end{alltt}
\end{kframe}

{\centering \includegraphics{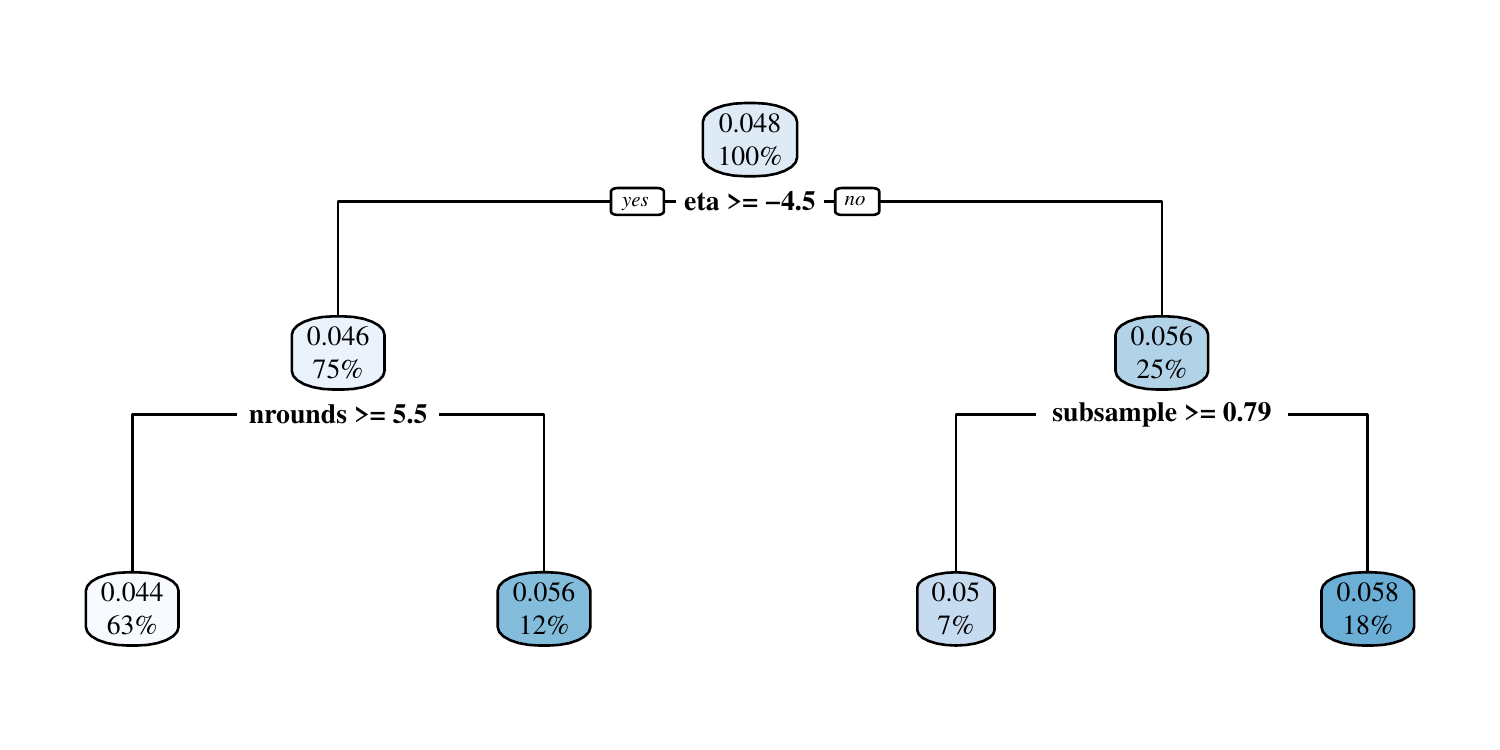} 

}

\end{knitrout}

Our visualizations are based on the transformed values.

\subsection{\acl{xgb}: Summary}

Data visualizations shown in Sec.~\ref{sec:visRpart} indicate that the hyperparameters
\parameta  and \paramnrounds have the greatest effect on the algorithm's performance.
The recommended values are 
$ \paramnrounds \approx 2^8 = 256$ and $ \parameta \approx 2^{-3} = 0.125$.

\section{Case Study III: Global Experimental Investigation}\label{sec:case3}

\subsection{Research Questions}\label{sec:frage}
As described in  Sec.~~\ref{ch:tuning}, \gls{SPOT} offers a robust approach for the tuning of \ac{ml} algorithms, 
especially if the training and/or evaluation runtime become large.

In practice, the learning process of models and hence the choice of their hyperparameters is influenced 
by a plethora of other factors.
On the one hand, this complex situation further motivates the use of tuning procedures,
since the \ac{ml} algorithms have to be adapted to new data or situations.
On the other hand, this raises the question of how such factors influence the tuning process itself.
We want to investigate this in a structured manner.

In detail, the following factors are objects of our investigation, testing
their influence on the tuning procedures.
\begin{itemize}
\item Number of numerical features in the data (\nnum).
\item Number of categorical features in the data (\nfac).
\item Cardinality of the categorical features, i.e., the maximal number of levels (\card).
\item Number of observations in the data ($m$).
\item Task type, classification or regression.
\item Choice of model.
\end{itemize}

We want to investigate the following hypotheses.

\begin{enumerate}[(H-1)]
\item Tuning is necessary to find good parameter values (compared to defaults).
\item Data: Properties of the data influence (the difficulty of) the tuning process.
\begin{itemize}
\item Information content: If the data has little information content, models are easier to tune,
since more parameter configurations achieve near-optimal quality.
In general, changing parameters has less impact on model quality in this case.
\item Number of features: 
A larger number of features leads to longer runtimes, which affects how many
evaluations can be made during tuning.
\item Type of features:
The number of numerical and / or categorical features as well as their cardinality
influences how much information is available to the model,
hence may affect the difficulty of tuning.
\item Number of observations $m$: 
With increasing $m$, the average runtime of model evaluations will increase.
\end{itemize}
\item Target variable: 
There is no fundamental difference between tuning regression or classification models.
\item Model: 
The choice of model (e.g., \ac{en} or \ac{svm}) 
affects the difficulty of the tuning task, but not necessarily the choice of 
tuning procedure.
\item Benchmark: 
The aptitude of the employed tuners can be measured in a statistically sound manner.
\end{enumerate}

\subsection{Setup}

\subsubsection{Generation of Data Sets}
For this investigation we choose a slighly modified version of the \gls{CID} data set, 
which was introduced in Sec.~\ref{sec:census}. 
This data set suits well to our research questions, 
as it is comparatively large,
has many categorical features,
several of the categorical features have a lot of levels,
and it can be easily used to generate different classification
as well as regression problems.

To investigate our research questions, the data set needs to be pre-processed
accordingly. 
This pre-processing is documented in the included source code,
and is briefly explained in the following.

\begin{itemize}
\item Feature 24 (instance weight MARSUPWT) is removed.
This feature describes the number of persons in the population who are
represented by the respective observation.
This is relevant for data understanding, but should not be an input
to the machine learning models.
\item Several features are encoded as numerical(integer) variables,
but are in fact categorical.
Example: Feature 3 (industry code ADTIND) is encoded as an integer.
Since the respective integers represent discrete codes for different sectors
of industry, they have no inherent order and should be encoded as categorical features.
\item The data set sometimes contains \NA values (missing data). 
These \NA values are replaced before modeling.
For categorical features, the most frequently observed category is imputed (mode). 
For integer features, the median is imputed, and for real-valued features the mean.
\item 
As the only model in the investigation, \modxgb is not able to work directly
with categorical features.
In that case (only for \modxgb) the categorical data features are transferred into
a dummy-coding.
For each category of the categorical feature, a new binary feature is created,
which specifies whether an observation is of the respective category or not.
\item Finally, we split the data randomly into test data
(40\% of the observations) 
and training data (60\%).
\end{itemize}

In addition to these general pre-processing steps, we change properties
of the data set for individual experiments, to cover our various hypotheses.

Arguably, we could have done this by using completely different data sets
where each covers different objects of investigation (i.e., different
numbers of features or different $m$).
We decided to vary a single data set instead, generating new data sets
with different properties, because this allows us to reasonably
compare results between the individual variations.
This way, we generate multiple data sets that cover different aspects and
problems in detail. 
While they all derive from the same data set (\gls{CID}), they all have different
characteristics:
Number of categorical features, number of numerical features, cardinality,
number of observations, and target variable.
These characteristics can be quantified with respect to difficulty
as discussed in  Sec.~~\ref{sec:schwierig}.

In detail, we vary:
\begin{itemize}
\item Target variable: 
The original target variable of the data set is the income class (below / above 50 000 USD).
We choose \textit{age} as the target variable instead.
For classification experiments, age will be discretized, 
into two classes: age $< 40$ and age $>= 40$.
For regression, age remains unchanged.
This choice intends to establish comparability between both experiment groups (classification, regression).
\item Number of categories (\texttt{cardinality}): 
To create variants of the data set with different cardinality of categorical features,
we merge categories into new, larger categories.
For instance, for feature 35 (country of birth self PENATVTY) 
the country of origin is first merged by combining all countries from a specific
continent.
This reduces the cardinality, with 6 remaining categories (medium cardinality).
For a further reduction (low cardinality) to three categories, the data is merged into
the categories unknown, US, and abroad.
Similar changes to other features are documented in the source code.
For our experiments, this pre-processing step results into data sets
with the levels of cardinality: low (up to 15 categories),
medium (up to 24 categories) and high (up to 52 categories).
\item 
Number of numerical features (\texttt{nnumericals}): 
To change the number of features, individual features are included or removed
from the data set.
This is done separately for categorical and numerical features
and results into four levels for \texttt{nnumericals} (low: 0, medium: 4, high: 6, complete: 7).
\item Number of categorical features (\texttt{nfactors}): 
Correspondingly, we receive four levels for \texttt{nfactors} (low: 0, medium: 8, high: 16, complete: 33).
Note, that these numbers become somewhat reduced, if cardinality is low (low: 0, medium: 7, high: 13, complete: 27).
The reason is, that some features might become redundant when merging categories.
\item Number of observations ($m$): 
To vary $m$, observations are randomly sampled from the data set.
We test five levels on a logarithmic scale from $10^4$ to $10^5$: 10 000,  17 783,  31 623,  56 234, and 100 000.
In addition, we made a separate test with the complete data set, i.e., 299 285 observations.
\end{itemize}

For classification, we performed for each tuner and each model the experiments
summarized in Table~\ref{table:explan}.
A reduced number of experiments were performed for regression, see Table~\ref{table:explanreg}.

\begin{table}[ht]
\centering
\caption{Experiments for classification: Investigated combinations of the number of categorical features (nfactors), numerical features(nnumericals), cardinality, and $m$. 
An empty field for cardinality occurs for low nfactors.
Here, no categorical features are present, so the number of categories is irrelevant.
}
\begin{tabular}{llll}
  \hline
nfactors & nnumericals & cardinality & $m$ \\ 
  \hline
high & low & low & $10^4, 10^{4.25}, ...,10^5$ \\ 
  medium & medium & low & $10^4, 10^{4.25}, ...,10^5$ \\ 
  low & high &  & $10^4, 10^{4.25}, ...,10^5$ \\ 
  high & high & low & $10^4, 10^{4.25}, ...,10^5$ \\ 
  high & low & medium & $10^4, 10^{4.25}, ...,10^5$ \\ 
  medium & medium & medium & $10^4, 10^{4.25}, ...,10^5$ \\ 
  high & high & medium & $10^4, 10^{4.25}, ...,10^5$ \\ 
  high & low & high & $10^4, 10^{4.25}, ...,10^5$ \\ 
  medium & medium & high & $10^4, 10^{4.25}, ...,10^5$ \\ 
  high & high & high & $10^4, 10^{4.25}, ...,10^5$ \\ 
     \hline
    \multicolumn{3}{c}{complete data set }   & 299285 \\ 
   \hline
\end{tabular}
\label{table:explan}
\end{table}

\begin{table}[ht]
\centering
\caption{Experiments for regression: Investigated combinations of the number of categorical features (nfactors), numerical features(nnumericals), cardinality, and $m$. 
An empty field for cardinality occurs for low nfactors.
Here, no categorical features are present, so the number of categories is irrelevant.
}
\begin{tabular}{llll}
  \hline
nfactors & nnumericals & cardinality & $m$ \\ 
  \hline
low & high &  & $10^4, 10^{4.25}, ...,10^5$ \\ 
  high & high & low & $10^4, 10^{4.25}, ...,10^5$ \\ 
  high & high & high & $10^4, 10^{4.25}, ...,10^5$ \\ 
   \hline
\end{tabular}
\label{table:explanreg}
\end{table}

To judge the impact of the creating random subsets of the data (i.e., to reduce $m$),
and to consider the test / train split, 
three data sets are generated for each configuration.
All experiments are repeated for each of those three configurations.

\subsubsection{Model Configuration}
The experiments include the models \ac{knn}, \ac{eb}, \ac{en}, \ac{rf}, \ac{xgb}, and \ac{svm}.
Their respective parameters are listed in Table~\ref{table:survey}).

For \ac{en}, \paramlambda is not optimized by the tuner.
Rather, the \modglmnet implementation itself tunes that parameter.
Here, a sequence of different \paramlambda values is tested~\citep{Hast16a,Hast20b}.

For \ac{svm}, the choice of the kernel (\paramkernel) 
is limited to \texttt{radial} and \texttt{sigmoid},
since we experienced surprisingly large runtimes for 
\texttt{linear} and \texttt{polynomial} 
in a set of preliminary experiments.
Hence, \paramdegree is also excluded, as it is only relevant for the \texttt{polynomial} kernel.
Due to experiment runtime, we also did not perform experiments with \ac{svm} and \ac{knn} on data sets with $m\geq 10^5$ observations.
These would require using model variants that are able to deal with huge data sets (e.g., some sparse SVM type).

Table~\ref{table:hyperExp} lists hyperparameters that are actually tuned in the experiments,
including data type, bounds and employed transformations.
Here, we mostly follow the bounds and transformations as used by \citet{Prob19a}.
Fundamentally, these are not general suggestions.
Rather, reasonable bounds will usually require some considerations with respect
to data understanding, modeling / analysis, and computational resources.
Bounds on values which affect run time should be chosen so that
experiments are still possible within a reasonable time frame.
Similar consideration can apply to memory requirements
Where increasing / decreasing parameters may lead to increasing / decreasing
sensitivity of the model, a suitable transformation (e.g., log-transformation)
should be applied.

Most other configurations of the investigated models remain at default values.
The only exceptions are:
\begin{itemize}
\item \modranger: 
For the sake of comparability with other models, model training and evaluation are performed
in a single thread, without parallelization (\texttt{num.threads} = 1).
\item \modxgb: Similarly to\modranger,we set \texttt{nthread}=1.
For regression, the evaluation metric is set to the root-mean-square error
(\verb|eval_metric="rmse"|).
For classification, log-loss is chosen (\verb|eval_metric="logloss"|).
\end{itemize}

\subsubsection{Further Configurations}
Overall, the implementation of our experiments is as follows:
The R package \packmlr is used as a uniform interface to the machine learning models. 
The R package \packSPOT is used to perform the actual tuning (optimization).
All additional code is provided together with this paper.
We also provide examples for creating visualizations of the tuning results.

The following configurations are of importance:

\begin{itemize}
\item General configuration: 
\begin{itemize}
\item Model quality measures: 
The evaluation of hyperparameter values requires a measure of quality,
which determines how well the resulting models perform.
\begin{itemize}
\item Regression: For the regression experiments, we use \ac{rmse}:
\ac{rmse}$=\sqrt{(1/m)\sum_{i=1}^m (y_i-\yhat_i)^2}$.
Here, $\yhat_i$ is the predicted value of the corresponding model for the i-th observation in the data set, and $y_i$ is the actually observed value.
\item Classification: 
For the classification experiments, we use \ac{mmce}:\\ \ac{mmce}$=(1/m)\sum_{i=1}^m \mathbf{I}(y_i\neq\yhat_i)$.
\end{itemize}
The tuner uses these two measures to determine better hyperparameter values.
\item Further measures:
In addition to \ac{rmse} and \ac{mmce}, we also record 
runtime (Overall runtime of a model evaluation, runtime for prediction, runtime for training).
Additional performance measure from the \texttt{mlr} package could be easily integrated when necessary\footnote{
A complete list of measures in \texttt{mlr} can be found at \url{https://mlr.mlr-org.com/articles/tutorial/measures.html}}.
\item Runtime budget: 
To mirror a realistic use case, we specify a fixed
runtime budget for the tuner.
This limits how long the tuner may take for finding potentially optimal hyperparameter values.
We set a budget of 5 hours for \ac{svm}, \ac{knn}, \ac{rf}, and \ac{xgb}.
Since \ac{en} and \ac{eb} are much faster to evaluate and less complex,
they receive a considerably lower budget (\ac{en}: 1 hour, \ac{eb}: 5 minutes).
\item Timeout: 
For a majority of the models, the runtime of a single evaluation (training + prediction)
is hard to predict and may easily become excessive if parameters are chosen poorly.
In extreme cases, the runtime of a single evaluation may become so large that
it takes up the majority of tuner runtime, or even more.
In such a case, there would be insufficient time to test different hyperparameter values.
To prevent this, we specify a limit for the runtime of a single evaluation, which we call timeout.
If the timeout is exceeded by the model, the evaluation will be aborted.
During the experiments we set the timeout to a twentieth of the tuner's overall runtime budget.
Exceptions are the experiments with \ac{eb} (\modrpart): 
Since \modrpart evaluates extremely quickly, (in our experiments: usually much less than a second)
and has a correspondingly reduced runtime budget (5 minutes), 
the timeout is not required.
In fact, using the timeout would add considerable overhead to the evaluation time in this case.
\item Returned quality values in case of errors:
If the evaluation gets aborted (e.g., due to timeout or in case of some numerical instability),
we still require a quality value to be returned to the tuner, so that the 
search can continue.
This return value should be chosen, so that, e.g., additional evaluations with high runtimes are avoided.
At the same time, the value should be on a similar scale as the actual quality measure,
to avoid a deterioration of the underlying surrogate model.
To achieve this, we return the following values when an evaluation aborts.
\begin{itemize}
\item Regression: Model quality for simply predicting the mean of the training data.
\item Classification: Model quality for simply predicting the mode of the training data.
\end{itemize}
\item Benchmark of runtime:
The experiments are not run on entirely identical hardware, but on somewhat diverse
nodes of a cluster.
Hence, we have to consider that single nodes are faster or slower than others
(i.e., a tuning run on one node may perform more model evaluations than on another node,
simply because of having more computational power).
To preserve some sort of comparability, we compute a corrective factor
that will be multiplied with the runtime budget.
Before each tuning run, we compute a short performance benchmark.
The time measured for that benchmark will be divided by the respective value
measured at a reference node, to determine the runtime multiplicator.
We use the \texttt{benchmark\_std} function from the \texttt{benchmarkme} R package~\citep{Gill21a}.
 \end{itemize}
\item Configuration of \ac{spot}:
\begin{itemize}
\item Design:
The initial design is created by Latin Hypercube Sampling~\citep{Lear03a}.
The size of that design (number of sampled configurations of hyperparameters)
corresponds to $5 d$. Here, $d$ is the number of hyperparameters. 
An exponential dependence such as $d^5$ would be possible in principle, but
might lead to an excessive size of the initial design
(e.g., just $d=4$ parameters yield $4^5=1024$ initial evaluations).
\item Surrogate model:
\gls{SPOT} can use arbitrary regression models as surrogates, e.g., random forest or Gaussian process models (Kriging). 
We chose a Gaussian process model, since it performs well in our experience,
and the Gaussian process implementation  can work well with discrete as well as
continuous hyperparameters.
Random forest is less suited for continuous parameters, as it has to approximate
them in a step-wise constant manner.

The Gaussian process model is fitted via Maximum Likelihood Estimation.
That is, using numerical optimization, the model parameters are determined,
so that the observed data are judged to have maximum likelihood according to the model.
We use Differential Evolution~\citep{Stor97a} for this purpose.

In one central issue, we deviate from more common configurations of surrogate
model based optimization algorithms:
For the determination of the next candidate solution to be evaluated, 
we directly use the predicted value of the Gaussian process model,
instead of the so-called expected improvement.
Our reason is, that the expected improvement may yield worse results if the number
of evaluations is low, or the dimensionality rather high~\citep{Rehb20a,Wess17a}.
With the strictly limited runtime budget, our experience is that
the predicted value is the better choice.
A similar observation is made by~\citet{Ath19a}.
\item Optimizing the predicted value:
To find the next candidate solution, the predicted value of the surrogate model
is optimized via Differential Evolution~\citep{Stor97a}\footnote{
We could use other global optimization algorithms as well.
Even random search would be a feasible strategy.
}.
Thus, \gls{SPOT} searches for the hyperparameter configuration that is predicted
to result into the best possible model quality..
We allow for $200 d$ surrogate evaluations in each iteration of \gls{SPOT}.
\item All other parameters of \ac{spot} remain at default values.
 \end{itemize}
\item Configuration of \ac{rs}:
\begin{itemize}
\item With \ac{rs}, hyperparameter values will be sample uniformly from the search space.
\item All related configurations (timeout, runtime budget, etc.) correspond to those of \gls{SPOT}.
 \end{itemize}
\item Defaults:
\begin{itemize}
\item 
As a comparison basis, we perform an additional experiment for each model,
 where all hyperparameter values remain at the models default settings.
\item However, in those cases we do \textit{not} 
set a timeout for evaluation.
Since no search takes place, the overall runtime for default values
is anyways considerable lower than the runtime of \gls{SPOT} or \ac{rs}.
\item All other settings correspond to those of \gls{SPOT} and \ac{rs}.
 \end{itemize}
\item Replications: 
To roughly estimate the variance of results, we repeat all experiments
(each tuner for each model on each data set) three times.
\end{itemize}

\subsection{Results}
In this  section, we provide an exploratory overview of the results of the experiments.
A detailed discussion of the results in terms of the research questions defined in Sec.~~\ref{sec:frage} follows in  Sec.~~\ref{ch:diskussion}.

To get an impression of the overall results, 
we show exemplary boxplots\footnote{Corresponding boxplots for all experiments can be found in the appendix of this document.}.
Since different results are achieved depending on the data set and optimized model,
a preprocessing step is performed first: For each drawn data set and each model the mean value of all observed results (model goodness of the best solutions found) is determined.
This mean is then subtracted from each individual observed value of the corresponding group.
Subsequently, these subtracted individual values are examined.
This allows a better visualization of the difference between the tuners without compromising interpretability. 
The resulting values are no longer on the original scale of the model quality, but the units remain unchanged. 
Thus, differences on this scale can still be interpreted well. 

\begin{figure}
\centering
\includegraphics[width=0.9\textwidth]{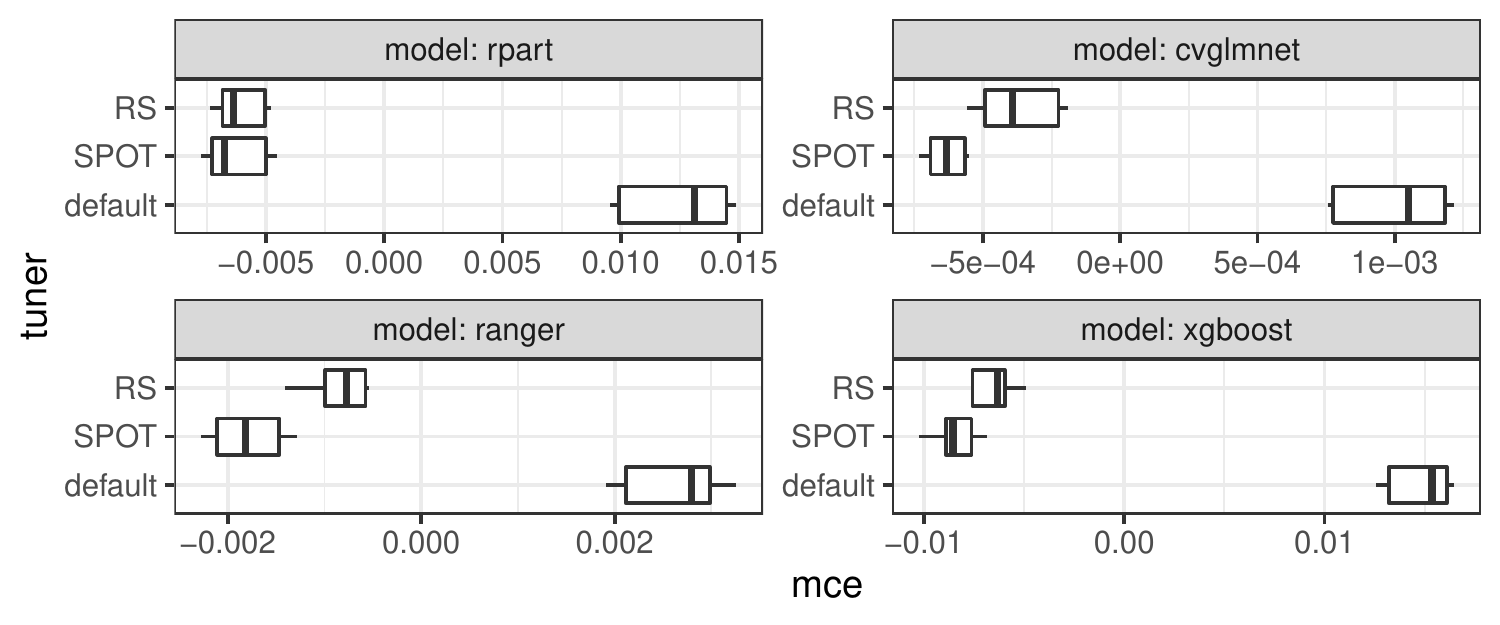}
\caption{Boxplot comparing tuners for different classification models, with $m=10^5$, nfactors=high, nnumericals=high, cardinality=high.
For this value of $m$, no experiments with \ac{knn} and \ac{svm} were performed.
The presented quality measure (mce) shows values after preprocessing, where the mean value for each problem instance is subtracted.
} \label{fig:box2}
\end{figure}

For the classification experiments, Figure~\ref{fig:box2} shows the results for a case with many features and observations (nfactors, nnumericals and cardinality are all set to high and $m=10^5$).
The figure first shows, 
that both tuners (\ac{rs}, \ac{spot}) achieve a significant improvement over default values.
The value of this improvement is in the range of about 1\% \ac{mmce}.
In almost all cases, the tuners show a smaller dispersion of the quality values than a model with default values. 
Except for the tuning of \modrpart, the results obtained by \gls{SPOT} are better than those of \ac{rs}.

\begin{figure}
\centering
\includegraphics[width=0.9\textwidth]{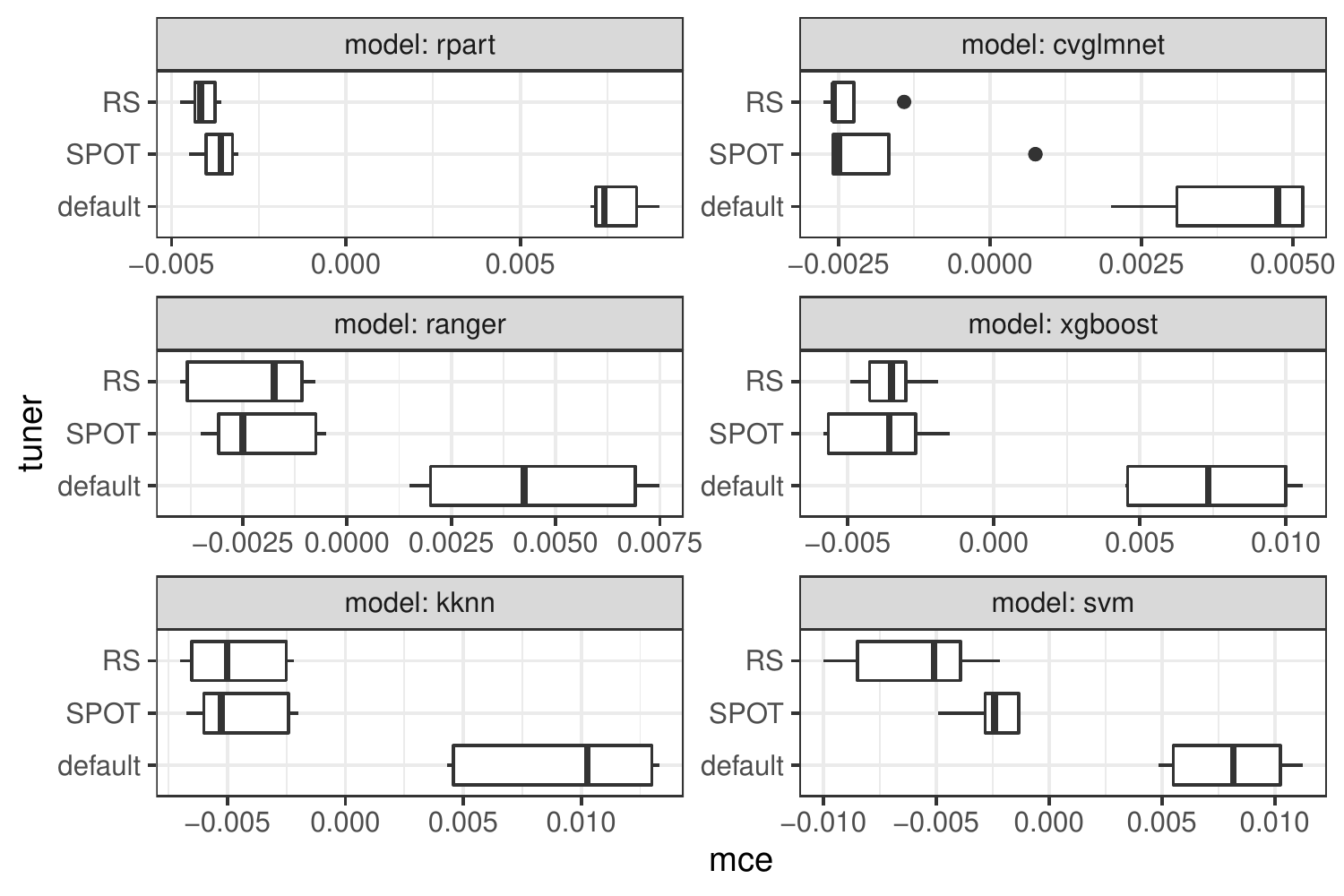}
\caption{Boxplot comparing tuners for different classification models, with $m=10^4$, nfactors=low, nnumericals=high.
The presented quality measure (mce) shows values after preprocessing, where the mean value for each problem instance is subtracted.
} \label{fig:box1}
\end{figure}

Likewise for classification, Figure~\ref{fig:box1} shows the case with $m=10^4$, without categorical features (nfactors=low) and with maximum number of numerical features (nnumericals=high).
Here the data set contains much less information, 
since a large part of the features is missing and only few observations are available.
Still, both tuners are significantly better than using default values.
In this case, it is mostly not clear which of the two tuners provides better results.
For \modrpart (\ac{eb}) and \ac{svm}, \ac{rs} seems to work better.

\begin{figure}
\centering
\includegraphics[width=0.9\textwidth]{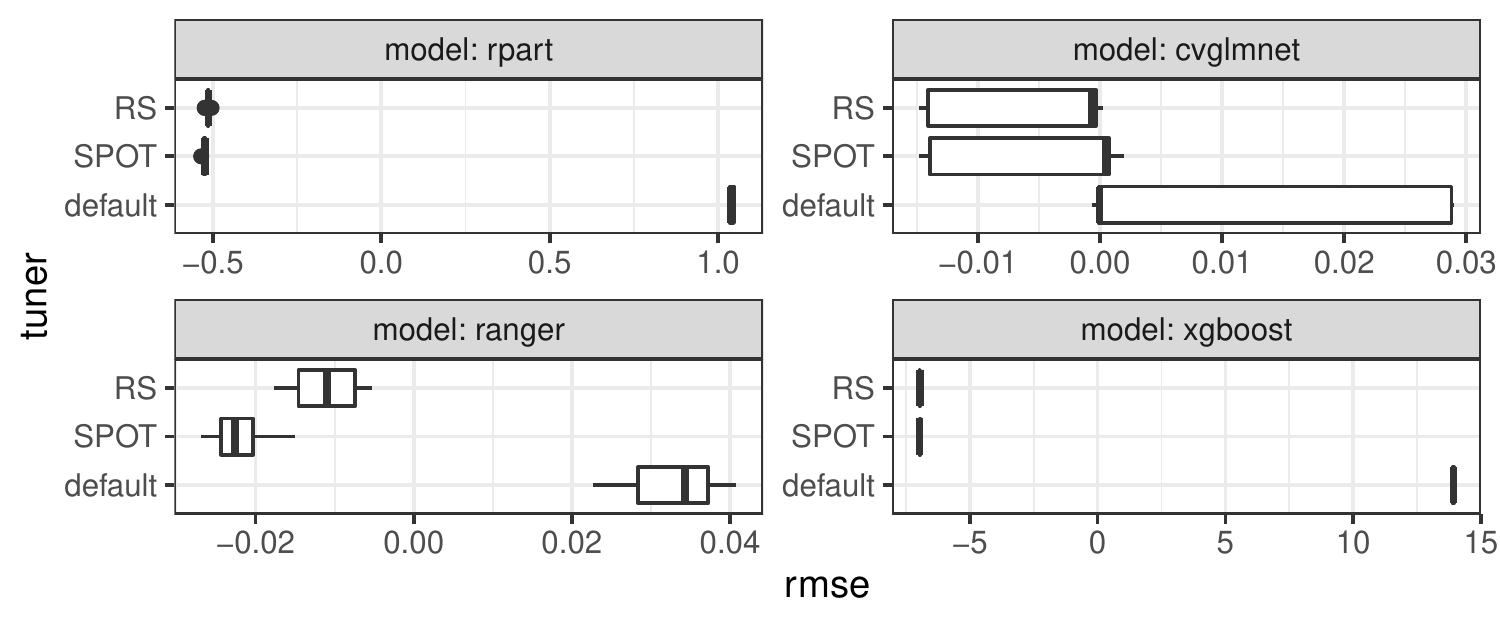}
\caption{Boxplot comparing tuners for different regression models, with $m=10^5$, nfactors=high, nnumericals=high, cardinality=high.
For this value of $m$ no experiments with \ac{knn} and \ac{svm} were performed.
The presented quality measure (mce) shows values after preprocessing, where the mean value for each problem instance is subtracted.
} \label{fig:box4}
\end{figure}

\begin{figure}
\centering
\includegraphics[width=0.9\textwidth]{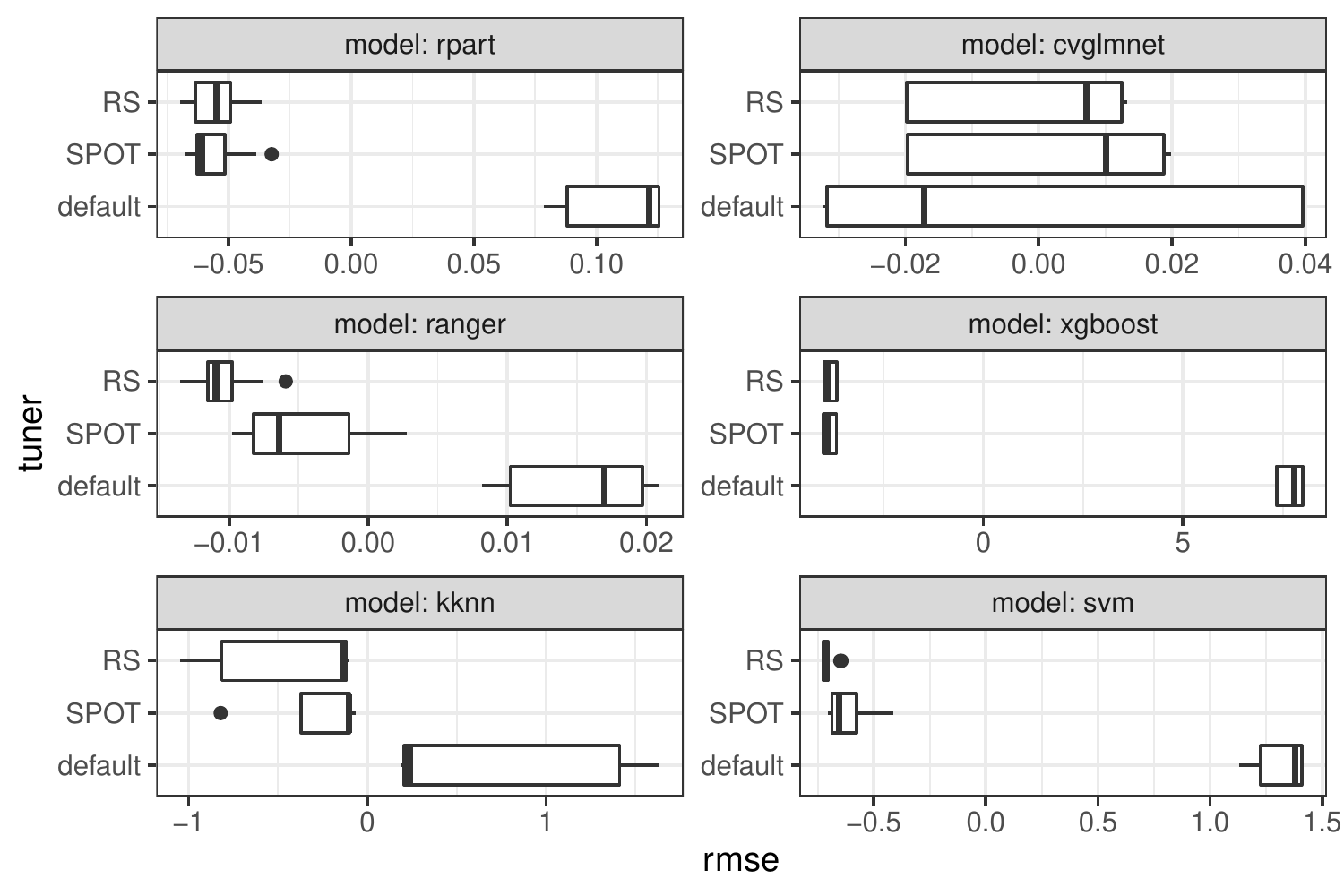}
\caption{Boxplot comparing tuners for different regression models, with $m=10^4$, nfactors=low, nnumericals=high.
The presented quality measure (mce) shows values after preprocessing, where the mean value for each problem instance is subtracted.
} \label{fig:box3}
\end{figure}

In the same form as for classification, figures \ref{fig:box4} and \ref{fig:box3} show results for regression models.
Unlike for classification, the results here are somewhat more diverse. 
In particular, \modglmnet shows a small difference between the tuners and default values.
There are also differences of several orders of magnitude between the individual models (e.g. \ac{rf} and \ac{svm}).
For example, for \ac{rf} the differences between tuners and default values are about 0.02 years (the target size is the age in years). 
As shown in figures \ref{fig:box4} and \ref{fig:box3}, the interquartile range is about 0.01 years.

For \ac{xgb}, on the other hand, there is a difference of about 20 years between tuners and default values.
Here, the default values seem to be particularly poorly suited.

\subsection{Discussion}\label{ch:diskussion}

\subsubsection{Rank-Analysis}\label{sec:rankbas}
We analyze the results of the experiments using rankings of the result values instead of the raw results.
Rankings are scale invariant, so aggregating the results from different problem instances (resulting from data set and model choice) with different scales for the result values is possible.
This is important because for the analysis of individual study items we have to aggregate over very diverse problem instances.

To aggregate rankings (also called consensus rankings), we follow the so-called ``{}optimal ranking'' approach of~\citet{Keme59a}.
Here, the consensus ranking is determined such that the mean distance between the consensus ranking and observed ranking is minimal.
The distance measure used for this purpose is Kendall's tau~\citep{Kend38a}, which counts the number of pairwise comparisons in which two rankings contradict each other.
The ranking determined using this approach can be interpreted as the ``median'' of the individual rankings.

This procedure has the following advantages~\citep{Horn07a,Mers10a}:
\begin{itemize}
  \item Scale invariant
  \item Invariant to irrelevant alternatives.
  \item Aggregation of large sets of comparisons, over arbitrary factors.
  \item Easy/intuitive to interpret results \& visualizable
  \item Generates relevant information: selection of the best one
  \item Additional weights for preferences can be inserted
  \item Fast evaluation
  \item Non-parametric method, no distribution assumptions
  \item Visualization of clusters over distance is possible, identification of problem classes with similar algorithm behavior.
\end{itemize}

However, in addition to these advantages, there are also disadvantages:
\begin{itemize}
  \item Estimating uncertainty in the ranking is difficult
  \item The ranking does not have to induce a strict ordering, ties are possible
\end{itemize}

We generate the consensus ranking by respectively combining rankings of tuners (\gls{SPOT}, \ac{rs}, default) of individual experiments.
We always summarize the rankings of 9 experiments (3 repetitions of the tuner runs on each of 3 randomly drawn data sets).
Then, to aggregate across different variables related to the study subjects (e.~B. $m$, nfactors), we count how often the tuners reach a certain consensus rank.

This count is divided by the total number of experiments (or number of rankings) on the corresponding problem instances.
Thus, for each tuner, the number of times a particular Consensus Rank is achieved is obtained.

Simplified example:
\begin{itemize}
\item For case 2 (i.e., for a fixed choice of nnumericals, nfactors, cardinality, model, $m$, target variable), the comparison of SPOT, RS, and default methods resulted in the ranks
	\[
		\{1,3,2\}\quad 
		\{1,2,3\}\quad
		\{2,1,3\}.
	\]
	The consensus ranking for this case is $\{1,2,3\}$.
\item For case 2 (i.e., for \emph{another} fixed choice of nnumericals, nfactors, cardinality, model, $m$, target variable), the comparison of SPOT, RS, and default methods resulted in the ranks
	\[
		\{3,2,1\}\quad
		\{1,2,3\}\quad
		\{2,1,3\}.
	\]
	The consensus ranking for this case is $\{2,1,3\}$.
\item When both experiments are combined for an analysis, the frequencies for the obtained rankings are as follows:
	\begin{itemize}
		\item SPOT: rank 1 with 50\%, rank 2 with 50\%, rank 3 with 0\%.
		\item RS: rank 1 with 50\%, rank 2 with 50\%, rank 3 with 0\%.
		\item Default: rank 1 with 0\%, rank 2 with 0\%, rank 3 with 100\%.
	\end{itemize}
\end{itemize}

\subsubsection{Rank-Analysis: Classification}
\begin{figure}
\centering
\includegraphics[width=0.9\textwidth]{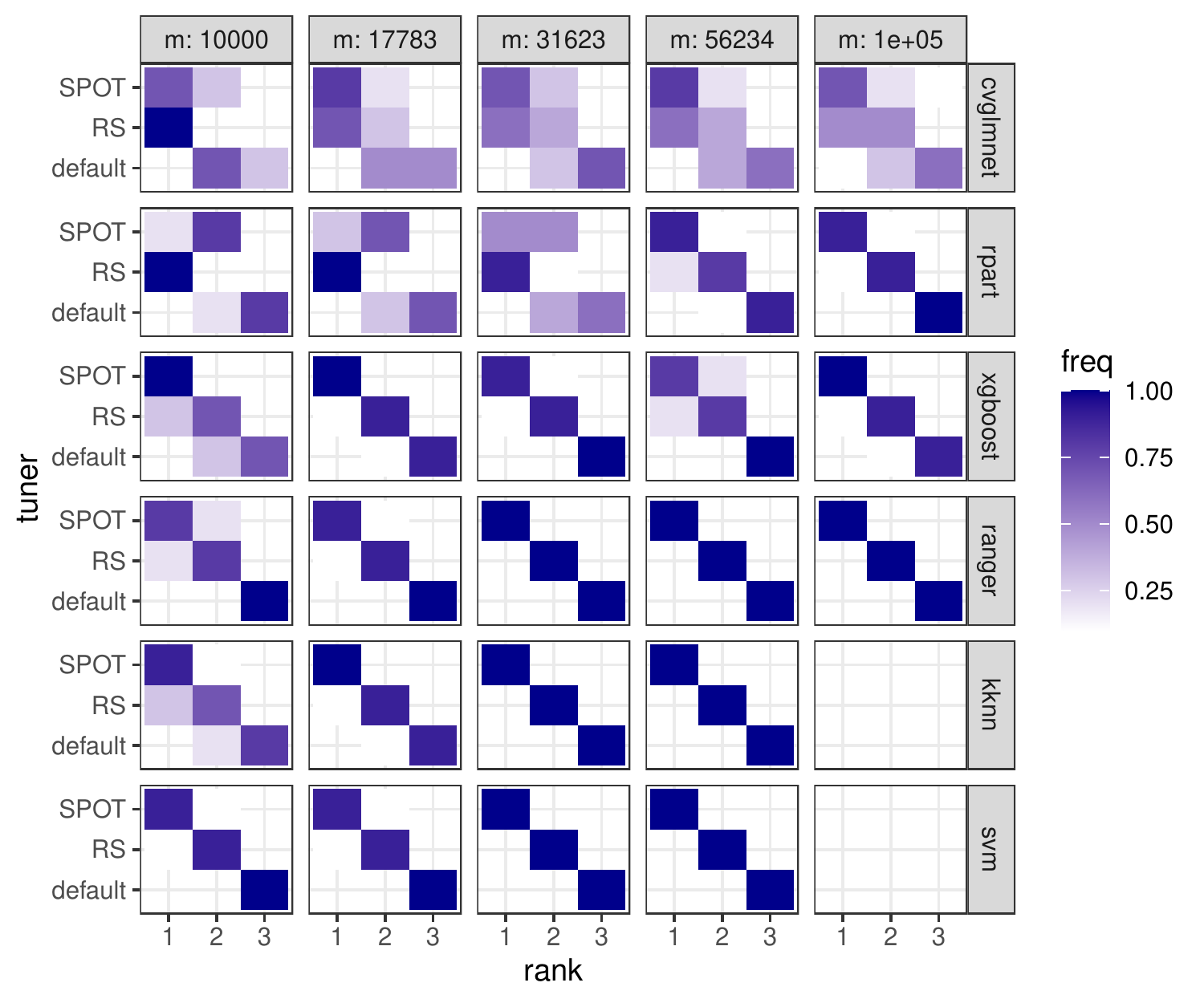}
\caption{Rank of tuners depending on number of observations ($m$) and model, for classification.} \label{fig:rankmodm}
\end{figure}

Based on the analysis method described in  Sec.~~\ref{sec:rankbas},
figure~\ref{fig:rankmodm} shows the relationship between the tuners, 
the number of observations $m$, and the optimized models.
It shows that mostly \ac{spot} and \ac{rs} beat the default setting and \ac{spot} also usually performs better than \ac{rs}.
However, some cases deviate from this. 
Especially, \modglmnet (\ac{en}) and \modrpart (\ac{eb}) seem to profit less from tuning: here
the distinction between the ranks of tuners is more difficult.
In addition, when the number of observations is small, there tends to be a greater uncertainty in the results.
These results can be partly explained by the required running time of the models.
With a smaller number of observations, the runtime of the individual models decreases, and \modglmnet and \modrpart
are the models with the lowest runtime (in the range of a few seconds, or below one second in the case of \modrpart).
If the evaluation of the models itself takes hardly any time, it is advantageous for \ac{rs} that the runtime overhead is low.
\ac{spot}, on the other hand, requires a larger overhead (for the surrogate model and the corresponding search for new parameter configurations).

\begin{figure}
\centering
\includegraphics[width=0.7\textwidth]{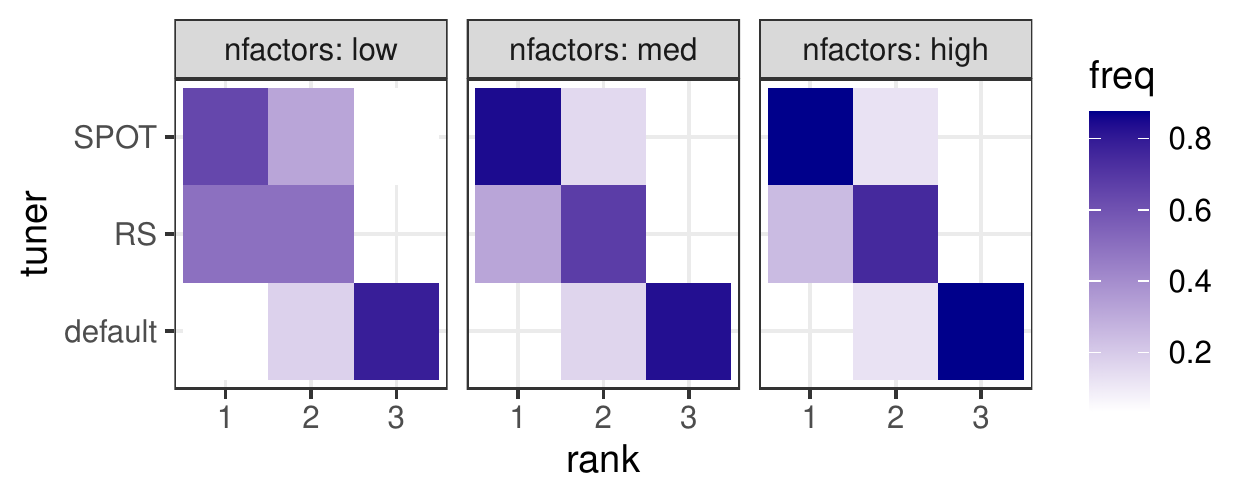}
\caption{Rank of tuners as a function of number of categorical features (nfactors) and model, for classification.} \label{fig:rankmodnf}
\end{figure}

Figure~\ref{fig:rankmodnf} shows the corresponding summary of the results depending on
the number of categorical features (nfactors).
Again, \ac{spot} usually performs best, followed by \ac{rs}.
There is a tendency for the greatest uncertainty to be found in case of a few categorical features.
Two explanations are possible: On the one hand, the reduction of features also means a reduction of the
required running time. 
On the other hand, the difficulty of the modeling increases, since with fewer features less information is available to separate the classes.

\begin{figure}
\centering
\includegraphics[width=0.7\textwidth]{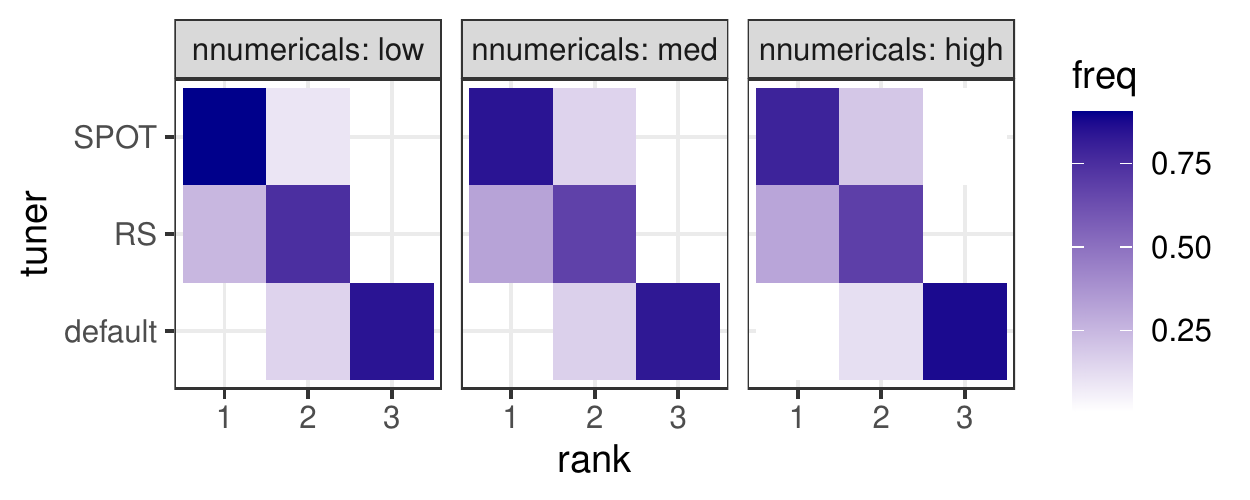}
\caption{Rank of tuners depending on number of numerical features (nnumericals) and model, for classification.} \label{fig:rankmodnn}
\end{figure}

The corresponding results for the number of numerical features can be found in Figure~\ref{fig:rankmodnn}.
There are hardly any differences, the number of numerical features seems to have little influence.
It should be noted that the data set contains fewer numerical than categorical features anyway.

\begin{figure}
\centering
\includegraphics[width=0.7\textwidth]{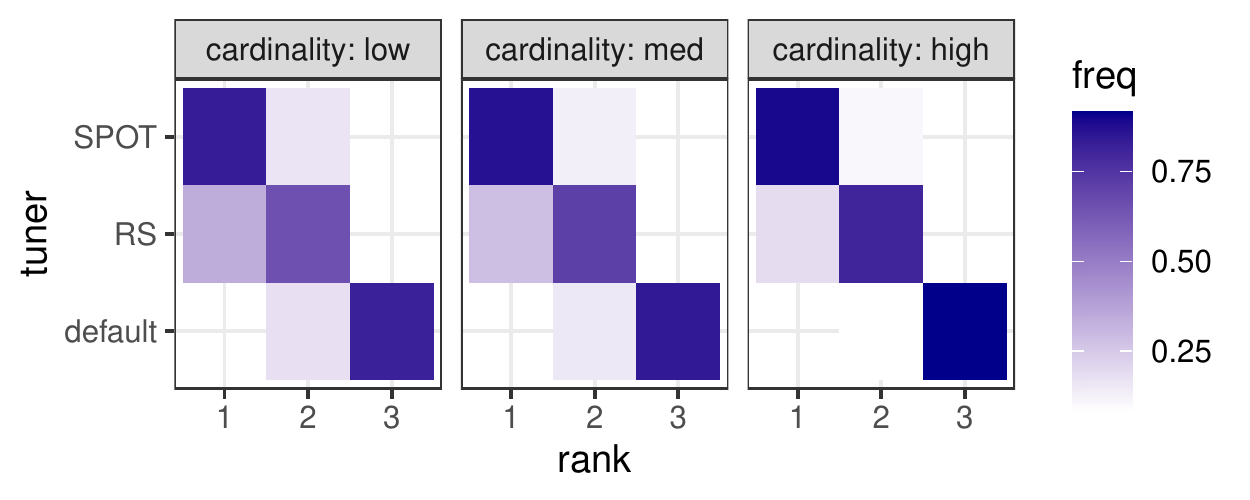}
\caption{Rank of tuners as a function of cardinality of categorical features and model for classification.
Note: This figure does not include the cases where the data set no longer contains categorical features (cardinality cannot be determined).} \label{fig:rankmodc}
\end{figure}

The cardinality of the categorical features also has little influence, see Figure~\ref{fig:rankmodc}.
However, there is a slight tendency: at higher cardinality, the distinction between the first rank (\ac{spot})
and second rank (\ac{rs}) is clearer.
This can be explained (similarly to nfactors) by the higher information content of the data set.

\begin{figure}
\centering
\includegraphics[width=0.9\textwidth]{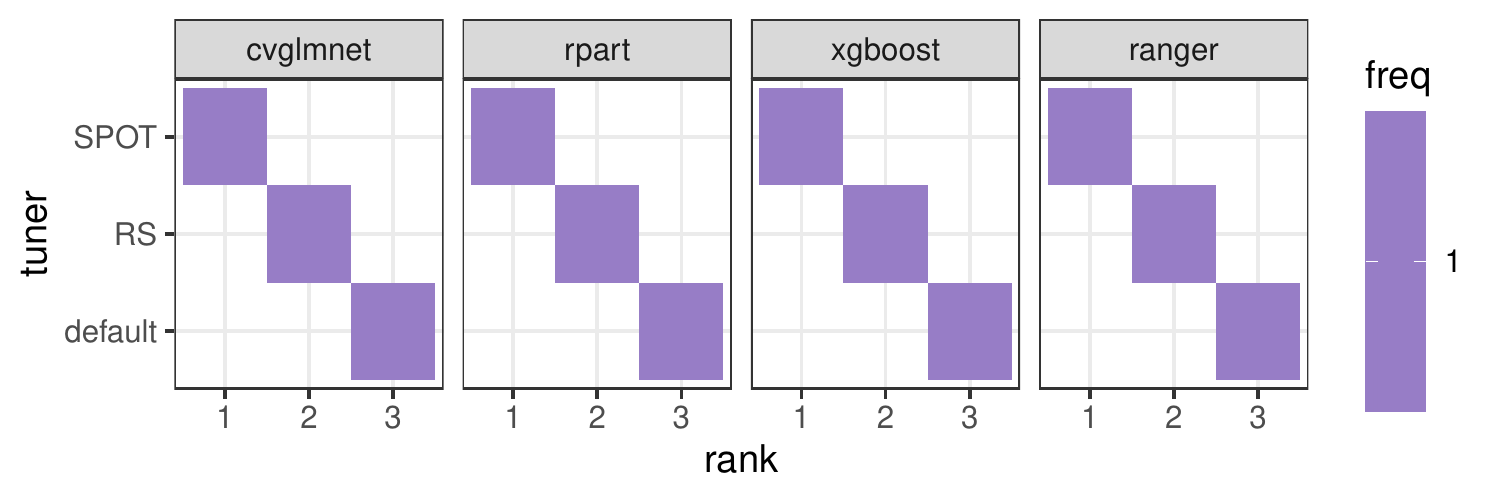}
\caption{Rank of tuners depending on the model for classification on complete data set.
Due to the increased runtime, \modkknn (\ac{knn}) and \modsvm (\ac{svm}) are not included.
} \label{fig:rankalldata}
\end{figure}

Finally, Figure~\ref{fig:rankalldata} shows the result of the tuners on the unmodified, complete data set.
For each case, \ac{spot} gets rank 1 and \ac{rs} rank 2.
This result is in line with the trends described above, since the complete data set contains the most information and also leads to the largest runtimes (for model evaluations).

\FloatBarrier
\subsubsection{Rank-Analysis: Regression}
\begin{figure}
\centering
\includegraphics[width=0.9\textwidth]{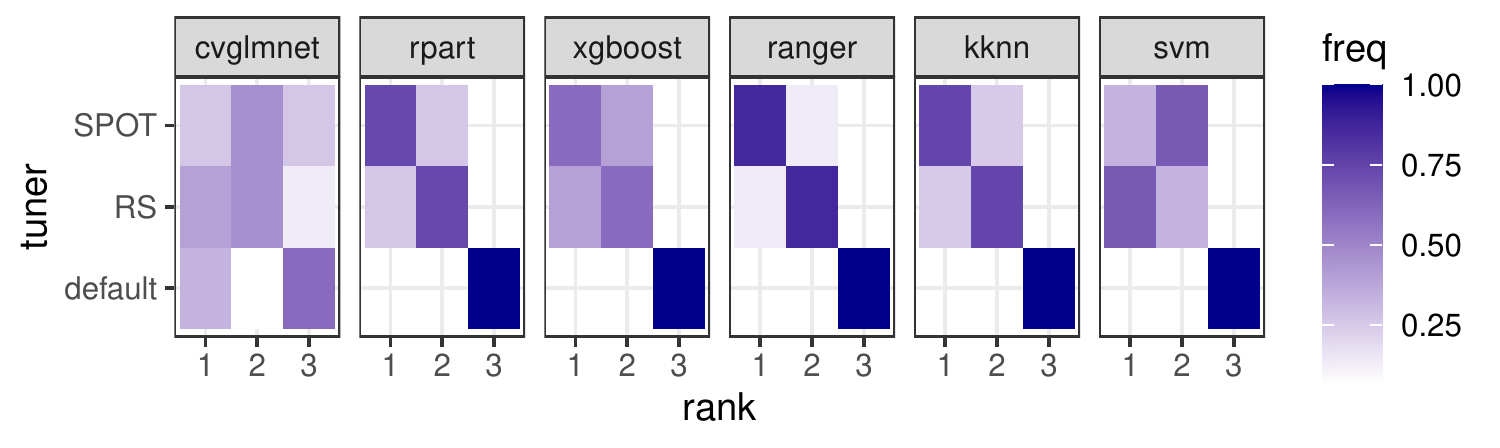}
\caption{Rank of tuners depending on model for regression.} \label{fig:rankmodreg}
\end{figure}

In addition to the classification experiments, a smaller set of experiments with regression as an objective were also conducted.
Figure~\ref{fig:rankmodreg} shows the results for this case separately for each optimized model.
Here, too, \ac{spot} is usually ranked 1. 
Unlike classification, however, there is more uncertainty.

For \modglmnet (\ac{en}), default values occasionally even achieve rank 1.
It seems that linear modeling with \modglmnet is unsuitable for regression with the present data set,
so that the tuners can hardly achieve differences to the default values.
This behavior was already indicated in Figure~\ref{fig:box3}.

In the case of \ac{svm}, \ac{rs} is more often ranked 1 than \ac{spot}, the reason for this is not clear.
Possible cause is a lower influence of the hyperparameters on the model goodness of fit of \ac{svm} for regression (compared to classification).
However, it should also be taken into account that for regression a smaller number of experiments is used for the evaluation.

\begin{figure}
\centering
\includegraphics[width=0.9\textwidth]{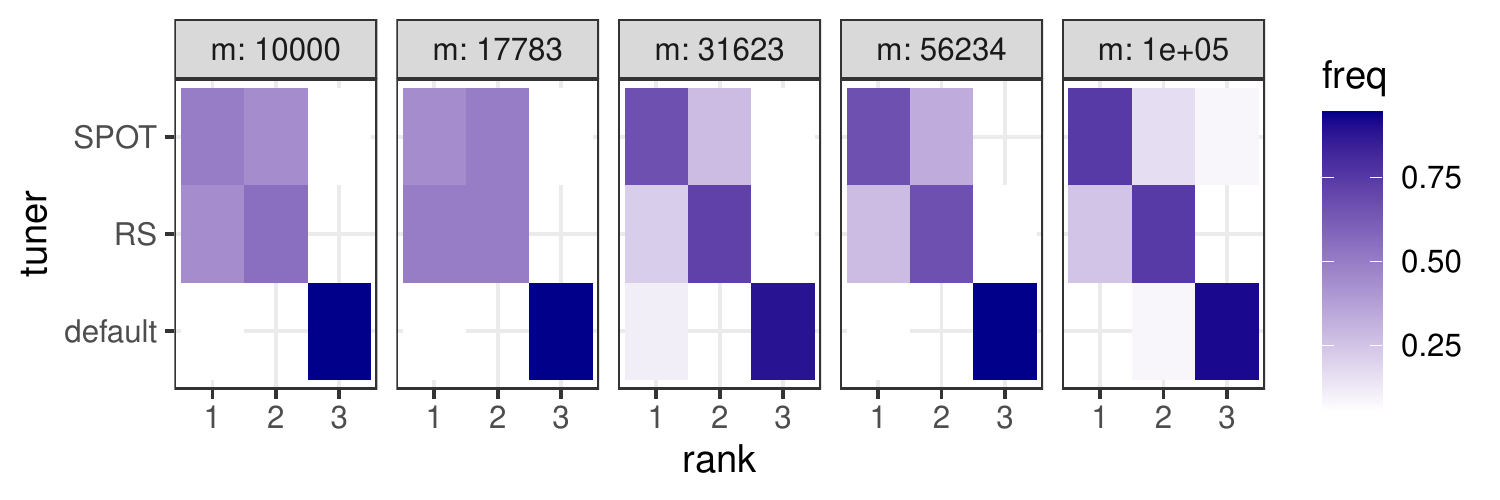}
\caption{Rank of tuners depending on number of observations ($m$) for regression.} \label{fig:rankmreg}
\end{figure}

The dependence on the number of observations $m$ is shown in Figure~\ref{fig:rankmreg}.
The results for regression with respect to $m$ are again largely consistent with those 
for classification.
With increasing number of observations, \ac{spot} is more clearly ahead of \ac{rs}.

\begin{figure}
\centering
\includegraphics[width=0.7\textwidth]{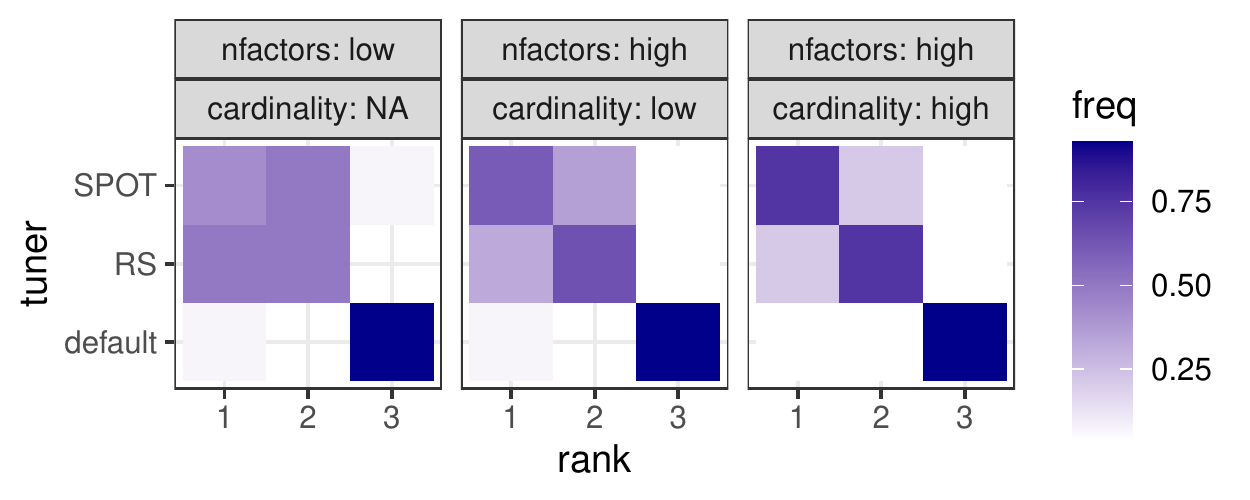}
\caption{Rank of tuners depending on cardinality and number of categorical features (nfactors) for regression.} \label{fig:ranknomodreg}
\end{figure}

The correlation with categorical features (number, cardinality) is also consistent with the classification results, see Figure~\ref{fig:ranknomodreg}. With larger number of features and larger cardinality, a clearer separation between the tuners is observed.

\subsubsection{Problem Analysis: Difficulty}\label{sec:schwierig}
In the context of this study, an interesting question arose as to how the difficulty of the modeling problem is related to the results of the tuning procedures.
We investigate this on the basis of the data obtained from the experiments.

In general, there are many measures of the difficulty of modeling problems in the literature.
An overview is given by~\citet{Lore19a}.
Of these measures, the overlap volume (F2, see~\citep{Lore19a}) is of interest, since it is easily interpretable and not specific to a particular model.
It is first determined for each feature individually how far the value ranges of two classes overlap.
This is called the \emph{overlap}\/ of the feature. 
The \emph{overlap volume}\/ of the data set is then the product of the individual overlap values.
However, this measure is unsuitable for categorical features (even after dummy coding). 
Furthermore, outliers are very problematic for this measure.

\begin{figure}
\centering
\includegraphics[width=0.7\textwidth]{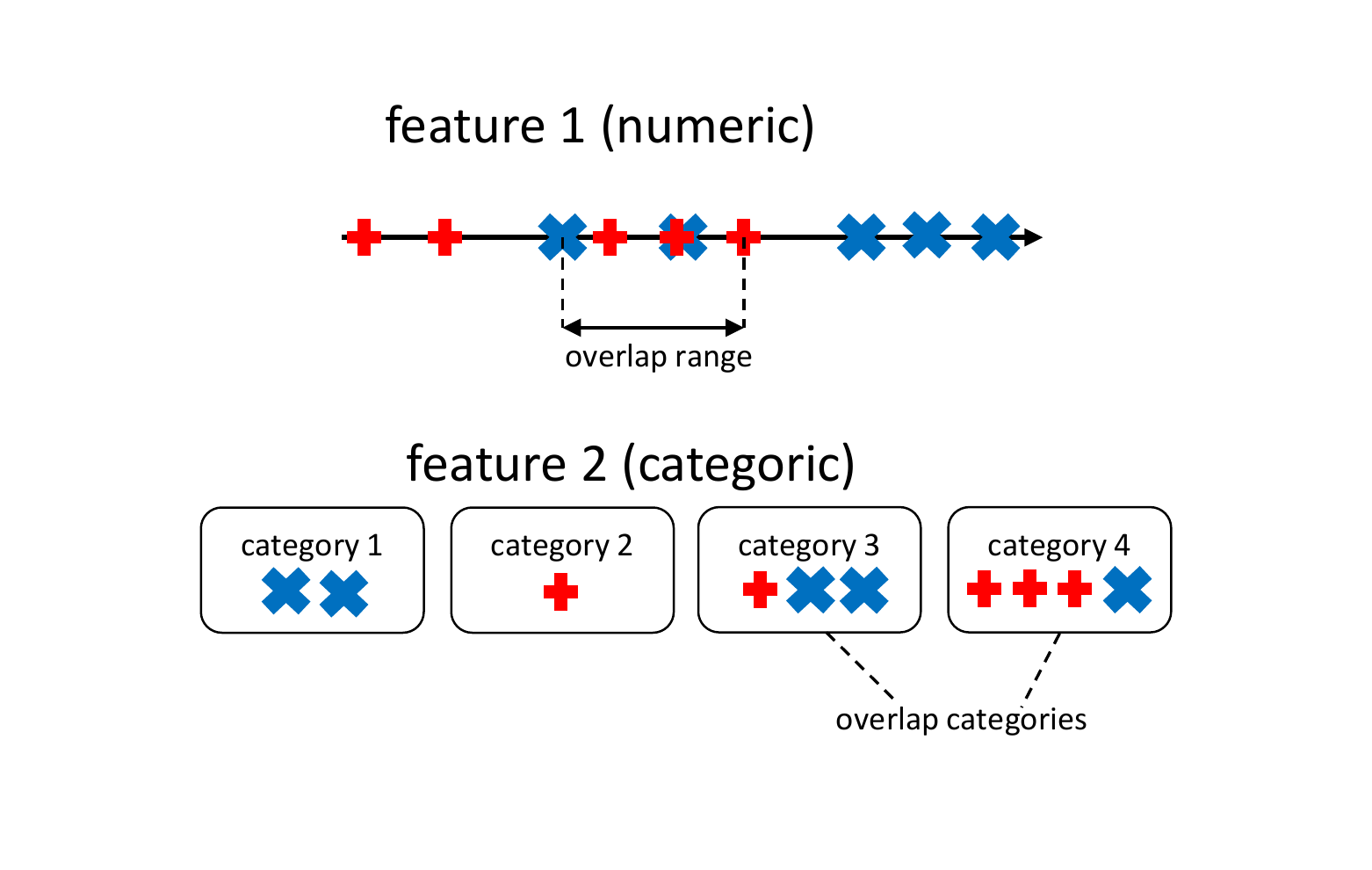}
\caption{Example of sample overlap for two features of a classification problem with two classes. 
The crosses show samples from the data set. 
All samples with class 1 are red (+).
All samples with class 2 are blue (x).
For feature 1, the overlap is 50\% (number of samples in the overlap area divided by the total number of samples).
For feature 2, the overlap is 70\% (number of samples in the overlap categories divided by total number of samples).
For both characteristics together, the sample overlap is $0.5 \times 0.7 = 0.35$.
} \label{fig:overlapexample}
\end{figure}

We therefore use a slight modification, 
by calculating the proportion of sample values for each feature,
which could occur in both classes (i.e. for which a swap of the classes based on the feature value is possible).
As an example, this is illustrated for a numeric and a categorical feature in Figure~\ref{fig:overlapexample}
for a numerical and a categorical feature.
For the overall data set, the individual overlap values of each feature are multiplied.
Subsequently, we refer to this measure as \emph{sample overlap}.

\begin{figure}
\centering
\includegraphics[width=0.99\textwidth]{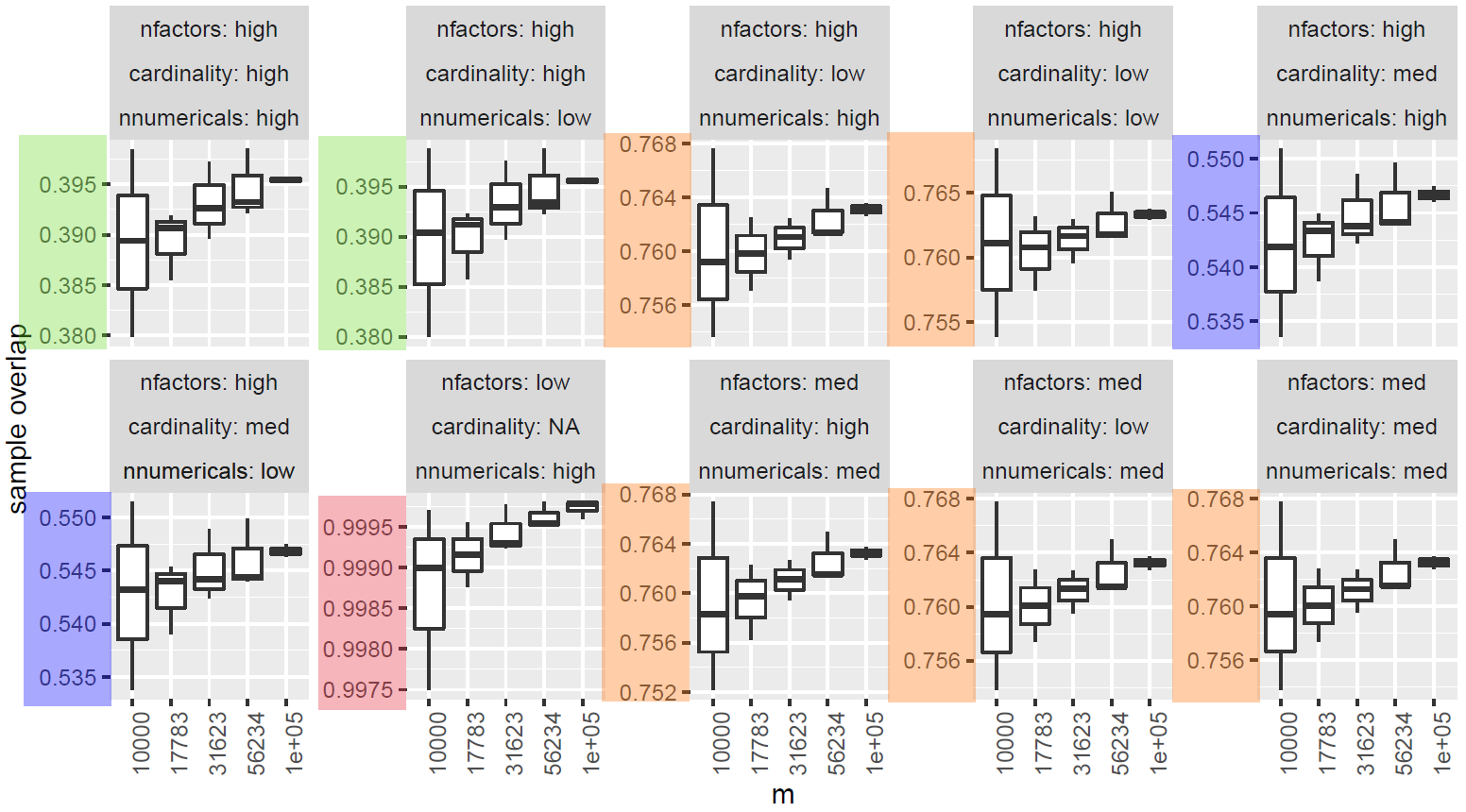}
\caption{Sample overlap depending on the properties of the data set varied in the experiments. The sample overlap is used as a measure of problem 
problem difficulty and leads to the definition of four difficulty levels (marked in color).} \label{fig:diff}
\end{figure}

Figure~\ref{fig:diff} shows the dependence of the sample overlap on our data properties ($m$, nfactors, nnumericals, cardinality).
Our data sets can be grouped into 4 difficulty levels based on these values of sample overlap:
\begin{enumerate}
\item Sample overlap $\approx 0.39$: nfactors=high and cardinality=high
\item Sample overlap $\approx 0.54$: nfactors=high and cardinality=med
\item Sample overlap $\approx 0.76$: all others
\item Sample overlap $\approx 1.00$: nfactors=low
\end{enumerate}
Here, 4 corresponds to the highest level of difficulty.
For the range relevant in the experiments, there is almost no change depending on $m$ or nnumericals.
For nfactors and cardinality a strong correlation can be seen. 

\begin{figure}
\centering
\includegraphics[width=0.99\textwidth]{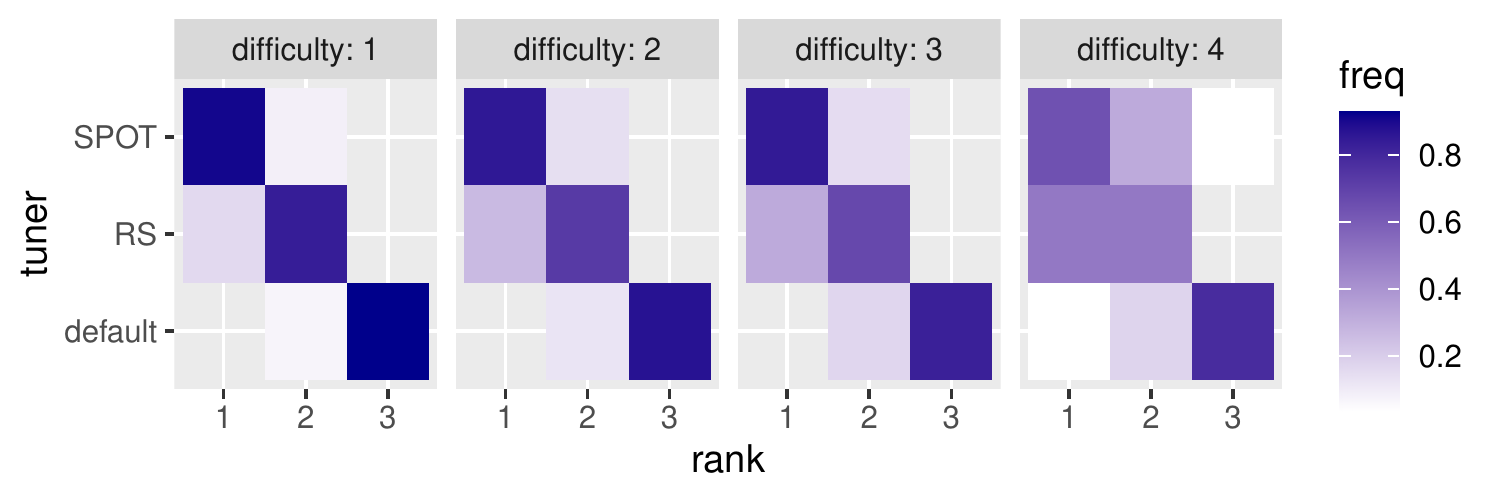}
\caption{Frequency of the achieved ranks of the tuners depending on the difficulty level.} \label{fig:rankdiff}
\end{figure}

Based on the 4 difficulty levels, the ranks already determined in previous sections can be re-ranked.
The result is summarized in Figure~\ref{fig:rankdiff}.
It turns out that as the difficulty of the problem increases, the rank differences are less robust.
That is, with larger sample overlap it is harder to estimate which tuner works best.
This is plausible with respect to the theoretical extreme case: 
A maximally difficult data set cannot be learned by any model, since the features no longer contain any information about the classes. 
In this case, the hyperparameters no longer have any influence and the tuners cannot improve.

\section{Conclusion}\label{sec:conclusion}

\subsection{Summary}
Finally, we discuss the hypotheses from  Sec.~~\ref{sec:frage}.
\begin{enumerate}[(H-1)]
\item \emph{Tuning is necessary:}
In almost all cases, significantly worse results are obtained with default settings.
In addition, the variance of the results decreases when tuners are used (see figures~\ref{fig:box2}~-~\ref{fig:box3}).
\item \emph{Data:}
Differences between tuners become apparent for data sets with a high information content and for larger data volumes.
If the number of levels or features (categorical and numerical) decreases, 
it can be observed that the differences between the tuners become smaller.

It has been confirmed that as the number of features and observations increases, the running time increases.
As the mean runtime of the models increases (i.e., for more complex models or larger data sets), \gls{SPOT} performs increasingly better than \ac{rs}, as the ratio of overhead to evaluation time decreases more for \gls{SPOT}.
\item \emph{Target variable:}
The execution of the tuning is not affected by the change of the target variable.
A peculiarity that requires further investigation occurred when tuning \ac{svm} for regression: \ac{rs} seems to perform better than \gls{SPOT}.
We recommend using a larger database for this case.
\item \emph{Model:} The choice of tuning method is not fundamentally influenced by the models.
Models that can be evaluated very quickly (e.g. \modrpart) benefit from the larger number of evaluations. 
This is due to the fact that time was chosen as a termination criterion.
\item \emph{Benchmark:} 
As described in  Sec.~~\ref{sec:rankbas}, analysis methods based on consensus ranking can be used to evaluate the suitability of tuners in a simple and statistically valid way.
\end{enumerate}

Overall, the result of the experimental investigation underlines, 
that tuning of machine learning models is essential to produce good configurations of the hyperparameters and thus models of high quality.

It is especially important to use tuners at all,
since the biggest differences are usually observed in comparison to default settings.
However, there are also differences between tuners, which in our investigation were mostly for the use of
\ac{spot}, i.e. a model-based method.
For the prospective focus on tuning with relatively large data sets (large $m$ and $n$), this is all the more evident, since the resulting high model runtime favors the use of model-based methods.

Since the number of hyperparameters is manageable ($<<100$), 
the addition of a few more parameters does not significantly increase the complexity of the tuning. 
For both Random Search\citep{Berg12a} and SPOT, the addition of a few parameters is harmless, 
even if they have only a small effect on the model quality. 

Rather, it results in a benefit: 
\begin{itemize}
\item Possible software bugs can be detected, 
\item seemingly unimportant parameters are detected as important, 
\item unknown interactions can be uncovered. 
\end{itemize}

The surrogate model allows to learn the influence of the hyperparameters on the model values. Unimportant parameters are thus automatically weighted less in the search.

Finally, we recommend for the selection of the tuner:
\begin{itemize}
\item Random Search can be used when either very little time (order of magnitude: time is sufficient for a single-digit number of sequential evaluations) but a lot of parallel computing capacity is available or when models can be evaluated extremely fast (seconds range).
\item 
If time-consuming computations with complex data and models need to be performed in more time (with relatively less parallel computing capacity), then we recommend model-based tuning with SPO. 
\item In exceptional cases, 
which deviate from the objective considered here for complex data and models (extremely large amount of computing time available, average evaluation times for the models), 
the use of surrogate model-free tuning methods can also be considered. 
\end{itemize}

\subsection{Special Case: Monotonous Parameters}\label{sec:mono}

\begin{table}[!htbp]
\centering
\caption{Global hyperparameter overview.
The column \enquote{Quality} shows all parameter, where a monotonous relationship between parameter values and model quality
is to be expected. 
(\upup: quality increases if parameter value increases, \updown: quality decreases if parameter value increases).
Correspondingly, the column \enquote{run time} shows the same information for the relationship of parameter values and run time.}
\begin{tabular}{l | r  | c | c }
\hline
Model & Hyperparameter & Quality & Run time\\
\hline
\ac{knn}&\paramk &&\upup\\
&\paramp &&\\
\hline
\ac{en}&\paramalpha &&\\
&\paramlambda  &&\\
&\paramthresh  &\updown & \updown \\
\hline
\ac{eb}&\paramminsplit &&\updown \\
&\paramminbucket &&\updown \\
&\paramcp &&\updown \\
&\parammaxdepth &&\upup \\
\hline
\ac{rf}&\paramnumtrees &\upup&\upup\\
&\parammtry  &&\upup \\
&\paramsamplefraction &&\upup \\
&\paramreplace && \\
&\paramrespect && \\
\hline
xgBoost &\parameta &&   \\
&\paramnrounds && \upup\\
&\paramlambda &&  \\
&\paramalpha && \\
&\paramsubsample &&\upup \\
&\paramcolsample &&\upup \\
&\paramgamma   &&\updown \\
&\parammaxdepthx  &&\upup \\
&\paramminchild  &&\updown \\
\hline
\ac{svm}&\paramkernel & & \\
&\paramdegree && \\
&\paramgamma  && \\
&\paramcoefz  && \\
&\paramcost  && \\
&\paramepsilon  && \\
\hline
\end{tabular}
\label{table:survey}
\end{table}

A special case are hyperparameters with monotonous effect
on the quality and runtime (and/or memory requirements) of the tuned model.
In our survey (see Table~\ref{table:survey}), two examples are included: \paramnumtrees (\ac{rf}), \paramthresh (\ac{en}).
Due to the monotonicity properties,
treating these parameters differently is an likely consideration.
In the following, we focus the discussion on \paramnumtrees as an example,
since this parameter is frequently discussed in literature and online communities~\citep{Prob18a}.

It is known from the literature that larger values of \paramnumtrees
generally lead to better models. 
As the size increases, a saturation sets in, leading to progressively lower quality gains.
It should be noted that this is not necessarily true for every quality measure.
\citet{Prob18a}, for example, show that this relation holds for log-loss and Brier score, but not for \ac{auc}.

Because of this relationship, Probst and Boulesteix recommend,
that \paramnumtrees should not be optimized.
Instead, they recommend setting the parameter to a ``computationally feasible large number''~\citep{Prob18a}.
For certain applications, especially for relatively small or medium sized data sets, we support this assessment.
However, at least in perspective, the present report considers tuning hyperparameters for very large data sets (many observations and/or many features). 
For this use case, we do not share the recommendation of Probst and Boulesteix,
because the required runtime of the model plays an increasingly important role and is not explicitly considered in the recommendation. 
In this case, a `computationally feasible large number' is not trivial to determine.

In total, we consider five alternatives:

\begin{enumerate}
\item Set manually: The parameter is set to the largest possible value that is still feasible with the available
computing resources.
Risks:
\begin{enumerate}
\item Single evaluations during tuning waste time unnecessarily.
\item Interactions with parameters (e.g. \parammtry) are not considered.
\item The value may be unnecessarily large (from a model quality point of view).
\item The determination of this value can be difficult, it requires detailed knowledge regarding: size of the data set, efficiency of the model implementation, available resources (memory / computer cores / time).
\end{enumerate}
\item Manual adjustment of the tuning: After a preliminary examination (as represented e.g. by the inital design step of \gls{SPOT})
a user intervention takes place. Based on the preliminary investigation, a value that seems reasonable is chosen by the user and is not changed in the further course of the tuning. 
Risks:
\begin{enumerate}
\item The preliminary investigation itself takes too much time.
\item The decision after the preliminary investigation requires intervention by the user (problematic for automation).
While this is feasible for individual cases, it is not practical for numerous experiments with different data (as in the experiments of the study in  Sec.~~\ref{sec:case3}).
Moreover, this reduces the reproducibility of the results.
\item Depending on the scope and approach of the preliminary study, interactions with other parameters may not be adequately accounted for.
\end{enumerate}
\item No distinction: parameters like \paramnumtrees are optimized by the tuning procedure just like all other hyperparameters.
Risks:
\begin{enumerate}
\item The upper bound for the parameter is set too low, so potentially good models are not explored by the tuning procedure. (Note: bounds set too tight for the search space are a general risk that can affect all other hyperparameters as well).
\item The upper bound is set too high, causing individual evaluations to use unnecessary amounts of time during tuning.
\item The best found value may become unnecessarily large (from a model quality point of view).
\end{enumerate}
\item Multi-objective: runtime and model quality can be optimized simultaneously in the context of multi-objective optimization. 
Risks:
\begin{enumerate}
\item Again, manual evaluation is necessary (selection of a sector of the Pareto front) to avoid that from a practical point of view irrelevant (but possibly Pareto-optimal) solutions are investigated.
\item This manual evaluation also reduces reproducibility.
\end{enumerate}
\item Regularization via weighted sum: The number of trees (or similar parameters) can be incorporated into the objective function.
In this case, the objective function becomes a weighted sum of model quality and number of trees (or runtime),
with a weighting factor $\theta$.
\begin{enumerate}
\item The new parameter of the tuning procedure, $\theta$, has to be determined.
\item Moreover, the optimization of a weighted sum cannot find certain Pareto-optimal solutions if the Pareto-front is non-convex.
\end{enumerate}
\end{enumerate}

In our experimental investigation, we use solution 3. 
That is, the corresponding parameters are tuned but do not undergo any special treatment during tuning. 
Due to the large number of experiments, user interventions would not be possible
and would also complicate the reproducibility of the results.
In principle, we recommend this solution for use in practice.

In individual cases, or if a good understanding of algorithms and data is available, solution 2 can also be used.
For this, \gls{SPOT} can be interrupted after the first evaluation step, in order to
set the corresponding parameters to a certain value or to adjust the bounds if necessary (e.g., if \paramnumtrees was examined with too low an upper bound).


\section*{Appendix}

\subsection*{Abbreviations}
\begin{acronym}
\acro{auc}[AUC]{Area Under the receiver operating characteristic Curve}
\acro{eb}[DT]{Decision Tree}
\acro{en}[EN]{Elastic Net}
\acro{knn}[\paramk-NN]{\paramk-Nearest-Neighbor}
\acro{ml}[ML]{Machine Learning}
\acro{mmce}[MCE]{mean Mis-Classification Error}
\acro{rf}[RF]{Random Forest}
\acro{rmse}[RMSE]{Root Mean Squared Error}
\acro{rs}[RS]{Random Search}
\acro{spot}[SPOT]{SPO Toolbox}
\acro{svm}[SVM]{Support Vector Machine}
\acro{xgb}[XGBoost]{eXtreme Gradient Boosting}
\end{acronym}

\bibliography{expertise}

\end{document}